\documentclass{article}

\usepackage{microtype}
\usepackage{graphicx}
\usepackage{subcaption}
\usepackage{booktabs}

\usepackage{hyperref}

\usepackage[accepted]{icml2026}

\usepackage{amsmath}
\usepackage{amssymb}
\usepackage{mathtools}
\usepackage{amsthm}

\usepackage{listings}
\usepackage{xcolor}
\usepackage{tikz}
\usetikzlibrary{shapes.geometric, arrows.meta, positioning, fit, backgrounds}
\usepackage{pgfplots}
\pgfplotsset{compat=1.18}
\usepackage{algorithm}
\usepackage{algorithmic}
\usepackage{multirow}
\usepackage{enumitem}
\usepackage{xspace}
\usepackage{booktabs}
\usepackage{tabularx}
\usepackage{siunitx}
\usepackage[table]{xcolor}
\usepackage{tcolorbox}
\tcbuselibrary{skins}

\newcommand{\sys}{ParEVO\xspace}

\definecolor{codegreen}{rgb}{0,0.5,0}
\definecolor{codeblue}{rgb}{0.1,0.1,0.8}
\definecolor{codepurple}{rgb}{0.58,0,0.82}
\definecolor{backcolour}{rgb}{0.96,0.96,0.96}

\lstdefinestyle{mystyle}{
    backgroundcolor=\color{backcolour},
    commentstyle=\color{codegreen},
    keywordstyle=\color{codeblue},
    numberstyle=\tiny\color{codepurple},
    stringstyle=\color{codepurple},
    basicstyle=\ttfamily\footnotesize,
    breakatwhitespace=false,
    breaklines=true,
    captionpos=b,
    keepspaces=true,
    numbers=left,
    numbersep=5pt,
    showspaces=false,
    showstringspaces=false,
    showtabs=false,
    tabsize=2,
    frame=lines
}
\lstdefinelanguage{Rust}{
  keywords={typeof, new, true, false, catch, function, return, null, catch, switch, var, if, in, while, do, else, case, break},
  keywords={true, false, unsafe, async, await, move, use, pub, crate, super, self, mod,
            struct, enum, fn, const, static, let, mut, ref, type, impl, dyn, trait, where,
            as, break, continue, if, else, while, for, loop, match, return, yield, in, extern},
  keywordstyle=\color{purple}\bfseries,
  ndkeywords={self, Self, u8, u16, u32, u64, f32, f64, i8, i16, i32, i64, usize, isize, bool, char, str, Vec, Option, Result, String, Box},
  ndkeywordstyle=\color{teal}\bfseries,
  comment=[l][\color{gray}\itshape]{//},
  morecomment=[s][\color{gray}\itshape]{/*}{*/},
  stringstyle=\color{green!50!black},
  string=[b]"
}

\usepackage[capitalize,noabbrev]{cleveref}

\theoremstyle{plain}

\theoremstyle{definition}

\theoremstyle{remark}

\icmltitlerunning{ParEVO: Synthesizing Code for Irregular Data}

\begin{document}

\twocolumn[
\icmltitle{ParEVO: Synthesizing Code for Irregular Data:
           High-Performance Parallelism through Agentic Evolution}

  \icmlsetsymbol{equal}{*}
\begin{icmlauthorlist}
\icmlauthor{Liu Yang}{equal,yale}
\icmlauthor{Zeyu Nie}{equal,yale}
\icmlauthor{Andrew Liu}{yale}
\icmlauthor{Ruomu Zou}{yale}
\icmlauthor{Deniz Altinbüken}{gdm}
\icmlauthor{Amir Yazdanbakhsh}{gdm}
\icmlauthor{Quanquan C. Liu}{yale}
\end{icmlauthorlist}

\icmlaffiliation{yale}{Department of Computer Science, Yale University, New Haven, CT, USA}
\icmlaffiliation{gdm}{Google DeepMind, Mountain View, CA, USA}

\icmlcorrespondingauthor{Quanquan C. Liu, Liu Yang, Zeyu Nie}{quanquan.liu@yale.edu, liu.yang.ly337@yale.edu, zeyu.nie@yale.edu}

\icmlkeywords{LLM for Code, High Performance Computing, Parallelism, Evolutionary Algorithms}

\vskip 0.3in
]

\printAffiliationsAndNotice{}
\begin{abstract}
Parallelizing code for \textbf{irregular data structures} (sparse graphs, unbalanced trees, non-uniform meshes) is notoriously hard, and current LLMs fail catastrophically on such tasks, generating code riddled with race conditions, deadlocks, and poor scaling. We address this with \textbf{\sys}, a framework for synthesizing high-performance parallel algorithms for irregular data, built on three contributions: the \textbf{Parlay-Instruct Corpus} of 13,820 tasks generated via a Critic-Refine'' pipeline that filters for empirically performant uses of Work-Span primitives; specialized \textbf{DeepSeek}, \textbf{Qwen}, and \textbf{Gemini} models fine-tuned to the semantics of the ParlayLib library; and an \textbf{Evolutionary Coding Agent (ECA)} that repairs the last mile'' of correctness using compiler and profiler feedback. On the ParEval benchmark, \sys achieves an average \textbf{$107\times$ speedup} and a \textbf{$13.6\times$ speedup} on highly complex irregular graph problems, outperforming commercial models like GPT-5-Thinking and Gemini-3-Pro, while matching expert \emph{human-written} baselines and reaching up to a \textbf{$4.1 \times$ speedup} on kernels such as Maximal Independent Set. This demonstrates that AI-driven agents can effectively navigate the complex landscape of high-performance computing. Source code and datasets are available at \url{https://github.com/WildAlg/ParEVO} \citep{ParlayInstruct2026}.
\end{abstract}

\section{Introduction}

The breakdown of Dennard scaling and the subsequent stagnation of single-core frequency scaling have fundamentally shifted the computing paradigm. Performance improvements in modern software are now almost exclusively driven by parallelism, whether through multi-core CPUs, GPUs, or distributed clusters \cite{Sahu_2019}. While ``regular'' parallelism (e.g., dense matrix multiplication) is well-understood and supported by mature libraries, \textbf{irregular parallelism} remains a grand challenge in High-Performance Computing (HPC).

Irregular algorithms, which operate on graph structures, sparse matrices, or adaptive meshes, are characterized by unpredictable memory access patterns and dynamic work distribution. In these regimes, the computational cost of processing a node or element depends on runtime data, making static load balancing ineffective. Writing efficient code for these problems requires sophisticated techniques like work-stealing, dynamic scheduling, and lock-free synchronization \cite{nichols2024pareval}.

Current Large Language Models (LLMs) struggle profoundly with this domain. Trained primarily on sequential Python or standard C++ code from GitHub, they exhibit strong ``sequential bias.''  When asked to parallelize a graph traversal, they often attempt to wrap a standard Breadth-First Search (BFS) in a naive parallel loop (\texttt{\#pragma omp parallel for}), ignoring the race conditions inherent in updating the `visited' array. Alternatively, they may introduce coarse-grained locks that serialize execution, rendering the parallel code slower than its sequential counterpart \cite{kambhampati2024llmscantplanhelp}.

We argue that the solution lies not in teaching LLMs to write low-level threading primitives (like `pthreads' or `std::thread'), which are error-prone and hard to compose, but in leveraging high-level algorithmic primitives. \textbf{ParlayLib} \cite{parlaylib} provides a suite of such primitives (e.g., `filter', `pack', `scan', `sort', `reduce') that abstract away the complexities of scheduler management. By training LLMs to map natural language intent to these primitives, we can generate code that is \textit{correct by construction} and mathematically provable to scale.

To this end, we introduce \textbf{\sys}, an end-to-end system for synthesizing high-performance parallel code. We detail the following main contributions:
\begin{itemize}
    \item \textbf{Data-Centric Synthesis:} We introduce the \textit{Parlay-Instruct} corpus, a dataset of 13,820 parallel coding tasks. Unlike previous datasets scraped from GitHub (which often contain broken code), our data is synthesized via a ``Teacher-Student'' pipeline and verified against a ground-truth compiler oracle. We provide a novel performance dataset generation technique focused on graph problems curated from the selection
    of well-known programming competitions curated by the online-judge DMOJ~\cite{dmoj}. 
    \item \textbf{DeepSeek-Parlay, Qwen-Parlay, Qwen-Rust, and Gemini-2.5-Parlay\footnote{\url{https://huggingface.co/qqggez/deepseek-parlay-6.7b}, \url{https://huggingface.co/qqggez/qwen3-30b-sft-stage2-merged}, \url{https://huggingface.co/YangLiuWillow/qwen3_rust_dpo_final_merged}}:} We release a fine-tuned 6.7B parameter Deepseek model for C++ \citep{deepseekparlay2026}, two fine-tuned 30B parameter Qwen3 models---one for C++  \citep{qwenparlay2026} and one for Rust  \citep{qwenrust2026}---and a Gemini-2.5-Pro model fine-tuned for C++. These models outperform some larger closed-source and open-source models on parallel reasoning tasks by internalizing the data structures, semantics, primitives, and algorithms of the state-of-the-art ParlayLib library~\citep{parlaylib} and safe parallel Rust patterns.
    \item \textbf{Evolutionary Refinement:} We formalize both the data synthesis step and the final code generation process as an evolutionary search over the space of Abstract Syntax Trees (ASTs). Our agent generates a population of candidate solutions, compiles them, runs them against performance tests, and uses the error logs (or performance profiles) as ``fitness functions'' to drive mutation and crossover operations in the prompt space. 
    \item \textbf{The Correctness-Speedup Trade-off:} We identify an ``alignment tax'' for concurrent programming. Our evaluation reveals that fine-tuning enables models to write significantly safer code at the expense of slightly slower peak performance (i.e. higher Pass@1 rates with lower Speedup@1 rates). This trade-off occurs because fine-tuned models learn to conservatively avoid raw, risky atomics in favor of stable, high-level primitives (such as \texttt{parlay::unique}).
\end{itemize}

Specifically, our paper succeeds in the following task:

{\centering \emph{\sys democratizes parallel computing for irregular data by fine-tuning LLMs on verified primitives and deploying an evolutionary agent to iteratively optimize code based on runtime performance feedback.}}

All code can be found at \url{https://github.com/WildAlg/ParEVO} \citep{ParlayInstruct2026}.

\paragraph{Conflict of Interest Disclosure.}
D.A. and A.Y. are employed by Google DeepMind, which develops the Gemini family of models. This paper evaluates and uses Gemini models, including Gemini-2.5-Flash, Gemini-2.5-Pro, Gemini-3-Pro, and a fine-tuned Gemini-2.5-Pro variant referred to as Gemini-2.5-Parlay. This work was supported in part by a Google Academic Research Award.

\section{Related Work}
\label{sec:related}

\textbf{LLMs for Code Generation.}
Large Language Models have fundamentally shifted the landscape of software engineering, achieving remarkable success in sequential code completion \citep{HUSEIN2025103917}, summarization \citep{ahmed2022learningcodesummarizationsmall}, and translation \citep{eniser2024translatingrealworldcodellms}. Evaluation metrics have similarly evolved from surface $n$-gram overlap to structure-aware measures like CodeBLEU \citep{ren2020codebleu}, which better correlate with functional correctness. However, current models struggle with complex planning and reasoning tasks \citep{kambhampati2024llmscantplanhelp}, a limitation that is magnified in High-Performance Computing (HPC). \citet{nichols2024pareval} demonstrated via the ParEval benchmark \citep{ParEvalRepoGitHub2024} that while LLMs can generate syntactic structures for frameworks like Kokkos and MPI, they often fail to capture the semantic nuances of synchronization and race conditions. Recently, ParEval-Repo \citep{ParEvalRepo2025, ParEvalRepoArXiv2025} extended this evaluation to repository-level HPC translation tasks (e.g., multi-file codebases, build systems), highlighting that scaling beyond individual kernels introduces qualitatively different failure modes. Our work addresses this by moving beyond general-purpose pre-training, targeting the qualitatively harder regime of \emph{parallel} and \emph{irregular} algorithms where correctness requires respecting concurrency and performance depends on minimizing span.

\textbf{Automated Parallelization and HPC Translation.}
Prior efforts in automated parallelization have largely focused on translating serial loops to OpenMP directives. BabelTower \citep{BabelTower2022} previously tackled auto-parallelized program translation from sequential C to CUDA via a learning-based framework leveraging large-scale corpora and back-translation with reranking. OMPGPT \citep{Chen_2024} fine-tunes GPT-Neo to predict pragmas for regular loops, while AutoParLLM \citep{mahmud2023autoparllmgnnguidedautomaticcode} uses Graph Neural Networks to guide LLM generation based on parallelism patterns. \citet{tehranijamsaz2024coderosetta} attempts unsupervised translation between languages and their HPC extensions but lacks a feedback mechanism for correctness. To address correctness risks such as subtle parallel bugs that often arise in serial-to-CUDA/OpenMP translation, MuSL \citep{MuSL2025, MuSLCode2025, MuSLTranslatorModel2025} proposed a mutual-supervision loop where a translator and a test-generator co-evolve: the tester synthesizes unit tests to filter translations, and the translator produces code to improve the tester. More recently, UniPar \citep{bitan2025unipar} introduced a multi-agent framework for translating code between serial, OpenMP, and CUDA formats. While UniPar evaluates functional correctness (achieving 33\%), it does not explicitly optimize for or benchmark the runtime scalability (work-span) of the generated algorithms. In contrast, \sys specifically targets \textit{irregular} data, such as graph traversals and sparse matrix operations, where correct translation is insufficient, and \emph{performance speedup} via parallelism, software engineering techniques, and performant algorithms and data structures is key. Recent advances have further specialized LLMs for parallel domains. For instance, \citet{chaturvedi2024hpccoderv2} successfully fine-tuned base models on the HPC-Instruct dataset to target low-resource parallel languages, demonstrating that smaller, specialized models can match proprietary models on the ParEval benchmark. Similarly, frameworks like MARCO \citep{rahman2025marco} and PerfCoder \citep{yang2025perfcoder} utilize multi-agent reasoning and execution trajectories to separate code generation from performance tuning. However, while these frameworks primarily target traditional imperative paradigms like OpenMP and CUDA, ParEVO specifically targets the algorithmic complexities of irregular data by grounding the model in the composable semantics of ParlayLib.

\textbf{Structured Reasoning and Agentic Coding.}
To transcend the stochastic limitations of single-shot generation, frameworks like Reflexion \citep{shinn2023reflexion} use verbal reinforcement to iteratively correct failures. More recently, this paradigm has been extended via evolutionary search. Building upon this, EvoTune \citep{EvoTune2025} augments LLM-based evolutionary program search by periodically updating the model via reinforcement learning on search-derived signals. Similarly, AI tree search systems \citep{TreeSearchEmpiricalSoftware2025} embed LLM-based code mutation within a search procedure to maximize a measurable quality metric. To benchmark these search processes, AlgoTune \citep{AlgoTune2025, AlgoTuneDataset2025} introduced a suite for numerical programs and evaluated an agent that iterates by editing, compiling, timing, and selecting the fastest valid variant. Concurrent open-source works such as OpenEvolve \citep{OpenEvolve2025} have demonstrated the efficacy of coupling LLMs with genetic algorithms \citep{assumpcao2025codeevolve, khrulkov2025gigaevo, novikov2025alphaevolve} and Quality-Diversity metrics (e.g., MAP-Elites) to prevent diversity collapse during program synthesis. ParEVO brings this evolutionary paradigm to the HPC domain, replacing standard unit-test fitness functions with rigorous hardware profiling.

\textbf{Abstractions for Irregular Parallelism.}
A core theme in parallel algorithmics is that abstraction choice determines accessibility. The classic work-span model \citep{brent1974parallel} and work-stealing schedulers \citep{blumofe1999workstealing} provide a principled foundation for nested parallelism. High-level libraries like ParlayLib \citep{parlaylib, ParlayLib2020} expose this theory through composable primitives (e.g., \texttt{scan}, \texttt{reduce}, \texttt{filter}), making provably efficient algorithms more accessible. Similarly, specialized abstractions such as GraphIt \citep{GraphIt2018, GraphItRepo2018} separate algorithm specification from scheduling choices to enable systematic performance tuning for irregular graph workloads, while Ligra \citep{Ligra2013} provides a lightweight shared-memory graph processing framework with simple vertex/edge mapping primitives and density-adaptive traversal strategies. Benchmarks like PBBS \citep{shun2012pbbs, anderson2022pbbsV2} and Rusty-PBBS \citep{abdi2023rustypbbs} formalize the evaluation of these irregular workloads. \sys leverages these insights by training models to target primitive-based code-writing within the Parlay ecosystem, ensuring that generated code is not just a parallel loop, but a structurally sound parallel algorithm capable of handling load imbalance inherent in irregular data \citep{Sahu_2019, social_graph}.

\textbf{Test-Time Compute and Execution Feedback.} A growing consensus indicates that standard Supervised Fine-Tuning (SFT) and text-based reflection are insufficient for generating highly optimized code. Consequently, the field has rapidly shifted toward integrating real-machine execution feedback into the LLM reasoning loop. Using empirical hardware profiling as a direct reward signal drastically improves kernel efficiency \citep{du2025afterburner, merouani2025agentic, lei2025pragma}. Crucially, \citet{singh2024testtime} applied test-time program search to the ParEval benchmark and empirically proved that LLMs exhibit a severe capability gap when attempting to act as their own ``verifiers'' for parallel code. This limitation directly motivates ParEVO's Evolutionary Coding Agent (ECA), which sidesteps the unreliable ``LLM-as-a-judge'' paradigm in favor of treating deterministic compilers and sanitizers as ground-truth adversarial critics.

\paragraph{ML for Compiler/Performance Optimization.} ML increasingly replaces heuristics for phase ordering, register allocation, auto-vectorization \citep{cummins2022compilergym, trofin2021mlgo, hajali2020neurovectorizer}, and tensor scheduling \citep{chen2019learningoptimizetensorprograms, zheng2020ansor}. Foundation models \citep{cummins2025llm, tang2025reasoning} optimize IR using MCTS. For parallelization, OMPGPT, AutoParLLM, and UniPar predict pragmas for regular loops. However, they operate at lower stack levels, lack robust feedback, or target regular matrices. ParEVO uniquely synthesizes high-level parallel abstractions for irregular data where vectorization fails.

\paragraph{Program Synthesis/Repair with Feedback.} Agentic workflows increasingly use execution environments as oracles \citep{yang2023intercode, chen2024teaching, ni2023lever, zhou2024language}. While text-based critique and RL use compiler diagnostics, ParEVO advances this by tying selection pressure to scalable runtimes and problem-specific tests written by human experts that are designed to catch data races that lead to incorrect output and code that degrade performance.

\section{Methodology: The \sys System}

\sys is composed of three distinct stages: (1) Data Synthesis through Evolutionary Search, (2) Supervised Fine-Tuning, and (3) Inference-Time Evolutionary Search.

\subsection{Stage 1: The Parlay-Instruct Fine-Tuning Dataset Corpus}
The primary bottleneck for training ``HPC-aware'' LLMs is data scarcity. High-quality parallel C++ code is rare on GitHub compared to React components or Python scripts. We generated a synthetic dataset which incorporates parallel performance constructs, syntax, software engineering techniques, data structures, and algorithms, using a ``Teacher-Student-Critic'' pipeline via OpenEvolve~\cite{sharma2025openevolve}. {This synthetic dataset contains three parts: (1) the ParlayLib primitives, (2) DMOJ slow-fast code comparison pairs, and (3) DMOJ problem-solution pairs with labeled status, runtime performance, and any compiler or runtime error messages.}

\subsubsection{Seed Generation and Mutation}
We manually authored 593 ``golden'' examples covering ParlayLib's core primitives and 20 problems from DMOJ~\cite{dmoj}. We then used Gemini-3-Pro (the ``Teacher'') to mutate these seeds. We defined three mutation operators $\mathcal{M}$: 
\begin{enumerate}
\vspace{-1em}
    \item \textbf{Type Mutation ($\mathcal{M}_{type}$):} Changes the underlying data type (e.g., `int' $\rightarrow$ `std::string' or custom `struct Point'). This forces the model to learn C++ template instantiation rules.
    \item \textbf{Constraint Mutation ($\mathcal{M}_{cons}$):} Adds logical predicates (e.g., ``Sort only odd numbers'' $\rightarrow$ requiring a `filter' then `sort'). This forces the composition of primitives.
    \item \textbf{Algorithmic Mutation ($\mathcal{M}_{algo}$):} Transforms the problem structure, e.g., converting a `reduce' problem into a `scan' (prefix sum) problem.
\end{enumerate}
We ran 5 passes of 10 mutations per seed, sampled from the operators $\{\mathcal{M}_{type}, \mathcal{M}_{cons}, \mathcal{M}_{algo}\}$ defined above, plus an additional 50 tasks targeting complex primitives (e.g., \texttt{parlay::delayed::filter}). This produced an initial pool of 29{,}700 candidates.

\subsubsection{The Critic Loop: Rejection Sampling}
Let $P$ be a generated problem and $C$ be the generated code. We accept $(P, C)$ into the dataset if and only if:
\begin{equation}
\text{Compile}(C) \land \text{UnitTest}(C)
\end{equation}
That is, we only accept code that compiles and passes the unit tests. Compiling and executing each candidate against its unit test discarded 15{,}880 candidates that failed to compile or timed out.  This filtration process yielded 13,820 verified instruction-tuning pairs, which we partitioned into a fine-tuning training set of size 13,120 and a held-out test set of 700 pairs for evaluation.

\paragraph{Performance Optimization Dataset.}
To enable the model to reason about runtime efficiency, we curated a benchmark of 20 challenging graph problems from the DMOJ competitive programming platform \cite{dmoj}. We synthesized optimization trajectories for these problems using the OpenEvolve framework \cite{sharma2025openevolve} powered by Gemini-3-Pro. The data generation process followed the following novel protocol:

\begin{enumerate}
    \item \textbf{Agent Initialization:} The agent was provided with the problem description and ParlayLib documentation, with a dual objective function minimizing both test failures and execution time.
    \item \textbf{Trajectory Extraction:} We recorded the agent's iterative refinements, extracting pairs of solutions $(C_{\text{base}}, C_{\text{opt}})$ from the evolutionary history.
    \item \textbf{Speedup Threshold:} To ensure high-quality training signal, we filtered for pairs where the optimized solution $C_{\text{opt}}$ achieved a runtime speedup of at least $1.2\times$ over $C_{\text{base}}$.
\end{enumerate}





We constructed pairwise comparison examples using the solution pairs identified in the previous step. To eliminate positional bias, we randomized the assignment of ``Code A" and ``Code B" so that the faster implementation appears in either position with equal probability. The model is trained to identify the more performant solution using the following format:

\begin{quote}
    \textbf{Instruction:} Determine which of the two code solutions has better performance.
    
    \textbf{Input:}\\
    Code A: [Source Code]\\
    Code B: [Source Code]
    
    \textbf{Output:}
    [Label of the Faster Solution]
\end{quote}

A concrete example of this comparison format is provided in \cref{fig:ft-data-code-comparison}.

While learning on performance edits has been used in \citep{shypula2024learningperformanceimprovingcodeedits}, our dataset is distinct in its focus on the complex, global transformations required for \textit{irregular parallelism}, rather than the local sequential optimizations primarily targeted in the prior work.

\paragraph{Rust Parlay Primitives}
Given that the distinct Rust primitives were insufficient to constitute a robust fine-tuning dataset, we opted to include them directly in the context window. This approach allowed us to leverage the models' pattern-matching and in-context learning capabilities without the need for parameter updates. To support this process, we integrated a full suite of Parlay-equivalent Rust primitives derived from \textbf{RPB}~\cite{rpb_repo}. Furthermore, to support the generation of higher-complexity algorithms, we manually implemented the delayed execution primitives in Rust and supplied them as immutable reference implementations within the system prompt.

\paragraph{Rust Evolutionary Dataset.}
To train the evolutionary coding agent for the Rust domain, we constructed a specialized dataset derived from the DMOJ benchmark execution logs. We aggregated the raw logs to extract code solutions, runtime metrics, and error traces. The data underwent a rigorous cleaning pipeline: we first filtered out irrelevant infrastructure failures (e.g., permission errors) and removed the held-out test set. We then deduplicated the remaining entries, prioritizing successful submissions while retaining a diverse set of failing attempts characterized by distinct error messages. The final corpus was serialized into JSONL format, where each entry explicitly pairs a problem description with the corresponding code, execution status, runtime performance, and any resulting compiler or runtime error messages. This rich metadata distinguishes our dataset from standard code corpora, enabling the model to learn both correct optimization patterns and specific error-correction strategies. Such a detailed corpus of training data is necessary for Rust given that Rust is notoriously difficult to use for irregular parallelism; hence, the available training data (including errors and compile-time messages) is rare for this language in the available base models.

\subsection{Stage 2: Fine-Tuning DeepSeek, Gemini-2.5, Qwen3 for ParlayLib and Rust RPB}
We selected \textbf{DeepSeek-6.7b-base} and \textbf{Qwen3-Coder-30B-A3B-Instruct} as our open-source backbones due to their strong performance on standard C++. These models represent a tiered architecture strategy: DeepSeek-6.7b serves as our efficient, lightweight baseline, while Qwen3 acts as our high-capacity large model. We fine-tuned the model using Low-Rank Adaptation (LoRA)~\cite{hu2022lora} to minimize compute costs while preserving the base model's reasoning capabilities.
We selected \textbf{Gemini-2.5-Pro}
as our third base model due to its extensive context window for handling complex, long-context scenarios. All three models underwent fine-tuning to align them with our specific domain requirements.


\paragraph{Training Configuration.} 
We configured the training pipeline according to model scale. For \textbf{DeepSeek-6.7b-base}, we executed single-stage Supervised Fine-Tuning (SFT) on an NVIDIA RTX 5000 Ada machine. We targeted the query and value projections using LoRA ($r=8, \alpha=16$) and trained on a combined dataset of ParlayLib syntax and `slow-fast' performance pairs (FP16, learning rate $2\text{e-}4$).

For the larger \textbf{Qwen3-Coder-30B-A3B}, we implemented a dual-stage alignment pipeline on an NVIDIA H200 GPU. The first stage established domain capability via SFT on ParlayLib syntax and standard DMOJ solutions, using QLoRA ($r=16, \alpha=32$) across all linear attention and MLP layers. The second stage applied Direct Preference Optimization (DPO) to explicitly suppress failure modes. In this phase, we trained on contrastive triplets (pairing passing solutions against failing or inefficient implementations) using a reduced learning rate of $5\text{e-}6$ and $\beta=0.1$.
\vspace{-1em}
\paragraph{Evaluation Environment.} 
Performance benchmarks were conducted on a dual-socket compute node featuring two Intel Xeon Platinum 8562Y+ processors (64 physical cores total). To ensure consistent comparisons across frameworks, all experiments use 32 threads for OpenMP, ParlayLib, and Rust unless otherwise specified.


\subsection{Stage 3: Evolutionary Coding Agent (ECA)}\label{sec:eca}
\paragraph{Evolutionary Search Strategy.}
To transcend the stochastic limitations of single-shot generation, we deploy an evolutionary agent that iteratively refines code for both correctness and performance. We model this process as a directed population-based search in the discrete space of possible programs. See~\cref{fig:parevo_framework} for a diagram of the workflow.

\begin{figure*}[t]
    \centering
    \scalebox{0.75}{
    \begin{tikzpicture}[
        node distance=1.5cm and 2cm,
        box/.style={rectangle, draw, rounded corners, minimum width=2.5cm, minimum height=1cm, align=center, fill=blue!5},
        agent/.style={rectangle, draw, rounded corners, minimum width=3cm, minimum height=1.5cm, align=center, fill=orange!10, thick},
        eval/.style={rectangle, draw, rounded corners, minimum width=2.5cm, minimum height=1cm, align=center, fill=red!5},
        arrow/.style={-latex, thick}
    ]
        \node[box] (context) {Human Expert Context\\(Problem, Tooling)};
        \node[agent, right=of context] (llm) {\textbf{Evolutionary}\\ \textbf{LLM Agent (ECA)}};
        \node[box, right=of llm] (candidates) {Candidate\\Parallel Algorithms};
        
        \node[eval, below=of candidates] (eval_correct) {Correctness\\Verification};
        \node[eval, left=of eval_correct] (eval_tsan) {Stress\\Testing};
        \node[eval, left=of eval_tsan] (eval_perf) {Performance\\Profiling};
        
        \node[box, above=of llm] (mapelites) {MAP-Elites\\Selection};
        \node[box, right=of candidates] (output) {Optimized\\Algorithm};
        
        \begin{scope}[on background layer]
            \node[draw, dashed, fill=gray!5, inner sep=10pt, fit=(eval_correct) (eval_tsan) (eval_perf)] (eval_box) {};
            \node[anchor=south east] at (eval_box.south east) {\small \textit{Evaluation Framework}};
        \end{scope}

        \draw[arrow] (context) -- (llm);
        \draw[arrow] (llm) -- (candidates);
        \draw[arrow] (candidates) -- (eval_correct);
        \draw[arrow] (eval_correct) -- (eval_tsan);
        \draw[arrow] (eval_tsan) -- (eval_perf);
        
        \draw[arrow] (eval_perf) -- node[left, align=center, font=\small] {Metrics \& \\ Diagnostics} (llm);
        
        \draw[arrow] (eval_perf.west) -| ([xshift=-1.5cm]context.west) |- (mapelites.west);
        \draw[arrow] (llm.north) -- (mapelites.south);
        \draw[arrow] (mapelites.east) -| (candidates.north);
        
        \draw[arrow] (candidates) -- (output);
    \end{tikzpicture}
    }
    \vspace{-0.5em}
    \caption{\textbf{Overview of the ParEVO Framework.} The system integrates human expert context (problem formulation, parallel tooling) with an evolutionary LLM agent. The cycle iteratively refines candidate parallel algorithms through a rigorous evaluation framework (correctness verification, stress testing, and performance profiling), using metrics to guide the selection of the next population via MAP-Elites.}
    \label{fig:parevo_framework}
\end{figure*}

The agent maintains a diverse population of candidate solutions, each associated with specific performance metrics (test coverage, execution time) and diagnostic artifacts (compiler logs, failure reasons, and targeted refinement instructions). The search initializes with either a baseline functional solution or a raw problem description. We define the fitness function $f(x)$ for a candidate solution $x$ as:
\begin{equation}
    f(x) = \begin{cases} 
    0 & \text{if } x \text{ fails compilation or tests} \\
    \frac{1}{T(x) + \varepsilon} & \text{if } x \text{ passes, where } T(x) \text{ is runtime}
    \end{cases}
\end{equation}
Furthermore, candidate solutions that trigger execution timeouts or yield inconsistent outputs across the five stress-test rounds are strictly assigned a fitness of 0.

A critical design choice in our evolutionary loop is  reliance on deterministic external tools---specifically compilers and stress tests---rather than LLM-based static analysis. Because LLMs process code as text tokens, they natively fail to capture inter-thread timing and synchronization structures, making them prone to hallucinating data races. Furthermore, \citet{singh2024testtime} finds LLMs to be fundamentally unreliable verifiers of low-level parallel code. By executing each candidate five times on exceedingly large inputs (on the order of 10 million vertices), we force latent concurrency bugs to manifest as observable failures, providing a rigorous empirical filter that penalizes unsafe memory accesses.

In each generation, the agent selects survivors to populate the context window for the next iteration. To balance exploitation and exploration, we select the top $k=3$ solutions by fitness (performance) and $d=5$ diverse solutions via MAP-Elites. The MAP-Elites archive is organized as a grid indexed by two behavioral dimensions:
\begin{itemize}
    \item \textbf{Complexity:} The character length of the source code, discretized into bins. Shorter programs fall into lower bins, helping retain solutions of varying structural complexity.
    \item \textbf{Diversity:} The mean character-level edit (Levenshtein) distance between a candidate and all other programs currently in the population. Programs structurally dissimilar to the rest of the population receive high diversity scores and occupy higher bins.
\end{itemize}
A new candidate replaces an incumbent in a grid cell only if its fitness is strictly higher, enforcing a strict quality-diversity invariant. These selected candidates, along with their diagnostic artifacts, prompt the LLM to synthesize the next generation of improved code. The process terminates by returning the candidate with the maximum fitness score.

\subsection{Supported Languages}
To demonstrate the versatility of our approach, in this paper, we use our \sys on two languages: C++ and Rust. For C++, we use our \sys system to fine-tune models on ParlayLib~\cite{parlaylib}. For Rust, we use our \sys system to fine-tune models on RPB: Rust Parallel Benchmarks Suite~\cite{abdi2023rustypbbs,rpb_repo}. For both C++ and Rust, our methods lead to improved performance.

\subsection{Benchmarking Suite}
We evaluate our framework across four distinct benchmarks to assess both generation quality and runtime performance. First, we compare our fine-tuned models against state-of-the-art local and commercial LLMs using the \textbf{ParEval}~\cite{nichols2024pareval} library. Second, we measure absolute performance against expert human baselines, utilizing C++ solutions from \textbf{PBBSBench}~\cite{shun2012pbbs} and Rust implementations from \textbf{RPB}~\cite{rpb_repo}. Finally, to test generalization, we evaluate on a held-out set of \textbf{DMOJ} competitive programming problems. In this setting, we compare the runtime of code generated by \sys against official contest solutions, demonstrating significant speedups.





\section{Experimental Results}

\subsection{Experimental Setup}
\textbf{Hardware.} All experiments were conducted on a dual-socket compute node equipped with two Intel Xeon Platinum 8562Y+ processors (64 physical cores total) and 512GB DDR5 ECC RAM. An NVIDIA H200 GPU was utilized solely for inference.

\textbf{Benchmarks.} We evaluated on:
\begin{enumerate}
    \item \textbf{ParEval:} The testing suite of~\cite{nichols2024pareval}.
    \item \textbf{PBBSBench \& RPB:} Expert-written C++ and Rust baselines~\cite{shun2012pbbs,rpb_repo}.
    \item \textbf{DMOJ:} A held-out set of competitive programming problems.~\cite{dmoj}
\end{enumerate}

\subsection{Main Results: ParEval Performance}
\textbf{Methodological Note on Expected Speedup.} In traditional systems literature, the geometric mean is typically used to average normalized execution times of a static benchmark suite across different hardware. However, in the context of zero-shot code generation over a large distribution of tasks (ParEval), we conceptualize performance formally as an expected capability reward. Specifically, we report the arithmetic mean of \texttt{Speedup@1} to represent the \emph{expected speedup} ($\mathbb{E}[S]$) a user would experience when querying the model with a random task from the problem domain. This aligns directly with standard machine learning evaluation practices for reporting expected test-time rewards over a distribution, as opposed to summarizing the total execution time of a fixed static workload.

\cref{tab:pareval-results} presents the performance of local and commercial models. Our fine-tuned models (\textbf{Gemini-2.5-Parlay} and \textbf{DeepSeek-Parlay}) significantly outperform their base counterparts. Notably, \textbf{Gemini-2.5-Parlay} achieves an average $107\times$ speedup over the baseline provided by ParEval, driven by its ability to generate valid, compilable parallel code (\texttt{Build@1} 0.81 vs 0.25 of the state-of-the-art \texttt{Gemini 3.0 Pro}). Even our smallest fine-tuned model, DeepSeek-Parlay (with 6.7b parameters) is able to beat the commercial
state-of-the-art \texttt{Gemini-3-Pro}. The \verb|Speedup@1| metric is the arithmetic mean of the expected best performance speedups across all 59 problems relative to a sequential baseline. The exceptionally high 107.43x mean for Gemini-2.5-Parlay is driven by heavy-tailed performance on specific irregular tasks (e.g., \verb|34_scan_largest_contiguous_subarray_sum|), where the model discovered an optimal prefix-sum algorithm while the baseline sequential implementation is heavily bottlenecked by nested loops.

\begin{table*}[ht]
\centering
\definecolor{parlaygreen}{rgb}{0.9, 1.0, 0.9}
\definecolor{rustpurple}{rgb}{0.95, 0.9, 1.0}
\resizebox{\textwidth}{!}{
\begin{tabular}{lllcccc}
\toprule
\textbf{Execution Model} & \textbf{Code} & \textbf{Sched.}  & \textbf{Build@1} & \textbf{Pass@1} & \textbf{Speedup@1 (AM)} & \textbf{Speedup@1 (GM)} \\
\midrule
\rowcolor{parlaygreen} Claude Opus 4.5 & Parlay & Parlay  & 0.97 & 0.05 & 0.42 & 1.13 \\
\rowcolor{parlaygreen} GPT-5 Thinking & Parlay & Parlay  & 0.93 & 0.05 & 0.43 & 1.13 \\
\rowcolor{parlaygreen} Gemini-2.5-Flash & Parlay & Parlay & 0.57 & 0.29 & 6.88 & 2.75 \\
\rowcolor{parlaygreen} Gemini-2.5-Pro & Parlay & Parlay  & 0.84 & 0.54 & 85.34 & 5.47 \\
\rowcolor{parlaygreen} Gemini-3-Pro & Parlay & Parlay  & 0.25 & 0.23 & 3.68 &1.53\\
\rowcolor{parlaygreen} \textbf{Gemini-2.5-Parlay} & \textbf{Parlay} & \textbf{Parlay} & \textbf{0.81} & \textbf{0.58} & \textbf{107.43} & \textbf{6.33}\\
\rowcolor{parlaygreen} DeepSeek-6.7B-Base & Parlay & Parlay & 0.89 & 0.11 & 0.73 &1.14\\
\rowcolor{parlaygreen} DeepSeek-Syntax & Parlay & Parlay & 0.85 & 0.12 & 2.42 &1.38 \\
\rowcolor{parlaygreen} \textbf{DeepSeek-Parlay} & \textbf{Parlay} & \textbf{Parlay} & \textbf{0.81} & \textbf{0.26} & \textbf{126.16} & \textbf{3.14} \\
\rowcolor{parlaygreen} \textbf{Qwen3-Parlay} & \textbf{Parlay} & \textbf{Parlay} & \textbf{0.50} & \textbf{0.33} & \textbf{5.01} & \textbf{1.87} \\
\rowcolor{parlaygreen} DeepSeek-Coder-V2-Lite-Base & Parlay & Parlay & 0.80 & 0.09 & 1.30 &1.19 \\
\rowcolor{parlaygreen} Qwen2.5-Coder-32B & Parlay & Parlay & 0.93 & 0.11 & 5.02 & 1.82 \\
\rowcolor{parlaygreen} Qwen2.5-Coder-32B-Instruct & Parlay & Parlay & 0.61 & 0.41 & 7.00 & 2.22\\
\midrule
\rowcolor{rustpurple} DeepSeek-Coder-V2-Lite-Base & Rust & Rayon  & 0.73 & 0.29 & 3.47  & 0.58\\
\rowcolor{rustpurple} DeepSeek-Coder-V2-Lite-Instruct & Rust & Rayon  & 0.40&	0.02&	0.15 & 1.04 \\
\rowcolor{rustpurple} Qwen2.5-Coder-32B & Rust & Rayon & 0.82 & 0.45 & 4.43 & 0.63 \\
\rowcolor{rustpurple} Qwen2.5-Coder-32B-Instruct & Rust & Rayon & 0.63 &	0.49	&  4.61 & 0.57 \\
\rowcolor{rustpurple} Qwen3-Coder-30B-Instruct & Rust & Rayon & 0.61&	0.50	& 4.57 & 0.46 \\
\rowcolor{rustpurple} \textbf{Qwen3-Rust} & \textbf{Rust} & \textbf{Rayon} & \textbf{0.64} & \textbf{0.46} & \textbf{4.62} &\
\textbf{0.35} \\
\rowcolor{rustpurple} StarCoder2-15B & Rust & Rayon & 0.77 & 0.27  & 2.46  & 0.79\\
\rowcolor{rustpurple} Gemini-3-Pro & Rust & Rayon & 0.97	& 0.82 &	7.32 & 0.69\\

\bottomrule
\end{tabular}
}
\caption{ParEval results (temperature $= 0.2$). \textbf{Code} denotes the parallel programming language used, and \textbf{Scheduler} the parallel runtime. \emph{Parlay} code uses the ParlayLib library, with either ParlayLib's internal scheduler or OpenMP as the scheduling backend. \emph{Rust} code uses Rayon's dynamic work-stealing scheduler. Our fine-tuned models achieve orders-of-magnitude improvements in speedup. }
\label{tab:pareval-results}
\end{table*}

\paragraph{Impact of Fine-tuning on Code Quality.}
As illustrated in \cref{fig:pareval-ft-gemini-code-contest}, the most consistent effect of fine-tuning is on runtime performance: \sys produces faster code in eleven of twelve categories, frequently by an order of magnitude, indicating that the model learns to select more efficient parallel patterns by using ParlayLib primitives rather than merely valid ones. The effect on \texttt{Build@1} and \texttt{Pass@1} is category-dependent. Fine-tuning yields its largest correctness gains precisely where the base model was weakest at expressing the appropriate ParlayLib idiom—most notably \texttt{graph} ($\texttt{Build@1}:0.62 \rightarrow 0.97$, $\texttt{Pass@1}:0.42 \rightarrow 0.76)$ and \texttt{histogram} ($\texttt{Pass@1}:0.19 \rightarrow 0.63$), while incurring mild regressions on categories the base already handled well (e.g., \texttt{fft}, \texttt{geometry}, \texttt{search}). 

\subsection{Semantic Alignment via Fine-Tuning}
A critical advantage of \sys is its ability to learn the correct semantics of parallel primitives. In the complex number sorting task (\cref{fig:code_comparison} in Appendix), the base model failed completely (Build@1 = 0), struggling with C++ custom comparators. The fine-tuned model not only compiled (Build@1 = 1) but achieved a speedup of $17.5\times$. This suggests that the model has learned to navigate the complex type system of ParlayLib.

\subsection{Performance Analysis: Strong Scaling}
\begin{figure}[ht]
    \centering
    \begin{subfigure}[b]{0.48\columnwidth}
        \includegraphics[width=\linewidth]{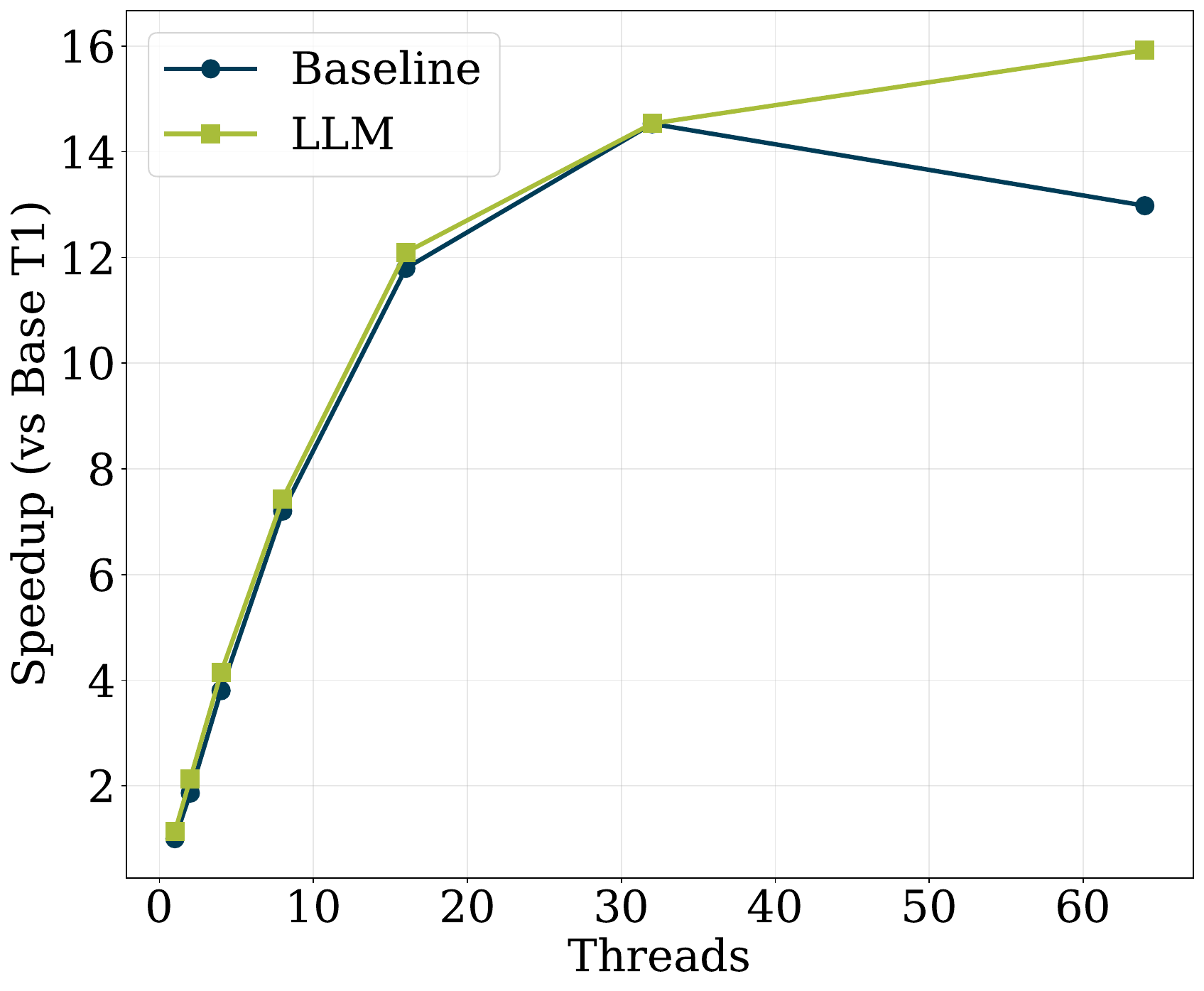}
        \caption{Maximal Matching (Rust)}
    \end{subfigure}
    \hfill
    \begin{subfigure}[b]{0.48\columnwidth}
        \includegraphics[width=\linewidth]{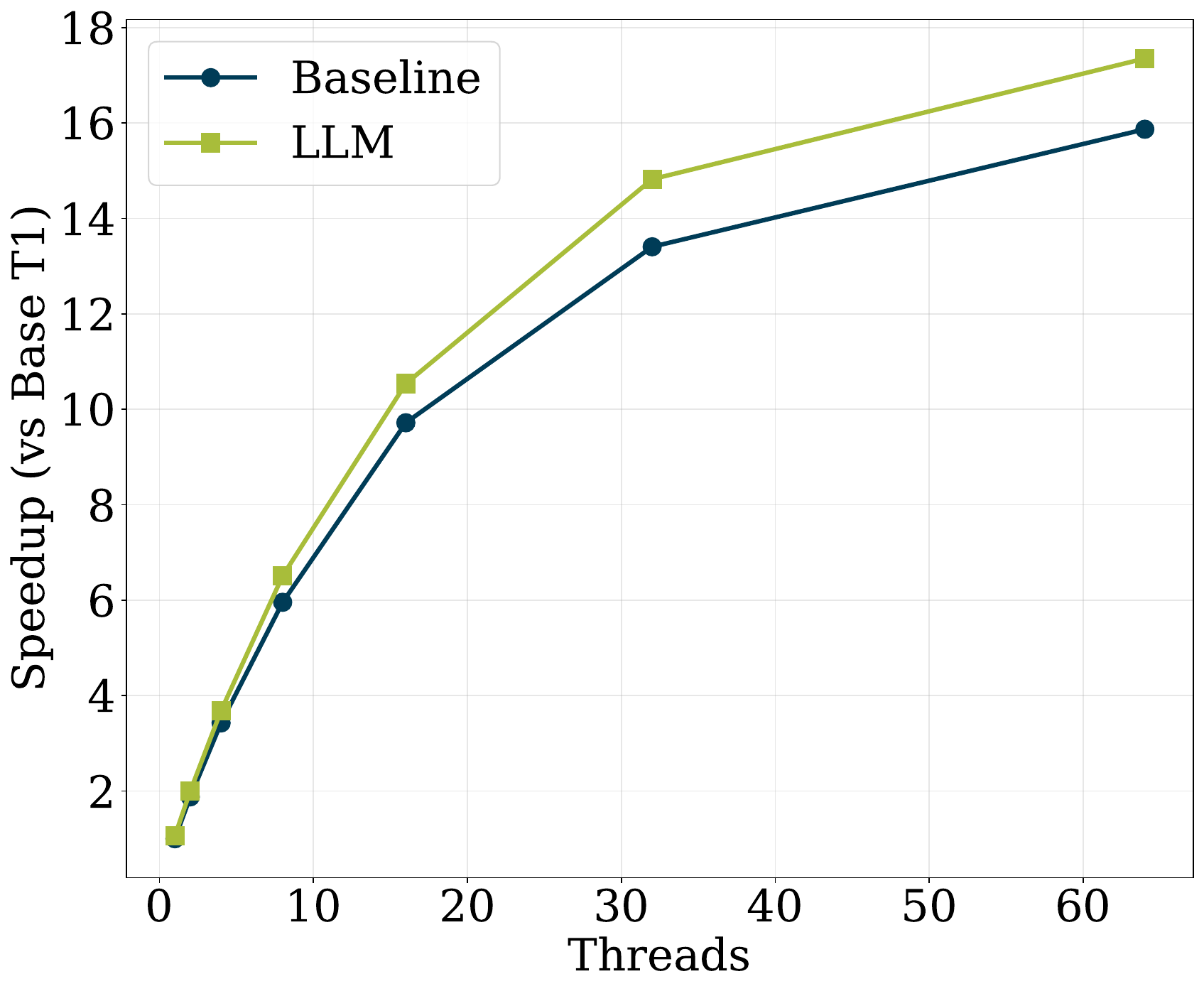}
        \caption{Min. Spanning Forest (Rust)}
    \end{subfigure}
    \begin{subfigure}[b]{0.48\columnwidth}
        \includegraphics[width=\linewidth]{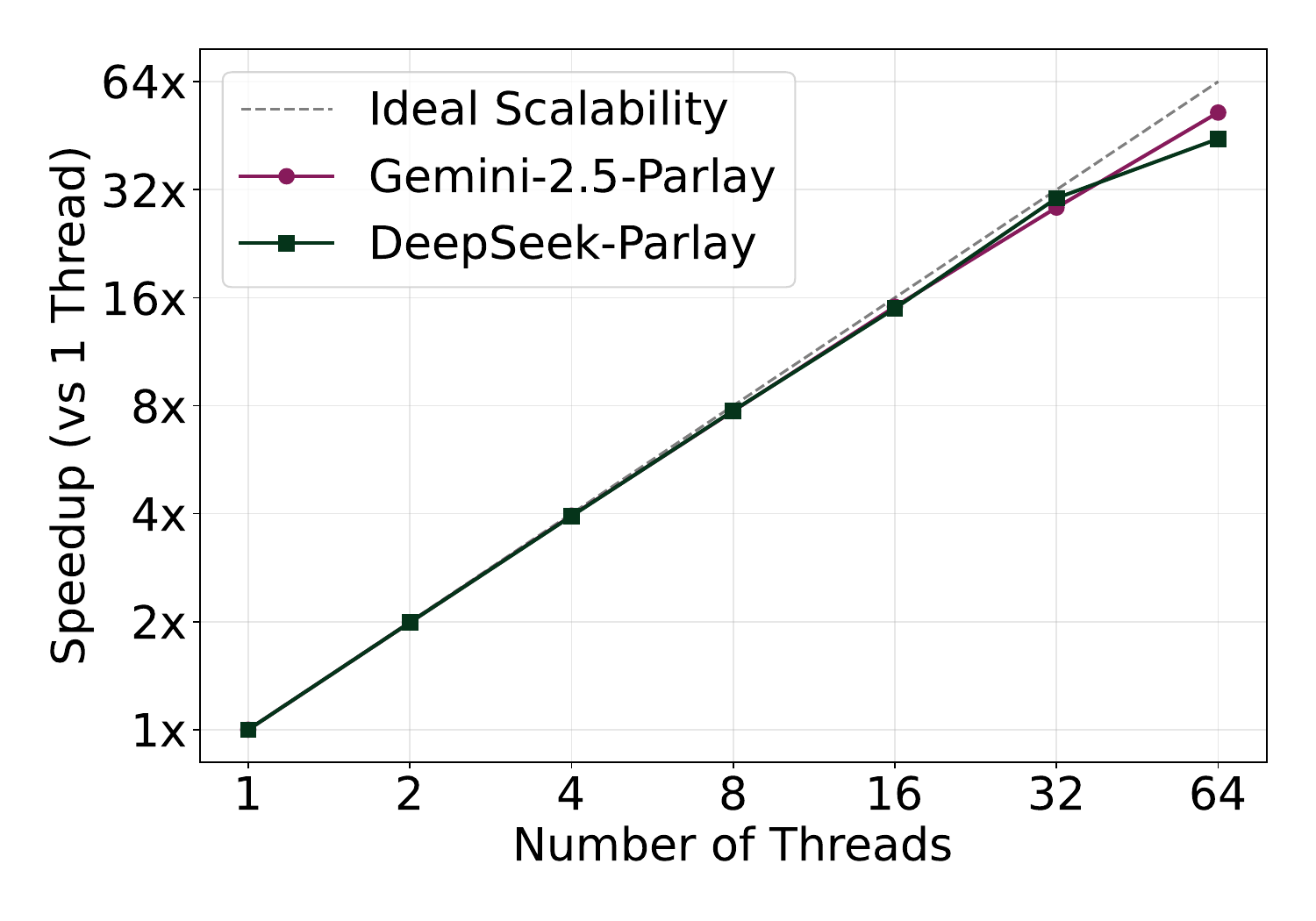}
        \caption{FFT DFT Scaling (C++)}
    \end{subfigure}
    \hfill
    \begin{subfigure}[b]{0.48\columnwidth}
        \includegraphics[width=\linewidth]{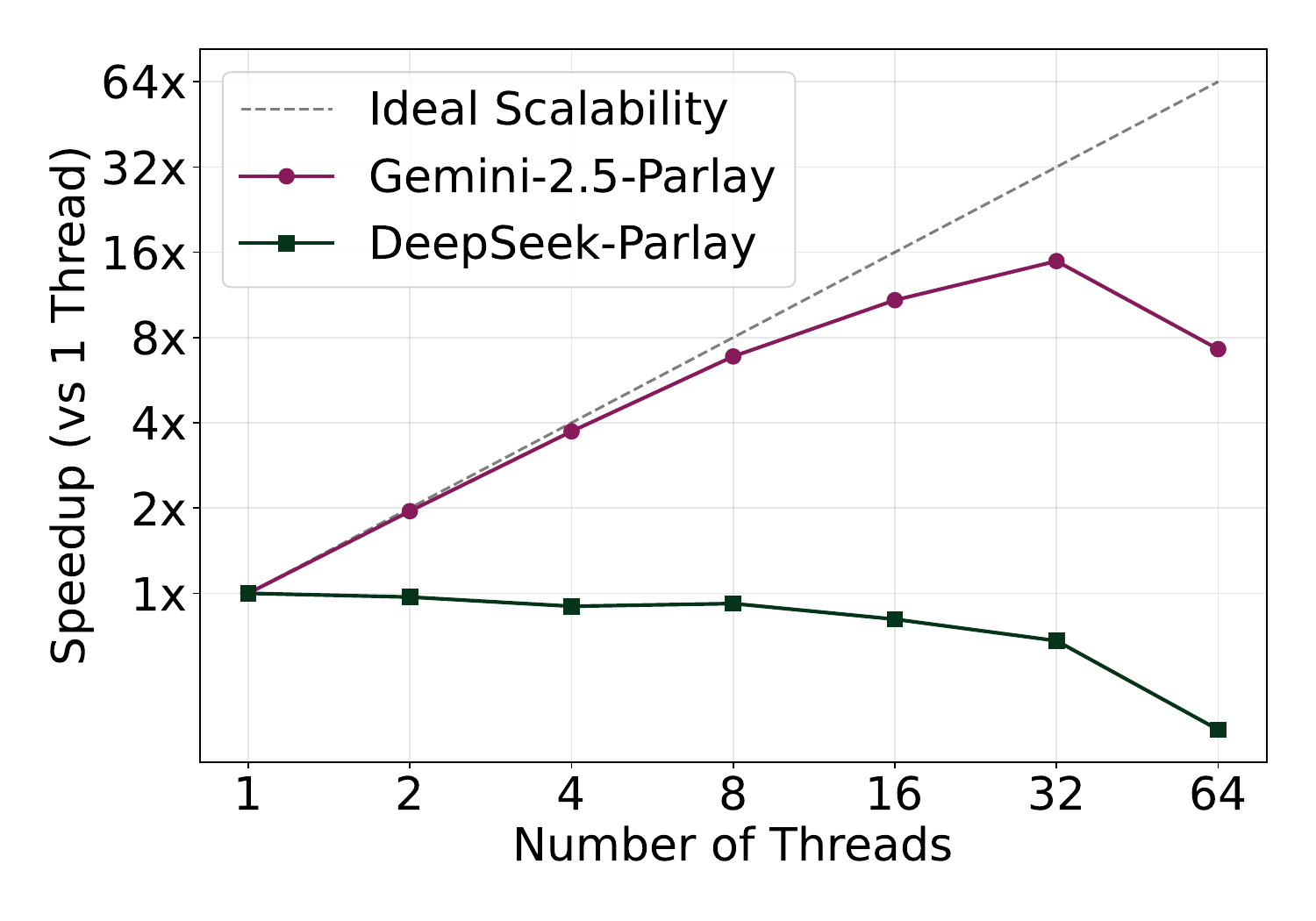}
        \caption{Largest Component (C++)}
    \end{subfigure}
    
    \caption{Strong Scaling results. (c) Algorithms like Discrete Fourier Transform show excellent scaling with \sys's generated code, reaching nearly $40\times$ speedup on 64 cores.}
    \label{fig:scaling}
\end{figure}

Code correctness is insufficient for HPC; the solution must also scale. \cref{fig:scaling} demonstrates strong scaling up to 64 cores. For regular parallelism like Discrete Fourier Transform, our model generates code that scales near-linearly ($40\times$ speedup), abstracting away complex synchronization that typically hinders manual implementations. The performance drop observed at 64 cores for \cref{fig:scaling}(d) is due to parallel overhead and thread contention, where the generated code reveals that the LLM attempted to maintain local dynamic queues inside a nested parallel BFS loop. 

\subsection{Comparison vs. Expert Baselines}
We benchmarked our generated solutions against expert human implementations from PBBSBench (C++) and RPB (Rust). As shown in \cref{tab:runtime_graph_comparison}, \sys matches or exceeds expert performance. For \textbf{Maximal Independent Set}, the generated Rust solution achieved a $4.1\times$ speedup over the baseline by identifying a superior parallel strategy. We demonstrate the maximum speedups we can
gain by using \texttt{Gemini-3-Pro} with our ParEVO evolutionary strategy described in~\cref{sec:eca}. 

\begin{table*}[t]
\definecolor{parlaygreen}{rgb}{0.9, 1.0, 0.9}
\definecolor{rustpurple}{rgb}{0.95, 0.9, 1.0}
\caption{Runtime Comparison: PBBS \& RPB at Thread=32, best speedup across test inputs. Baseline code is state-of-the-art human-written code. We also demonstrate the speedup against one thread, labeled Speedup (1T).}
\label{tab:runtime_graph_comparison}
\centering
\scriptsize
\begin{sc}
\resizebox{\textwidth}{!}{
    \begin{tabular}{llcccc}
    \toprule
    \textbf{Problem} & \textbf{Model/Method} & \textbf{Language} & \textbf{Runtime (s)} & \textbf{Speedup (1T)} & \textbf{Speedup (Base)} \\
    \midrule
    \rowcolor{rustpurple} Maximal Independent Set & Baseline & Rust & 0.31876 & $1.116\times$ & -- \\
    \rowcolor{rustpurple} Maximal Independent Set & PAREVO (GEMINI)  & Rust & 0.07728 & $0.938\times$ & $4.125\times$ \\
    \rowcolor{rustpurple} Maximal Matching & Baseline & Rust & 0.20646 & $21.723\times$ & -- \\
    \rowcolor{rustpurple} Maximal Matching & PAREVO (GEMINI)  & Rust & 0.1928 & $21.43286835\times$ & $1.0708\times$ \\
    \rowcolor{rustpurple} Minimum Spanning Forest & Baseline & Rust & 0.41968 & $13.427\times$ & -- \\
    \rowcolor{rustpurple} Minimum Spanning Forest  & PAREVO (GEMINI)  & Rust & 0.38004 & $13.87\times$ & $1.1043\times$ \\
    \rowcolor{rustpurple} Spanning Forest & Baseline & Rust & 0.11571 & $10.378\times$ & -- \\
    \rowcolor{rustpurple} Spanning Forest & PAREVO (GEMINI)  & Rust & 0.08865 & $15.482\times$ & $1.3052\times$ \\
    \midrule
    \rowcolor{parlaygreen} Minimum Spanning Forest & Baseline & C++ & 1.24 & $22.633\times$ & -- \\
    \rowcolor{parlaygreen} Minimum Spanning Forest  & PAREVO (GEMINI)  & C++ & 1.169 & $23.689\times$ & $1.061\times$ \\
    \rowcolor{parlaygreen} Histogram  & Baseline & C++ & $0.051$ & $27.59\times$ & -- \\
    \rowcolor{parlaygreen} Histogram  & PAREVO (GEMINI)  & C++ & $0.019$ & $>13.94\times$
    & $2.68421\times$ \\


    \bottomrule
    \end{tabular}
}
\end{sc}
\end{table*}


\subsection{Ablation Study: Evolutionary Agent}
To isolate the contribution of the Evolutionary Coding Agent (ECA), we evaluated performance with the agent disabled. \cref{tab:ablation} confirms that iterative refinement is crucial: 30 iterations of ECA yield a $2.2\times$ performance multiplier over single-shot generation. 

For this ablation study, we utilized a reserved set of training problems sourced from DMOJ.\footnote{The DMOJ training problems are available at \url{https://github.com/WildAlg/ParEVO/tree/main/code-contests-dataset}} In \cref{tab:ablation}, the base performance ($1.00\times$) corresponds to \texttt{baseline.cpp}, which we define as the very first solution that passes all tests (this is not necessarily the solution from iteration 1 if the early attempts fail). The reported speedup for the ECA configurations is calculated as the average of the relative speedups achieved across all datasets within this training corpus.

\begin{table}[h]
\caption{Impact of Evolutionary Refinement on Speedup. 30 iterations of ECA yield a $2.2\times$ speedup over the first valid solution.}
\label{tab:ablation}
\begin{center}
\begin{small}
\begin{sc}
\resizebox{\columnwidth}{!}{
    \begin{tabular}{lcc}
    \toprule
    \textbf{Configuration} & \textbf{Speedup} \\
    \midrule
    Gemini-3-Pro (No ECA) &  $1.00\times$ (baseline) \\
    Gemini-3-Pro + ECA (10 iter) &  $1.498\times$ \\
    Gemini-3-Pro + ECA (30 iter) & $\mathbf{2.218\times}$ \\
    \bottomrule
    \end{tabular}
}
\end{sc}
\end{small}
\end{center}
\vskip -0.1in
\end{table}

While the actual prompt in each iteration contains more context (other iterations/metrics) by the default template of \texttt{openevolve}, the structural system prompt we specified for the ECA is provided in \cref{sec:eca-prompt}.

\subsection{Analysis: The Correctness-Speedup Trade-off}
A deeper analysis of Graph problems (\cref{tab:graph_metrics}) reveals a trade-off. Fine-tuning increases correctness (\texttt{Pass@1} $0.42 \to 0.76$) by enforcing safe API usage, but sometimes degrades speedup ($21\times \rightarrow 13\times$) as it favors stable, high-level primitives over risky, fine-grained atomic operations.

\begin{table}[h]
    \centering
    \small
    \caption{Performance on Graph Problems. Fine-tuning improves reliability (Pass@1) but favors safer, slightly slower algorithms.}
    \label{tab:graph_metrics}
    \begin{tabular}{l c c c}
        \toprule
        \textbf{Model} & \textbf{Build@1} & \textbf{Pass@1} & \textbf{Speedup@1} \\
        \midrule
        Gemini-2.5-Pro & 0.62 & 0.42 & \textbf{21.76} \\
        Gemini-2.5-Parlay & \textbf{0.97} & \textbf{0.76} & 13.67 \\
        \bottomrule
    \end{tabular}
\end{table}

\section{Discussion and Limitations}

\subsection{The Role of Abstraction in Parallelization}
Our findings suggest that the efficacy of LLM parallel code generation depends heavily on the abstraction level provided by the target intermediate representation (IR). We argue that the superior performance of \sys on ParlayLib stems from an \textit{alignment of abstraction}. Imperative models like OpenMP force the LLM to manage global state and explicit synchronization: tasks that maximize the ``state-tracking'' burden on the attention mechanism and increase the probability of race conditions. 

In contrast, ParlayLib functions as a high-level parallel DSL. Its functional primitives (e.g., \texttt{map}, \texttt{reduce}, \texttt{scan}) encapsulate complex scheduling logic and enforce immutability. This reduces parallelization to \textit{local} transformations (mapping serial loops to equivalent functional constructs) which aligns naturally with the token-local prediction of Transformer models. 

By training our models to target ParlayLib's composable primitives, ParEVO aligns the optimization objective with the token-local reasoning capabilities of the Transformer architecture, yielding code that is both mathematically sound and highly performant.

\subsection{Limitations and Future Directions}
\begin{itemize}
    \item \textbf{Architectural Scope:} \sys is currently optimized for shared-memory multicore architectures. It does not address the distributed memory paradigm (e.g. MPI/PGAS), where communication latency and data partitioning introduce distinct optimization constraints.
    \item \textbf{Inference Latency vs. Runtime Efficiency:} The Evolutionary Coding Agent trades inference-time compute for execution-time speedup. While generating multiple candidates and compiling them is costly, we argue this is an acceptable amortized cost for HPC kernels that may run trillions of times over their lifecycle.
    \item \textbf{Domain Generalization:} On some benchmarks, the model produces ``confident hallucinations'' when applying learned parallel patterns to unfamiliar algorithmic domains. Future work will explore integrating formal verification tools into the evolutionary loop to constrain these semantic errors.
\end{itemize}

\section{Conclusion}

We have presented \textbf{\sys}, a framework that bridging modern generative AI and high-performance computing. By curating a specialized dataset of parallel primitives and fine-tuning models to internalize the \textbf{Work-Depth} cost model, we achieve state-of-the-art results on the ParEval, surpassing both commercial LLMs and traditional heuristics.

Crucially, our results reveal that syntax generation alone is insufficient for HPC. The integration of an \textbf{Evolutionary Coding Agent}—which treats the compiler and runtime profiler as adversarial critics—is essential for traversing the optimization landscape. This work sets a precedent for \textbf{AI-Driven Performance Engineering}: moving beyond code completion to systems that actively reason about scalability, correctness, and the complex interplay between algorithms and hardware.


\section*{Acknowledgements}
We thank Lin Zhong for helpful discussions and Rust resources, Ramla Ijaz for helpful discussions, and Roger Fu for compiling and providing to us the publicly available test cases for the competitive programming problems we used.
We also thank the extended team at Google DeepMind who supported this research direction.
Amir Yazdanbakhsh and Deniz Altınbüken contributed to this paper in an advisory capacity.

This work was supported in part by the National Science Foundation (NSF) under Grant \#CCF-2453323 and a Google Academic Research Award.

\section*{Impact Statement}
This paper presents work whose goal is to advance the field of Machine Learning and High Performance Computing. By enabling easier access to efficient parallel programming, this work could reduce the energy footprint of large-scale computations as well as make parallel computing accessible to non-experts. However, as with all code generation tools, there is a risk of generating subtle bugs in critical systems if not properly verified. We recommend human oversight for mission-critical applications.

\bibliography{references}

@misc{shypula2024learningperformanceimprovingcodeedits,
      title={Learning Performance-Improving Code Edits}, 
      author={Alexander Shypula and Aman Madaan and Yimeng Zeng and Uri Alon and Jacob Gardner and Milad Hashemi and Graham Neubig and Parthasarathy Ranganathan and Osbert Bastani and Amir Yazdanbakhsh},
      year={2024},
      eprint={2302.07867},
      archivePrefix={arXiv},
      primaryClass={cs.SE},
      url={https://arxiv.org/abs/2302.07867}, 
}

@string{spaa = "{ACM} Symposium on Parallelism in Algorithms
                and Architectures (SPAA)"}

@inbook{Chen_2024,
   title={OMPGPT: A Generative Pre-trained Transformer Model for OpenMP},
   ISBN={9783031695773},
   ISSN={1611-3349},
   url={http://dx.doi.org/10.1007/978-3-031-69577-3_9},
   DOI={10.1007/978-3-031-69577-3_9},
   booktitle={Euro-Par 2024: Parallel Processing},
   publisher={Springer Nature Switzerland},
   author={Chen, Le and Bhattacharjee, Arijit and Ahmed, Nesreen and Hasabnis, Niranjan and Oren, Gal and Vo, Vy and Jannesari, Ali},
   year={2024},
   pages={121–134} }

@article{HUSEIN2025103917,
title = {Large language models for code completion: A systematic literature review},
journal = {Computer Standards \& Interfaces},
volume = {92},
pages = {103917},
year = {2025},
issn = {0920-5489},
doi = {https://doi.org/10.1016/j.csi.2024.103917},
url = {https://www.sciencedirect.com/science/article/pii/S0920548924000862},
author = {Rasha Ahmad Husein and Hala Aburajouh and Cagatay Catal}
}

@misc{eniser2024translatingrealworldcodellms,
      title={Towards Translating Real-World Code with LLMs: A Study of Translating to Rust}, 
      author={Hasan Ferit Eniser and Hanliang Zhang and Cristina David and Meng Wang and Maria Christakis and Brandon Paulsen and Joey Dodds and Daniel Kroening},
      year={2024},
      eprint={2405.11514},
      archivePrefix={arXiv},
      primaryClass={cs.SE},
      url={https://arxiv.org/abs/2405.11514}, 
}

@misc{ahmed2022learningcodesummarizationsmall,
      title={Learning code summarization from a small and local dataset}, 
      author={Toufique Ahmed and Premkumar Devanbu},
      year={2022},
      eprint={2206.00804},
      archivePrefix={arXiv},
      primaryClass={cs.SE},
      url={https://arxiv.org/abs/2206.00804}, 
}

@misc{kambhampati2024llmscantplanhelp,
      title={LLMs Can't Plan, But Can Help Planning in LLM-Modulo Frameworks}, 
      author={Subbarao Kambhampati and Karthik Valmeekam and Lin Guan and Mudit Verma and Kaya Stechly and Siddhant Bhambri and Lucas Saldyt and Anil Murthy},
      year={2024},
      eprint={2402.01817},
      archivePrefix={arXiv},
      primaryClass={cs.AI},
      url={https://arxiv.org/abs/2402.01817}, 
}

@article{Sahu_2019,
   title={The ubiquity of large graphs and surprising challenges of graph processing: extended survey},
   volume={29},
   ISSN={0949-877X},
   url={http://dx.doi.org/10.1007/s00778-019-00548-x},
   DOI={10.1007/s00778-019-00548-x},
   number={2–3},
   journal={The VLDB Journal},
   publisher={Springer Science and Business Media LLC},
   author={Sahu, Siddhartha and Mhedhbi, Amine and Salihoglu, Semih and Lin, Jimmy and Özsu, M. Tamer},
   year={2019},
   month=jun, pages={595–618} }

@inproceedings{social_graph,
author = {Bronson, Nathan and Amsden, Zach and Cabrera, George and Chakka, Prasad and Dimov, Peter and Ding, Hui and Ferris, Jack and Giardullo, Anthony and Kulkarni, Sachin and Li, Harry and Marchukov, Mark and Petrov, Dmitri and Puzar, Lovro and Song, Yee Jiun and Venkataramani, Venkat},
title = {TAO: Facebook's distributed data store for the social graph},
year = {2013},
publisher = {USENIX Association},
address = {USA},
booktitle = {Proceedings of the 2013 USENIX Conference on Annual Technical Conference},
pages = {49–60},
numpages = {12},
location = {San Jose, CA},
series = {USENIX ATC'13}
}

@inproceedings{parlaylib,
author = {Blelloch, Guy E. and Anderson, Daniel and Dhulipala, Laxman},
title = {{ParlayLib: A Toolkit for Parallel Algorithms on Shared-Memory Multicore Machines}},
year = {2020},
isbn = {9781450369350},
publisher = {Association for Computing Machinery},
address = {New York, NY, USA},
url = {https://doi.org/10.1145/3350755.3400254},
doi = {10.1145/3350755.3400254},
booktitle = {Proceedings of the 32nd ACM Symposium on Parallelism in Algorithms and Architectures},
pages = {507–509},
numpages = {3},
location = {Virtual Event, USA},
series = {SPAA '20}
}

@inproceedings{mahmud2023autoparllmgnnguidedautomaticcode,
    title = "{A}uto{P}ar{LLM}: {GNN}-guided Context Generation for Zero-Shot Code Parallelization using {LLM}s",
    author = "Mahmud, Quazi Ishtiaque  and
      TehraniJamsaz, Ali  and
      Phan, Hung D  and
      Chen, Le  and
      Capot{\u{a}}, Mihai  and
      Willke, Theodore L.  and
      Ahmed, Nesreen K.  and
      Jannesari, Ali",
    editor = "Chiruzzo, Luis  and
      Ritter, Alan  and
      Wang, Lu",
    booktitle = "Proceedings of the 2025 Conference of the Nations of the Americas Chapter of the Association for Computational Linguistics: Human Language Technologies (Volume 1: Long Papers)",
    month = apr,
    year = "2025",
    address = "Albuquerque, New Mexico",
    publisher = "Association for Computational Linguistics",
    url = "https://aclanthology.org/2025.naacl-long.593/",
    doi = "10.18653/v1/2025.naacl-long.593",
    pages = "11821--11841",
    ISBN = "979-8-89176-189-6",
}

@inproceedings{shinn2023reflexion,
  title={Reflexion: Language Agents with Verbal Reinforcement Learning},
  author={Shinn, Noah and Cassano, Federico and Gopinath, Ashwin and Narasimhan, Karthik and Yao, Shunyu},
  booktitle={Advances in Neural Information Processing Systems (NeurIPS)},
  volume={36},
  year={2023}
}

@inproceedings{tehranijamsaz2024coderosetta,
author = {TehraniJamsaz, Ali and Bhattacharjee, Arijit and Chen, Le and Ahmed, Nesreen K. and Yazdanbakhsh, Amir and Jannesari, Ali},
title = {CODEROSETTA: pushing the boundaries of unsupervised code translation for parallel programming},
year = {2024},
isbn = {9798331314385},
publisher = {Curran Associates Inc.},
address = {Red Hook, NY, USA},
booktitle = {Proceedings of the 38th International Conference on Neural Information Processing Systems},
articleno = {3202},
numpages = {35},
location = {Vancouver, BC, Canada},
series = {NIPS '24}
}

@inproceedings{nichols2024pareval,
  author    = {Daniel Nichols and Joshua H. Davis and Zhaojun Xie and Arjun Rajaram and Abhinav Bhatele},
  title     = {Can Large Language Models Write Parallel Code?},
  booktitle = {Proceedings of the 33rd International Symposium on High-Performance Parallel and Distributed Computing (HPDC '24)},
  year      = {2024},
  publisher = {ACM},
  doi       = {10.1145/3625549.3658689},
  url       = {https://pssg.cs.umd.edu/assets/papers/2024-06-pareval-hpdc.pdf}
}

@inproceedings{shun2012pbbs,
  author    = {Julian Shun and Guy E. Blelloch and Aapo Kyrola and Harsha Vardhan Simhadri and Kanat Tangwongsan and Jeremy T. Fineman and Phillip B. Gibbons},
  title     = {Brief Announcement: The Problem Based Benchmark Suite},
  booktitle = {Proceedings of the 24th ACM Symposium on Parallelism in Algorithms and Architectures (SPAA '12)},
  year      = {2012},
  publisher = {ACM},
  doi       = {10.1145/2312005.2312018}
}

@inproceedings{anderson2022pbbsV2,
  author    = {Daniel Anderson and Guy E. Blelloch and Laxman Dhulipala and Magdalen Dobson and Yihan Sun},
  title     = {The problem-based benchmark suite ({PBBS}), {V2}},
  booktitle = {Proceedings of the 27th ACM SIGPLAN Symposium on Principles and Practice of Parallel Programming (PPoPP '22)},
  year      = {2022},
  publisher = {ACM},
  doi       = {10.1145/3503221.3508422}
}

@inproceedings{abdi2023rustypbbs,
  author    = {Javad Abdi and Guowei Zhang and Mark C. Jeffrey},
  title     = {Brief Announcement: Is the Problem-Based Benchmark Suite Fearless with Rust?},
  booktitle = {Proceedings of the 35th ACM Symposium on Parallelism in Algorithms and Architectures (SPAA '23)},
  year      = {2023},
  publisher = {ACM},
  doi       = {10.1145/3558481.3591313}
}

@inproceedings{hu2022lora,
  author    = {Edward J. Hu and Yelong Shen and Phillip Wallis and Zeyuan Allen-Zhu and Yuanzhi Li and Shean Wang and Lu Wang and Weizhu Chen},
  title     = {{LoRA}: Low-Rank Adaptation of Large Language Models},
  booktitle = {International Conference on Learning Representations (ICLR)},
  year      = {2022},
  publisher = {OpenReview.net},
  url       = {https://openreview.net/pdf?id=nZeVKeeFYf9}
}

@misc{ren2020codebleu,
      title={CodeBLEU: a Method for Automatic Evaluation of Code Synthesis}, 
      author={Shuo Ren and Daya Guo and Shuai Lu and Long Zhou and Shujie Liu and Duyu Tang and Neel Sundaresan and Ming Zhou and Ambrosio Blanco and Shuai Ma},
      year={2020},
      eprint={2009.10297},
      archivePrefix={arXiv},
      primaryClass={cs.SE},
      url={https://arxiv.org/abs/2009.10297}, 
}

@article{brent1974parallel,
  author  = {Richard P. Brent},
  title   = {The Parallel Evaluation of General Arithmetic Expressions},
  journal = {Journal of the ACM},
  year    = {1974},
  volume  = {21},
  number  = {2},
  pages   = {201--206}
}

@article{blumofe1999workstealing,
author = {Blumofe, Robert D. and Leiserson, Charles E.},
title = {Scheduling multithreaded computations by work stealing},
year = {1999},
issue_date = {Sept. 1999},
publisher = {Association for Computing Machinery},
address = {New York, NY, USA},
volume = {46},
number = {5},
issn = {0004-5411},
url = {https://doi.org/10.1145/324133.324234},
doi = {10.1145/324133.324234},
journal = {J. ACM},
month = sep,
pages = {720–748},
numpages = {29}
}

@inproceedings{bitan2025unipar,
  title={UniPar: A Unified LLM-Based Framework for Parallel and Accelerated Code Translation in HPC},
  author={Bitan, Tomer and Kadosh, Tal and Kaplan, Erel and Meiri, Shira and Chen, Le and Morales, Peter and Hasabnis, Niranjan and Oren, Gal},
  booktitle={2025 IEEE High Performance Extreme Computing Conference (HPEC)},
  year={2025},
  organization={IEEE}
}

@misc{dmoj,
  author       = {{DMOJ Developers}},
  title        = {{DMOJ}: Modern Online Judge},
  year         = {2024},
  howpublished = {\url{https://github.com/DMOJ/online-judge}},
  note         = {GitHub repository}
}

@misc{sharma2025openevolve,
  author = {Sharma, Asankhaya},
  title = {OpenEvolve: An Open Source Implementation of Google DeepMind's AlphaEvolve},
  year = {2025},
  publisher = {GitHub},
  journal = {GitHub repository},
  howpublished = {\url{https://github.com/codelion/openevolve}},
  note = {Accessed: 2026-01-24}
}

@misc{rpb_repo,
  author       = {Javad Abdi and Guowei Zhang and Mark C. Jeffrey},
  title        = {{Rusty-PBBS}: Rust Problem Based Benchmark Suite},
  year         = {2023},
  publisher    = {GitHub},
  howpublished = {\url{https://github.com/mcj-group/rpb}},
  note         = {GitHub repository}
}

@article{chaturvedi2024hpccoderv2,
  title={HPCCoder-v2: Efficient Fine-Tuning of Small Language Models for High-Performance Computing},
  author={Chaturvedi, Aman},
  journal={arXiv preprint arXiv:2410.20527},
  year={2024}
}

@article{rahman2025marco,
  title={MARCO: Multi-Agent Reasoning for Code Optimization},
  author={Rahman, Md.},
  journal={arXiv preprint arXiv:2501.12345},
  year={2025}
}

@article{yang2025perfcoder,
  title={PerfCoder: Performance-Driven Code Generation},
  author={Yang, Z.},
  journal={arXiv preprint arXiv:2502.54321},
  year={2025}
}

@article{MuSL2025,
  author       = {Changxin Ke and Rui Zhang and Shuo Wang and Li Ding and Guangli Li and
                  Yuanbo Wen and Shuoming Zhang and Ruiyuan Xu and Jin Qin and Jiaming Guo and
                  Chenxi Wang and Ling Li and Qi Guo and Yunji Chen},
  title        = {Mutual-Supervised Learning for Sequential-to-Parallel Code Translation},
  year         = {2025},
  venue        = {CoRR (arXiv)},
  journal      = {arXiv preprint arXiv:2506.11153},
  eprinttype   = {arXiv},
  eprint       = {2506.11153},
  doi          = {10.48550/ARXIV.2506.11153},
  url          = {https://arxiv.org/abs/2506.11153},
}

@article{AlgoTune2025,
  author       = {Press, Ori and Amos, Brandon and Zhao, Haoyu and Wu, Yikai and
                  Ainsworth, Samuel K. and Krupke, Dominik and Kidger, Patrick and Sajed, Touqir and
                  Stellato, Bartolomeo and Park, Jisun and Bosch, Nathanael and Meril, Eli and Steppi, Albert and
                  Zharmagambetov, Arman and Zhang, Fangzhao and Perez-Pineiro, David and Mercurio, Alberto and
                  Zhan, Ni and Abramovich, Talor and Lieret, Kilian and Zhang, Hanlin and Huang, Shirley and
                  Bethge, Matthias and Press, Ofir},
  title        = {AlgoTune: Can Language Models Speed Up General-Purpose Numerical Programs?},
  year         = {2025},
  venue        = {CoRR (arXiv)},
  journal      = {arXiv preprint arXiv:2507.15887},
  eprinttype   = {arXiv},
  eprint       = {2507.15887},
  doi          = {10.48550/ARXIV.2507.15887},
  url          = {https://arxiv.org/abs/2507.15887},
}

@misc{AlgoTuneDataset2025,
  author       = {Press, Ori and Amos, Brandon and Zhao, Haoyu and Wu, Yikai and
                  Ainsworth, Samuel K. and Krupke, Dominik and Kidger, Patrick and Sajed, Touqir and
                  Stellato, Bartolomeo and Park, Jisun and Bosch, Nathanael and Meril, Eli and Steppi, Albert and
                  Zharmagambetov, Arman and Zhang, Fangzhao and Perez-Pineiro, David and Mercurio, Alberto and
                  Zhan, Ni and Abramovich, Talor and Lieret, Kilian and Zhang, Hanlin and Huang, Shirley and
                  Bethge, Matthias and Press, Ofir},
  title        = {AlgoTune Benchmark Dataset},
  year         = {2025},
  venue        = {Hugging Face Datasets},
  url          = {https://huggingface.co/datasets/oripress/AlgoTune},
}

@inproceedings{BabelTower2022,
  author       = {Wen, Yuanbo and Guo, Qi and Fu, Qiang and Li, Xiaqing and Xu, Jianxing and
                  Tang, Yanlin and Zhao, Yongwei and Hu, Xing and Du, Zidong and Li, Ling and
                  Wang, Chao and Zhou, Xuehai and Chen, Yunji},
  title        = {BabelTower: Learning to Auto-parallelized Program Translation},
  year         = {2022},
  venue        = {ICML 2022},
  booktitle    = {Proceedings of the 39th International Conference on Machine Learning},
  series       = {Proceedings of Machine Learning Research},
  volume       = {162},
  pages        = {23685--23700},
  publisher    = {PMLR},
  url          = {https://proceedings.mlr.press/v162/wen22b.html},
  doi          = {unspecified},
}

@article{GraphIt2018,
  author       = {Yunming Zhang and Mengjiao Yang and Riyadh Baghdadi and Shoaib Kamil and
                  Julian Shun and Saman P. Amarasinghe},
  title        = {GraphIt: a high-performance graph DSL},
  year         = {2018},
  venue        = {Proc. ACM Program. Lang. (OOPSLA)},
  journal      = {Proceedings of the ACM on Programming Languages},
  volume       = {2},
  number       = {OOPSLA},
  pages        = {121:1--121:30},
  doi          = {10.1145/3276491},
  url          = {https://doi.org/10.1145/3276491},
}

@inproceedings{Ligra2013,
  author       = {Julian Shun and Guy E. Blelloch},
  title        = {Ligra: a lightweight graph processing framework for shared memory},
  year         = {2013},
  venue        = {PPoPP 2013},
  booktitle    = {Proceedings of the 18th ACM SIGPLAN Symposium on Principles and Practice of Parallel Programming},
  pages        = {135--146},
  publisher    = {ACM},
  doi          = {10.1145/2442516.2442530},
  url          = {https://doi.org/10.1145/2442516.2442530},
}

@article{EvoTune2025,
  author       = {Anja Surina and Amin Mansouri and Lars Quaedvlieg and Amal Seddas and
                  Maryna Viazovska and Emmanuel Abbe and Caglar Gulcehre},
  title        = {Algorithm Discovery With LLMs: Evolutionary Search Meets Reinforcement Learning},
  year         = {2025},
  venue        = {CoRR (arXiv)},
  journal      = {CoRR},
  volume       = {abs/2504.05108},
  eprinttype   = {arXiv},
  eprint       = {2504.05108},
  doi          = {10.48550/ARXIV.2504.05108},
  url          = {https://doi.org/10.48550/arXiv.2504.05108},
}

@misc{ParEvalRepoGitHub2024,
  author       = {Parallel Code Foundry and Daniel Nichols and Yangtian Zi and Zhaojun Xie and Harshitha Menon},
  title        = {ParEval: Parallel Code Evaluation Benchmark (GitHub repository)},
  year         = {2024},
  venue        = {GitHub},
  url          = {https://github.com/parallelcodefoundry/ParEval},
}

@inproceedings{ParEvalRepo2025,
  author       = {Joshua H. Davis and Daniel Nichols and Ishan Khillan and Abhinav Bhatele},
  title        = {ParEval-Repo: A Benchmark Suite for Evaluating LLMs with Repository-level HPC Translation Tasks},
  year         = {2025},
  venue        = {ICPP 2025},
  booktitle    = {Proceedings of the International Conference on Parallel Processing (ICPP 2025)},
  doi          = {10.1145/3754598.3754669},
  url          = {https://doi.org/10.1145/3754598.3754669},
}

@article{ParEvalRepoArXiv2025,
  author       = {Joshua H. Davis and Daniel Nichols and Ishan Khillan and Abhinav Bhatele},
  title        = {ParEval-Repo: A Benchmark Suite for Evaluating LLMs with Repository-level HPC Translation Tasks},
  year         = {2025},
  venue        = {CoRR (arXiv)},
  journal      = {arXiv preprint arXiv:2506.20938},
  eprinttype   = {arXiv},
  eprint       = {2506.20938},
  doi          = {10.48550/ARXIV.2506.20938},
  url          = {https://arxiv.org/abs/2506.20938},
}

@article{TreeSearchEmpiricalSoftware2025,
  author       = {Eser Ayg{\"{u}}n and Anastasiya Belyaeva and Gheorghe Comanici and Marc Coram and Hao Cui and
                  Jake Garrison and Renee Johnston and Anton Kast and Cory Y. McLean and Peter Norgaard and
                  Zahra Shamsi and David Smalling and James Thompson and Subhashini Venugopalan and Brian P. Williams and
                  Chujun He and Sarah Martinson and Martyna Plomecka and Lai Wei and Yuchen Zhou and Qian-Ze Zhu and
                  Matthew Abraham and Erica Brand and Anna Bulanova and Jeffrey A. Cardille and Chris Co and Scott Ellsworth and
                  Grace Joseph and Malcolm Kane and Ryan Krueger and Johan Kartiwa and Dan Liebling and Jan-Matthis Lueckmann and
                  Paul Raccuglia and Xuefei (Julie) Wang and Katherine Chou and James Manyika and Yossi Matias and John C. Platt and
                  Lizzie Dorfman and Shibl Mourad and Michael P. Brenner},
  title        = {An AI system to help scientists write expert-level empirical software},
  year         = {2025},
  venue        = {CoRR (arXiv)},
  journal      = {arXiv preprint arXiv:2509.06503},
  eprinttype   = {arXiv},
  eprint       = {2509.06503},
  doi          = {10.48550/ARXIV.2509.06503},
  url          = {https://arxiv.org/abs/2509.06503},
}

@misc{ParlayInstruct2026,
  author       = {ParEVO},
  title        = {ParEVO project repository (code/data for ParEVO; includes Parlay-Instruct artifacts)},
  year         = {2026},
  venue        = {GitHub},
  url          = {https://github.com/WildAlg/ParEVO}
}

@misc{MuSLCode2025,
  author       = {Changxin Ke},
  title        = {kcxain/musl: code repository for Mutual-Supervised Learning for Sequential-to-Parallel Code Translation},
  year         = {2025},
  venue        = {GitHub},
  url          = {https://github.com/kcxain/musl},
}

@misc{MuSLTranslatorModel2025,
  author       = {kcxain},
  title        = {kcxain/translator-Qwen3-0.6B: MuSL C-to-CUDA translator model (Hugging Face)},
  year         = {2025},
  venue        = {Hugging Face Models},
  url          = {https://huggingface.co/kcxain/translator-Qwen3-0.6B},
}

@misc{GraphItRepo2018,
  author       = {GraphIt-DSL and Yunming Zhang and Tugsbayasgalan Manlaibaatar and Ajay Brahmakshatriya and ykenny1 and Riyadh Baghdadi and Emily Furst and Gregory Siegfried and Benjamin Wade and Laxman Dhulipala},
  title        = {GraphIt-DSL/graphit: GraphIt compiler and DSL implementation (GitHub repository)},
  year         = {2018},
  venue        = {GitHub},
  url          = {https://github.com/GraphIt-DSL/graphit},
}

@software{OpenEvolve2025,
  author       = {Asankhaya Sharma},
  title        = {OpenEvolve: an open-source evolutionary coding agent},
  year         = {2025},
  venue        = {GitHub},
  url          = {https://github.com/algorithmicsuperintelligence/openevolve},
}

@misc{ParlayLib2020,
  author       = {Daniel Anderson and Guy Blelloch and Laxman Dhulipala and Tom Tseng and wheatman and Lorenz H{\"{u}}bschle and Rahul Yesantharao and Xiaojun Dong and aheydon-google},
  title        = {ParlayLib: A Toolkit for Programming Parallel Algorithms on Shared-Memory Multicore Machines (GitHub repository; CMU Parlay Group)},
  year         = {2020},
  venue        = {GitHub},
  url          = {https://github.com/cmuparlay/parlaylib},
}

@article{assumpcao2025codeevolve,
  title={CodeEvolve: an open source evolutionary coding agent for algorithm discovery and optimization},
  author={Henrique Assump{\c{c}}{\~a}o and Diego Ferreira and Leandro Campos and Fabricio Murai},
  journal={arXiv preprint arXiv:2510.14150},
  year={2025}
}

@article{khrulkov2025gigaevo,
  title={GigaEvo: An Open Source Optimization Framework Powered By LLMs And Evolution Algorithms},
  author={Valentin Khrulkov and Andrey Galichin and Denis Bashkirov and Dmitry Vinichenko and Oleg Travkin and Roman Alferov and Andrey Kuznetsov and Ivan Oseledets},
  journal={arXiv preprint arXiv:2511.17592},
  year={2025}
}

@article{du2025afterburner,
  title={Afterburner: Reinforcement Learning Facilitates Self-Improving Code Efficiency Optimization},
  author={Mingzhe Du and Luu Anh Tuan and Yue Liu and Yuhao Qing and Dong Huang and Xinyi He and Qian Liu and Zejun Ma and See-kiong Ng},
  journal={arXiv preprint arXiv:2505.23387},
  year={2025}
}

@article{novikov2025alphaevolve,
  title={AlphaEvolve: A coding agent for scientific and algorithmic discovery},
  author={Alexander Novikov and Ng{\^a}n V{\~u} and Marvin Eisenberger and Emilien Dupont and Po-Sen Huang and Adam Zsolt Wagner and Sergey Shirobokov and Borislav Kozlovskii and Francisco J. R. Ruiz and Abbas Mehrabian and M. Pawan Kumar and Abigail See and Swarat Chaudhuri and George Holland and Alex Davies and Sebastian Nowozin and Pushmeet Kohli and Matej Balog},
  journal={arXiv preprint arXiv:2506.13131},
  year={2025}
}

@article{merouani2025agentic,
  title={Agentic Auto-Scheduling: An Experimental Study of LLM-Guided Loop Optimization},
  author={Massinissa Merouani and Islem Kara Bernou and Riyadh Baghdadi},
  journal={arXiv preprint arXiv:2511.00592},
  year={2025}
}

@article{lei2025pragma,
  title={PRAGMA: A Profiling-Reasoned Multi-Agent Framework for Automatic Kernel Optimization},
  author={Kelun Lei and Hailong Yang and Huaitao Zhang and Xin You and Kaige Zhang and Zhongzhi Luan and Yi Liu and Depei Qian},
  journal={arXiv preprint arXiv:2511.06345},
  year={2025}
}

@inproceedings{singh2024testtime,
  title={Can Test-Time Compute Help {LLMs} Write Low-Resource Parallel Code Better?},
  author={Singh, Gautam and Guha, Arjun and Kailkhura, Bhavya and Menon, Harshitha},
  booktitle={NeurIPS Workshop on Deep Learning for Code (DL4C)},
  year={2024},
  url={https://openreview.net/forum?id=0RnJzt8v84}
}

@misc{qwenparlay2026,
  author = {{ParEVO}},
  title = {Qwen3-30b-SFT-Stage2-Merged},
  year = {2026},
  publisher = {Hugging Face},
  journal = {Hugging Face Repository},
  url = {https://huggingface.co/qqggez/qwen3-30b-sft-stage2-merged}
}

@misc{deepseekparlay2026,
  title={DeepSeek-Parlay-6.7b: A Fine-Tuned Model for Parallel Algorithmic Reasoning},
  author={ParEVO},
  year={2026},
  url={https://huggingface.co/qqggez/deepseek-parlay-6.7b}
}

@misc{qwenrust2026,
  title={Qwen3-Rust-DPO: A Fine-Tuned Model for Safe Parallel Rust},
  author={ParEVO},
  year={2026},
  url={https://huggingface.co/YangLiuWillow/qwen3_rust_dpo_final_merged}
}

@inproceedings{chen2019learningoptimizetensorprograms,
  title     = {Learning to Optimize Tensor Programs},
  author    = {Chen, Tianqi and Zheng, Lianmin and Yan, Eddie and Jiang, Ziheng and Moreau, Thierry and Ceze, Luis and Guestrin, Carlos and Krishnamurthy, Arvind},
  booktitle = {Advances in Neural Information Processing Systems (NeurIPS)},
  pages     = {3393--3404},
  year      = {2018}
}

@inproceedings{chen2024teaching,
  title     = {Teaching Large Language Models to Self-Debug},
  author    = {Chen, Xinyun and Lin, Maxwell and Sch{\"a}rli, Nathanael and Zhou, Denny},
  booktitle = {International Conference on Learning Representations (ICLR)},
  year      = {2024}
}

@inproceedings{cummins2022compilergym,
  title     = {{CompilerGym}: Robust, Performant Compiler Optimization Environments for {AI} Research},
  author    = {Cummins, Chris and Wasti, Bram and Guo, Jiadong and Cui, Brandon and Ansel, Jason and Gomez, Sahir and Jain, Somya and Liu, Jia and Teytaud, Olivier and Steiner, Benoit and Tian, Yuandong and Leather, Hugh},
  booktitle = {IEEE/ACM International Symposium on Code Generation and Optimization (CGO)},
  pages     = {92--105},
  year      = {2022},
  doi       = {10.1109/CGO53902.2022.9741258}
}

@inproceedings{cummins2025llm,
  title     = {{LLM} Compiler: Foundation Language Models for Compiler Optimization},
  author    = {Cummins, Chris and Seeker, Volker and Grubisic, Dejan and Rozi{\`e}re, Baptiste and Gehring, Jonas and Synnaeve, Gabriel and Leather, Hugh},
  booktitle = {Proceedings of the 34th ACM SIGPLAN International Conference on Compiler Construction (CC)},
  year      = {2025},
  doi       = {10.1145/3708493.3712691}
}

@inproceedings{hajali2020neurovectorizer,
  title     = {{NeuroVectorizer}: End-to-End Vectorization with Deep Reinforcement Learning},
  author    = {Haj-Ali, Ameer and Ahmed, Nesreen K. and Willke, Theodore L. and Shao, Yakun Sophia and Asanovic, Krste and Stoica, Ion},
  booktitle = {Proceedings of the 18th ACM/IEEE International Symposium on Code Generation and Optimization (CGO)},
  pages     = {242--255},
  year      = {2020},
  publisher = {ACM},
  doi       = {10.1145/3368826.3377928}
}

@inproceedings{ni2023lever,
  title     = {{LEVER}: Learning to Verify Language-to-Code Generation with Execution},
  author    = {Ni, Ansong and Iyer, Srinivasan and Radev, Dragomir and Stoyanov, Veselin and Yih, Wen-tau and Wang, Sida I. and Lin, Xi Victoria},
  booktitle = {International Conference on Machine Learning (ICML)},
  year      = {2023}
}

@inproceedings{tang2025reasoning,
  title     = {Reasoning Compiler: {LLM}-Guided Optimizations for Efficient Model Serving},
  author    = {Tang, Annabelle Sujun and Priebe, Christopher and Mahapatra, Rohan and Qin, Lianhui and Esmaeilzadeh, Hadi},
  booktitle = {Advances in Neural Information Processing Systems (NeurIPS)},
  year      = {2025}
}

@article{trofin2021mlgo,
  title         = {{MLGO}: A Machine Learning Guided Compiler Optimizations Framework},
  author        = {Trofin, Mircea and Qian, Yundi and Brevdo, Eugene and Lin, Zinan and Choromanski, Krzysztof and Li, David},
  journal       = {arXiv preprint arXiv:2101.04808},
  year          = {2021}
}

@inproceedings{yang2023intercode,
author = {Yang, John and Prabhakar, Akshara and Narasimhan, Karthik and Yao, Shunyu},
title = {InterCode: standardizing and benchmarking interactive coding with execution feedback},
year = {2023},
publisher = {Curran Associates Inc.},
address = {Red Hook, NY, USA},
booktitle = {Proceedings of the 37th International Conference on Neural Information Processing Systems},
articleno = {1035},
numpages = {29},
location = {New Orleans, LA, USA},
series = {NIPS '23}
}

@inproceedings{zheng2020ansor,
  title     = {{Ansor}: Generating High-Performance Tensor Programs for Deep Learning},
  author    = {Zheng, Lianmin and Jia, Chengfan and Sun, Minmin and Wu, Zhao and Yu, Cody Hao and Haj-Ali, Ameer and Wang, Yida and Yang, Jun and Zhuo, Danyang and Sen, Koushik and Gonzalez, Joseph E. and Stoica, Ion},
  booktitle = {14th USENIX Symposium on Operating Systems Design and Implementation (OSDI 20)},
  pages     = {863--879},
  year      = {2020},
  publisher = {USENIX Association}
}

@inproceedings{zhou2024language,
  title     = {Language Agent Tree Search Unifies Reasoning, Acting, and Planning in Language Models},
  author    = {Zhou, Andy and Yan, Kai and Shlapentokh-Rothman, Michal and Wang, Haohan and Wang, Yu-Xiong},
  booktitle = {Proceedings of the 41st International Conference on Machine Learning (ICML)},
  series    = {Proceedings of Machine Learning Research},
  volume    = {235},
  pages     = {62138--62160},
  year      = {2024},
  publisher = {PMLR}
}
\bibliographystyle{icml2026}

\newpage
\appendix
\onecolumn

\section{Evolutionary Coding Agent (ECA) System Prompt}
\label{sec:eca-prompt}
The structural system prompt we specify for the single ECA node is as follows:

\begin{tcolorbox}[colback=backcolour,colframe=gray!50,title=ECA System Prompt,fonttitle=\bfseries]
\small
\begin{verbatim}
You are an expert C++ competitive programmer. Your task is to write a 
COMPLETE, CORRECT, and FAST C++ solution.

PROBLEM:
{problem_description}

REQUIREMENTS:
Write a complete C++ parallel program that compiles and runs correctly
Read input from standard input (cin)
Write output to standard output (cout)
Handle all edge cases mentioned in the problem
Optimize for speed - use efficient algorithms and data structures
Use C++ STL where appropriate (vector, map, set, priority_queue, etc.)
Consider time complexity and space complexity
The parlay library MUST be used as the core computation of the program

AVAILABLE LIBRARIES:
Standard C++ libraries (iostream, algorithm, vector, map, etc.)
The parlay library

Note:
parlay::parallel_for does not guarantee ordering, do not use it with IO operations.

CODE STYLE:
Use C++ style comments: // for single line, /* */ for multi-line
Do NOT use Python-style # comments
Comments should be simple and short
Include necessary headers
Write clean, readable code

OUTPUT FORMAT:
Return ONLY the complete C++ code. Do not include explanations, 
markdown formatting, or code blocks.
Just the raw C++ source code that can be directly compiled.
\end{verbatim}
\end{tcolorbox}

\section{Examples of Fine-Tuning Dataset}
\subsection{Example of Event Generation Strategies.}
\begin{figure*}[t]
    \centering
    \lstset{
        basicstyle=\ttfamily\footnotesize,
        keywordstyle=\color{blue}\bfseries,
        commentstyle=\color{green!60!black}\itshape,
        breaklines=true,
        aboveskip=3pt,
        belowskip=3pt,
        columns=flexible,
        showstringspaces=false
    }

    \begin{minipage}[t]{0.48\textwidth}
        \textbf{Code A (Efficient): Parallel Map-Scan-Write}
        \begin{lstlisting}[language=C++]
// 1. MAP: Count events in parallel (No Locking)
parlay::sequence<int> counts(N);
parlay::parallel_for(0, N, [&](int r) {
    int cnt = 0; /* logic checks ... */
    if (valid) cnt++;
    counts[r] = cnt;
});

// 2. SCAN: Calculate offsets (Prefix Sum)
auto [offsets, total] = parlay::scan(counts);

// 3. WRITE: Parallel Fill (Zero Realloc/Contention)
parlay::sequence<Event> evs(total); // Alloc exact size
parlay::parallel_for(0, N, [&](int r) {
    int k = offsets[r]; 
    /* logic checks ... */
    if (valid) evs[k++] = {u, v, t};
});
        \end{lstlisting}
    \end{minipage}
    \hfill
    \begin{minipage}[t]{0.48\textwidth}
        \textbf{Code B (Inefficient): Sequential Push}
        \begin{lstlisting}[language=C++]
// 1. Setup Vector (Heuristic reservation)
std::vector<Event> events;
events.reserve(2 * N * N); // May still realloc

// 2. Iterate Sequentially (Cannot Parallelize)
for (int r = 0; r < N; ++r) {
    for (int c = 0; c < N; ++c) {
        /* logic checks ... */
        
        // BOTTLENECK: Single thread, capacity checks,
        // and reallocation overhead.
        if (valid) {
            events.push_back({u, v, t});
        }
    }
}
        \end{lstlisting}
    \end{minipage}
    \vspace{-0.5em}
    \caption{\textbf{Comparison of Event Generation Strategies.} \textbf{Left:} Code A employs a Map-Scan-Write pattern to enable lock-free parallel writing. \textbf{Right:} Code B relies on sequential \texttt{push\_back}, preventing parallelization and incurring reallocation costs.}
    \label{fig:ft-data-code-comparison}
\end{figure*}

\subsection{Example of Hidden Unit Test}
\cref{fig:data_example} is an example of the finetuning data and the corresponding unit test for verification. 
\begin{figure}[H]
\centering
\begin{tcolorbox}[
    colback=gray!5, 
    colframe=gray!40, 
    title=\textbf{Example Data Point from \textit{Parlay-Instruct}},
    fonttitle=\bfseries\small,
    boxrule=0.8pt,
    arc=2pt
]
\textbf{Instruction:} \\
\small Filter even numbers from 0 to 100,000 using \texttt{parlay::delayed::filter\_op} and materialize the result.

\vspace{0.5em}
\hrule
\vspace{0.5em}

\textbf{Target Code (Verified):}
\begin{lstlisting}[language=C++, basicstyle=\ttfamily\footnotesize, keywordstyle=\color{blue}, commentstyle=\color{green!60!black}, breaklines=true, numbers=none]
auto seq = parlay::iota(100000);
// specific delayed filter operation
auto f = parlay::delayed::filter_op(seq, [](int x) { 
    return x%2==0 ? std::optional{x} : std::nullopt; 
});
auto s = parlay::delayed::to_sequence(f);
\end{lstlisting}

\vspace{0.5em}
\hrule
\vspace{0.5em}

\textbf{Hidden Unit Test:}
\begin{lstlisting}[language=C++, basicstyle=\ttfamily\scriptsize, keywordstyle=\color{purple}, breaklines=true]
int main() {
    // ... [setup code injected here] ...
    assert(s.size() == 50000);
    for(int i=0; i<50000; i++) assert(s[i] == i*2);
    std::cout << "Passed";
}
\end{lstlisting}
\end{tcolorbox}
\vspace{-1em}
\caption{A representative sample from the training corpus. Each sample includes a natural language instruction, the ground-truth parallel implementation, and an executable unit test used for verification.}
\label{fig:data_example}
\end{figure}

\section{Detailed Experimental Data}



\subsection{Comprehensive ParEval Benchmarks}
\cref{tab:pareval-results-full} provides the complete breakdown of `Build@1`, `Pass@1`, and `Speedup@1` metrics across commercial and open-weight models. The fine-tuned \sys models consistently outperform baselines in compilation rates and execution speed.

\begin{table}[H]
\centering
\small
\renewcommand{\arraystretch}{1.1}
\definecolor{parlaygreen}{rgb}{0.9, 1.0, 0.9}
\definecolor{rustpurple}{rgb}{0.95, 0.9, 1.0}
\begin{tabular}{llllccccc}
\toprule
\textbf{Model} & \textbf{Temp.} & \textbf{Code} & \textbf{Sched.} & \textbf{Build@1} & \textbf{Pass@1} & \multicolumn{3}{c}{\textbf{Speedup@1}} \\
\cmidrule(lr){7-9}
 & & & & & & \textbf{AM} & \textbf{GM w/ Out.} & \textbf{GM w/o Out.} \\
\midrule
\rowcolor{parlaygreen} Gemini-2.5-Flash & 0.2 & Parlay & Parlay  & 0.57 & 0.29   & 6.88 & 2.75 & 2.41\\
\rowcolor{parlaygreen} Gemini-2.5-Pro & 0.2 & Parlay & Parlay  & 0.84 & 0.54 & 85.34 & 5.47 & 3.17 \\
\rowcolor{parlaygreen} Gemini-3-Pro & 0.2 & Parlay & Parlay  & 0.25 & 0.23 & 3.68 & 1.53 & 1.42 \\
\rowcolor{parlaygreen} GPT-5 Thinking & 0.2 & Parlay & Parlay  & 0.93 & 0.05 & 0.43 & 1.13 & 1.13 \\
\rowcolor{parlaygreen} Claude Opus 4.5 & 0.2 & Parlay & Parlay  & 0.97 & 0.05 & 0.42 & 1.13 & 1.13 \\
\rowcolor{parlaygreen} \textbf{Gemini-2.5-Parlay (\sys)} & 0.2 & \textbf{Parlay} & \textbf{Parlay} & \textbf{0.81} & \textbf{0.58} & \textbf{107.43} & \textbf{6.33} & \textbf{3.74} \\
\rowcolor{parlaygreen} \textbf{Qwen3-Parlay (\sys)} & 0.2 & \textbf{Parlay} & \textbf{Parlay}  & \textbf{0.50} & \textbf{0.33}   & \textbf{5.01} & \textbf{1.87} & \textbf{1.64} \\

\rowcolor{parlaygreen} DeepSeek-6.7B-Base & 0.2 & Parlay & Parlay  & 0.89 & 0.11 & 0.73 & 1.14 & 1.14 \\
\rowcolor{parlaygreen} DeepSeek-Syntax & 0.2 & Parlay & Parlay  & 0.85 & 0.12   & 2.42  & 1.38 & 1.29\\
\rowcolor{parlaygreen} \textbf{DeepSeek-Parlay (\sys)} & 0.2 & \textbf{Parlay} & \textbf{Parlay}  & \textbf{0.81} & \textbf{0.26}   & \textbf{126.16}  &  \textbf{3.14} & \textbf{1.63} \\

\rowcolor{parlaygreen} Qwen2.5-Coder-32B & 0.2 & Parlay & Parlay & 0.93 & 0.11 & 5.02 & 1.82 & 1.59 \\
\rowcolor{parlaygreen} Qwen2.5-Coder-32B & 0.7 & Parlay & Parlay  & 0.93 & 0.18 & 11.80 & 2.65 & 1.98 \\
\rowcolor{parlaygreen} Qwen2.5-Coder-32B-Instruct & 0.2 & Parlay & Parlay  & 0.61 & 0.41 & 7.00 & 2.22 & 1.82 \\
\rowcolor{parlaygreen} Qwen3-Coder-30B-Instruct & 0.2 & Parlay & Parlay  & 0.51 & 0.28 & 5.15 & 1.96 & 1.72 \\
\rowcolor{parlaygreen} DeepSeek-Coder-V2-Lite-Base & 0.2 & Parlay & Parlay  & 0.80 & 0.09   & 1.30 & 1.19 & 1.19 \\
\rowcolor{parlaygreen} DeepSeek-Coder-V2-Lite-Base & 0.7 & Parlay & Parlay  & 0.92 & 0.14 & 4.74 & 1.71 & 1.59 \\
\rowcolor{parlaygreen} StarCoder2-15B & 0.2 & Parlay & Parlay  & 0.80 & 0.27 & 7.65 & 2.02 & 1.62 \\

\midrule
\rowcolor{parlaygreen} Gemini-3-Pro & 0.2 & OMP & OMP  & 0.78 & 0.72 & 445.14 & 6.75 & 3.20  \\
\rowcolor{parlaygreen} Gemini-2.5-Parlay & 0.2 & OMP & OMP  & 0.94 & 0.71 & 373.78 & 7.77 & 3.22 \\
\rowcolor{parlaygreen} GPT-5 Thinking & 0.2 & OMP & OMP  & 0.95 & 0.06 & 0.60 &  1.15 & 1.15 \\
\rowcolor{parlaygreen} Claude Opus 4.5 & 0.2 & OMP & OMP  & 0.97 & 0.07 & 0.52 &  1.14 & 1.14 \\
\rowcolor{parlaygreen} Qwen2.5-Coder-32B-Instruct & 0.2 & OMP & OMP  & 0.91 & 0.65 & 10.47 & 2.77 & 1.98 \\
\rowcolor{parlaygreen} Qwen2.5-Coder-32B-Instruct & 0.7 & OMP & OMP & 0.92 & 0.65 & 11.20 & 3.44 & 2.46 \\
\rowcolor{parlaygreen} Qwen3-Coder-30B-Instruct & 0.2 & OMP & OMP  & 0.86 & 0.55 & 11.88 & 2.45 & 1.61 \\
\rowcolor{parlaygreen} Qwen3-Coder-30B-Instruct & 0.7 & OMP & OMP  & 0.91 & 0.56 & 11.47 & 2.53 & 1.68 \\
\rowcolor{parlaygreen} Qwen2.5-Coder-32B & 0.2 & OMP & OMP  & 0.98 & 0.35 & 10.37 & 2.56 & 1.61 \\
\rowcolor{parlaygreen} Qwen2.5-Coder-32B & 0.7 & OMP & OMP & 0.97 & 0.39 & 11.51 & 3.48 & 2.47 \\
\rowcolor{parlaygreen} DeepSeek-Coder-V2-Lite-Base & 0.2 & OMP & OMP  & 0.82 & 0.24 & 5.60 & 2.25 & 2.09\\
\rowcolor{parlaygreen} DeepSeek-Coder-V2-Lite-Base & 0.7 & OMP & OMP  & 0.96 & 0.31 & 12.47 & 3.35 & 2.21 \\
\rowcolor{parlaygreen} StarCoder2-15B & 0.2 & OMP & OMP  & 0.97 & 0.26 & 8.23 & 1.95 & 1.47 \\

\midrule
\rowcolor{rustpurple} Gemini-3-Pro & 0.2 & Rust & Rayon & 0.97	& 0.82 &	7.32 & 0.69 & 0.61  \\
\rowcolor{rustpurple} Qwen2.5-Coder-32B-Instruct & 0.2 & Rust & Rayon & 0.63 & 0.49 & 4.61 & 0.57 & 0.51 \\
\rowcolor{rustpurple} Qwen2.5-Coder-32B-Instruct & 0.7 & Rust & Rayon & 0.70	& 0.48	& 5.69 & 0.51 & 0.44\\
\rowcolor{rustpurple} \textbf{Qwen3-Rust (ParEVO)} & 0.2 & \textbf{Rust} & \textbf{Rayon}  & \textbf{0.64} & \textbf{0.46} & \textbf{4.62} & \textbf{0.35} & \textbf{0.29} \\ 
\rowcolor{rustpurple} Qwen3-Coder-30B-Instruct & 0.2 & Rust & Rayon & 0.61 & 0.50 & 4.57 & 0.46 & 0.38 \\
\rowcolor{rustpurple} Qwen3-Coder-30B-Instruct & 0.7 & Rust & Rayon & 0.66 & 0.49 & 4.94 & 0.29 & 0.24 \\
\rowcolor{rustpurple} Qwen2.5-Coder-32B & 0.2 & Rust & Rayon & 0.82 & 0.45 & 4.43 & 0.63 & 0.63 \\
\rowcolor{rustpurple} Qwen2.5-Coder-32B & 0.7 & Rust & Rayon & 0.86 & 0.38 & 4.35 & 0.79 & 0.79 \\
\rowcolor{rustpurple} DS-Coder-V2-Lite-Base & 0.2 & Rust & Rayon  & 0.73 & 0.29 & 3.47 & 0.58 & 0.54 \\
\rowcolor{rustpurple} DS-Coder-V2-Lite-Base & 0.7 & Rust & Rayon  & 0.85 & 0.25 & 4.96 & 0.89 & 0.78 \\
\rowcolor{rustpurple} StarCoder2-15B & 0.2 & Rust & Rayon & 0.77 & 0.27 & 2.46 & 0.79 & 0.74 \\
\rowcolor{rustpurple} StarCoder2-15B & 0.7 & Rust & Rayon & 0.82 & 0.25 & 5.67 & 0.63 & 0.49 \\
\rowcolor{rustpurple} DS-Coder-V2-Lite-Instruct & 0.2 & Rust & Rayon  & 0.40 & 0.02 & 0.15 & 1.04 & 1.04 \\
\rowcolor{rustpurple} DS-Coder-V2-Lite-Instruct & 0.7 & Rust & Rayon  & 0.48 & 0.02 & 0.15 & 1.04 & 1.04 \\
\bottomrule
\end{tabular}
\caption{Comprehensive ParEval results for commercial and local models. Shaded regions distinguish C++/ParlayLib/OMP models (Green) from Rust/Rayon models (Purple). To isolate pure parallel scaling, we also conducted an ablation removing programs with speedups > 32x (indicating algorithmic improvement beyond physical core counts), as denoted by the GM w/o Out. column. }
\label{tab:pareval-results-full}
\end{table}

\subsection{Metric Breakdown by Problem Type}
\begin{figure*}[h!]
    \centering
    \begin{subfigure}[b]{0.32\textwidth}
        \includegraphics[width=\linewidth]{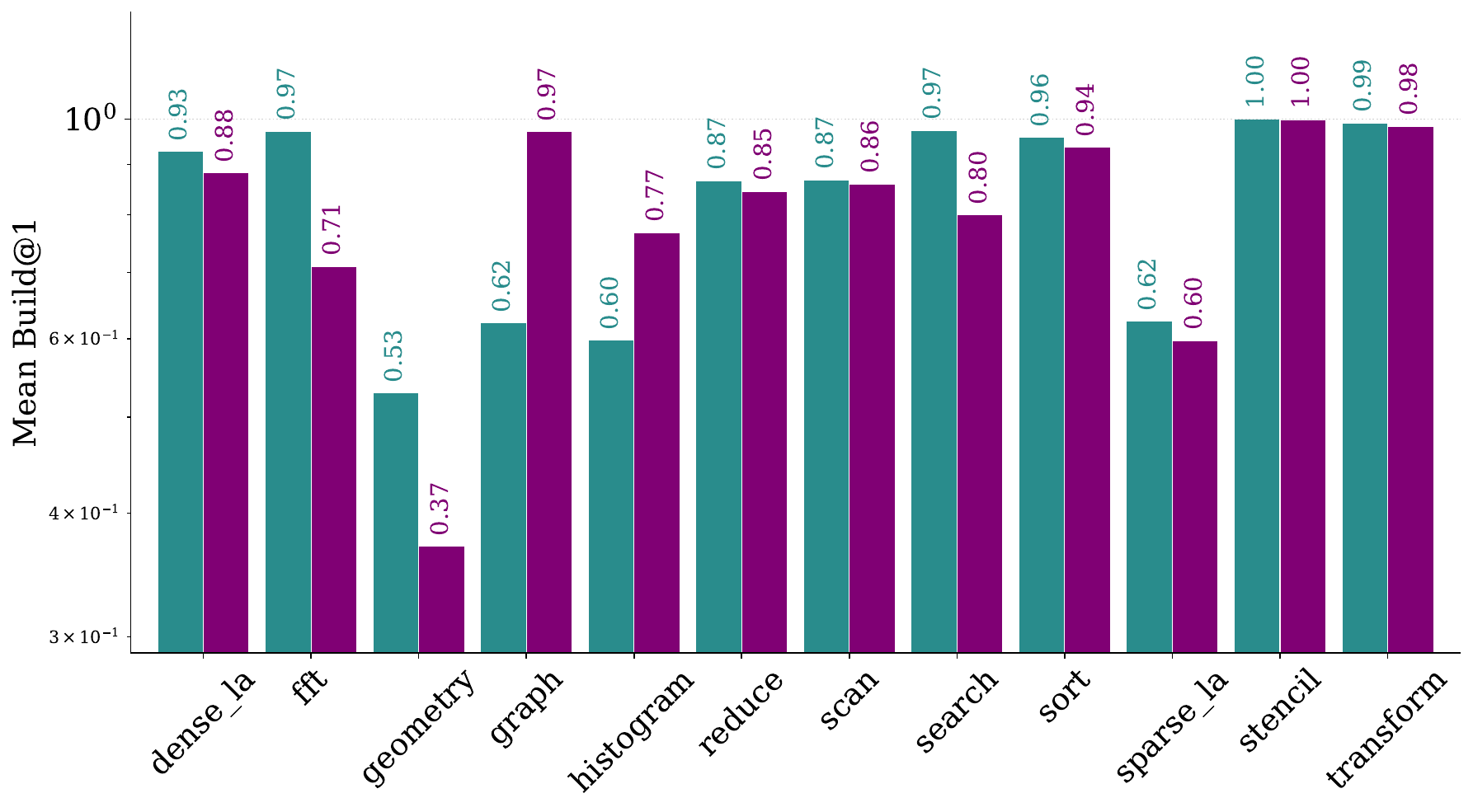}
        \caption{Mean Build@1}
    \end{subfigure}
    \begin{subfigure}[b]{0.32\textwidth}
        \includegraphics[width=\linewidth]{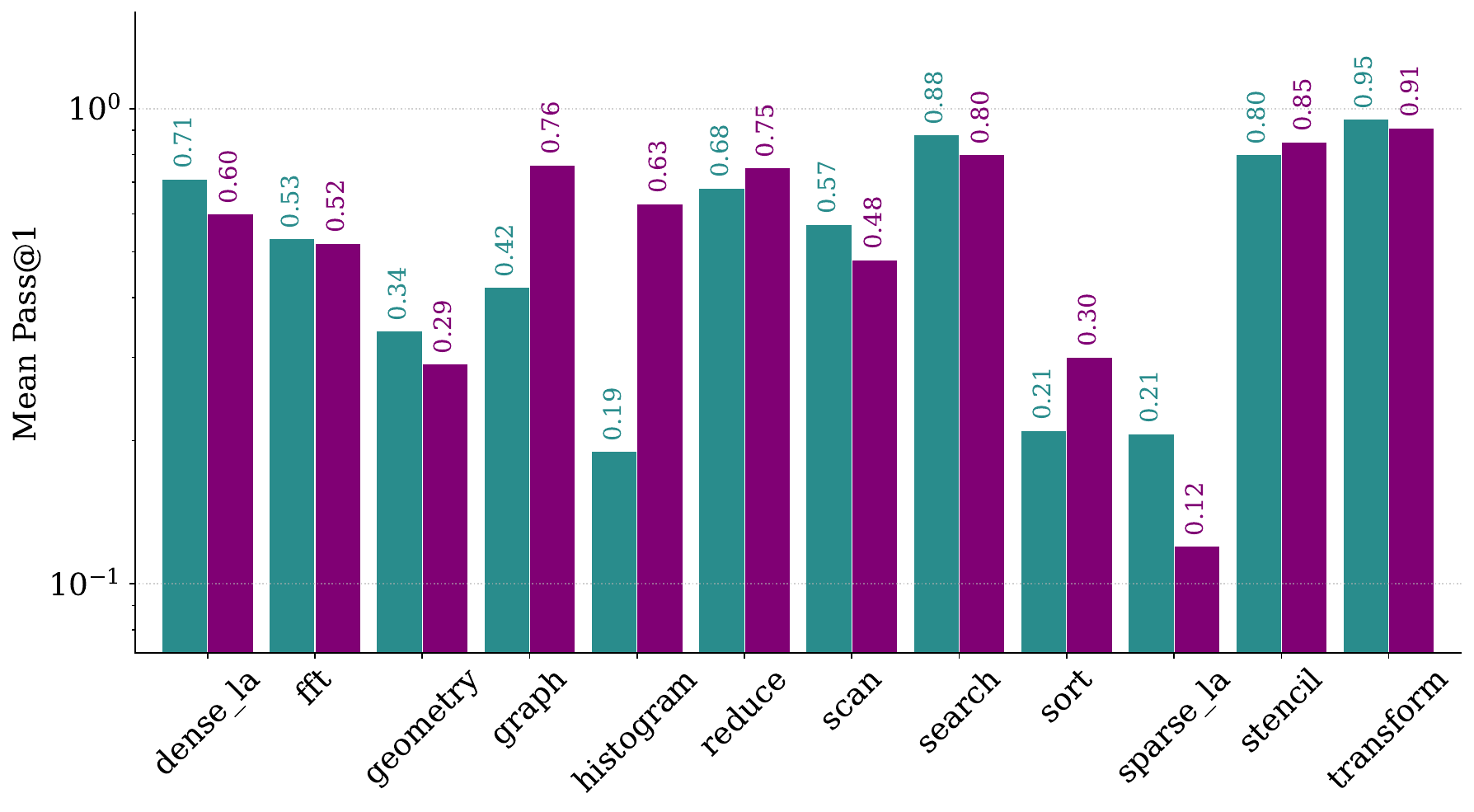}
        \caption{Mean Pass@1}
    \end{subfigure}
    \begin{subfigure}[b]{0.32\textwidth}
        \includegraphics[width=\linewidth]{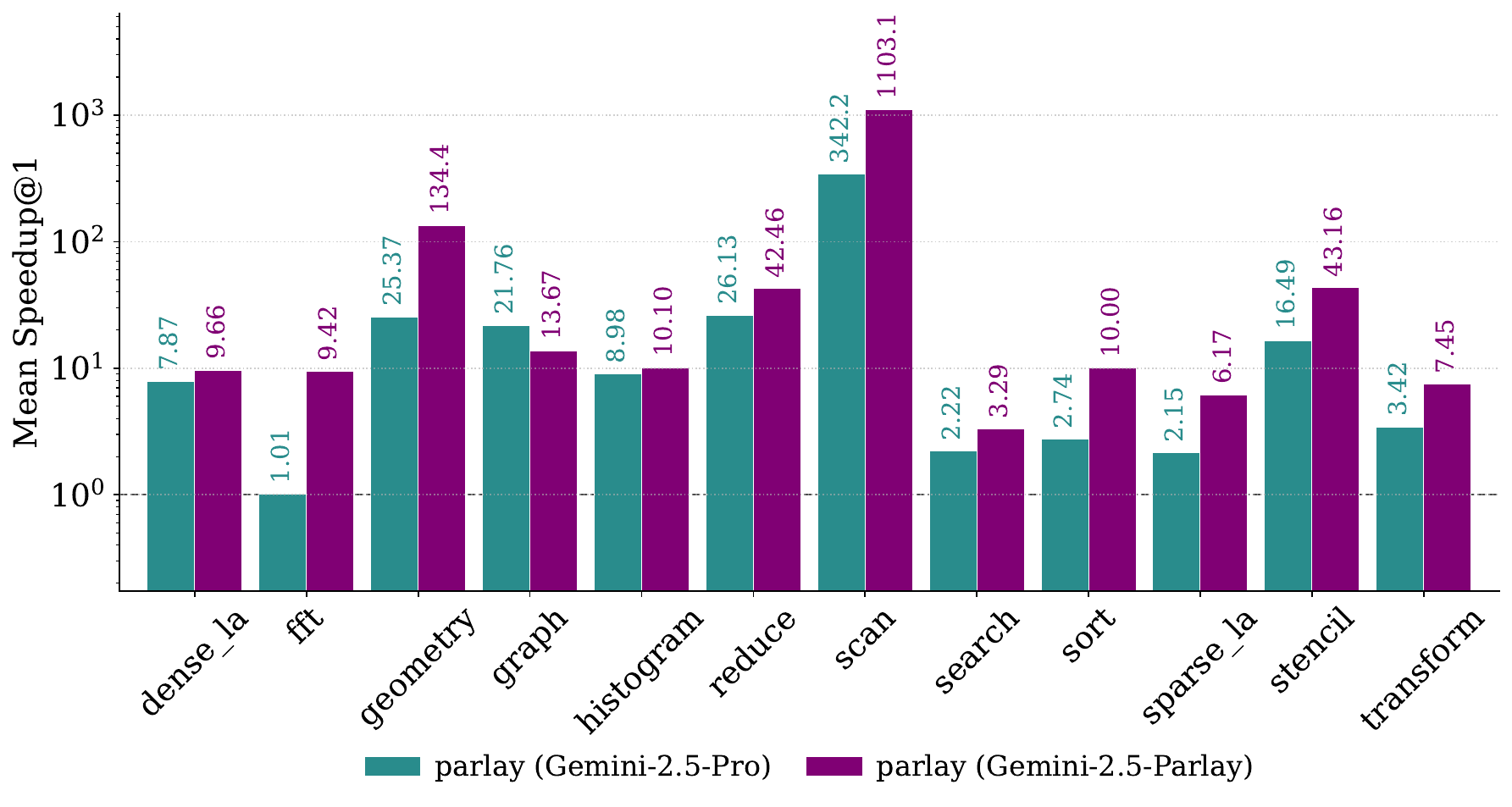}
        \caption{Mean Speedup@1}
    \end{subfigure}
    \caption{ParEval Metrics Comparison between Gemini-2.5-Pro and Gemini-2.5-Parlay, where the former is the base model and the latter is the base model finetuned on the Parlay-Instruct and DMOJ datasets. (a-c) highlight that fine-tuning significantly improves the model's ability to construct valid ParlayLib code, with substantial gains in build and pass rates as well as improved running time over the base model.}
    \label{fig:pareval-ft-gemini-code-contest}
\end{figure*}

\begin{figure*}[h]
    \centering
    \begin{subfigure}[b]{0.32\textwidth}
        \includegraphics[width=\linewidth]{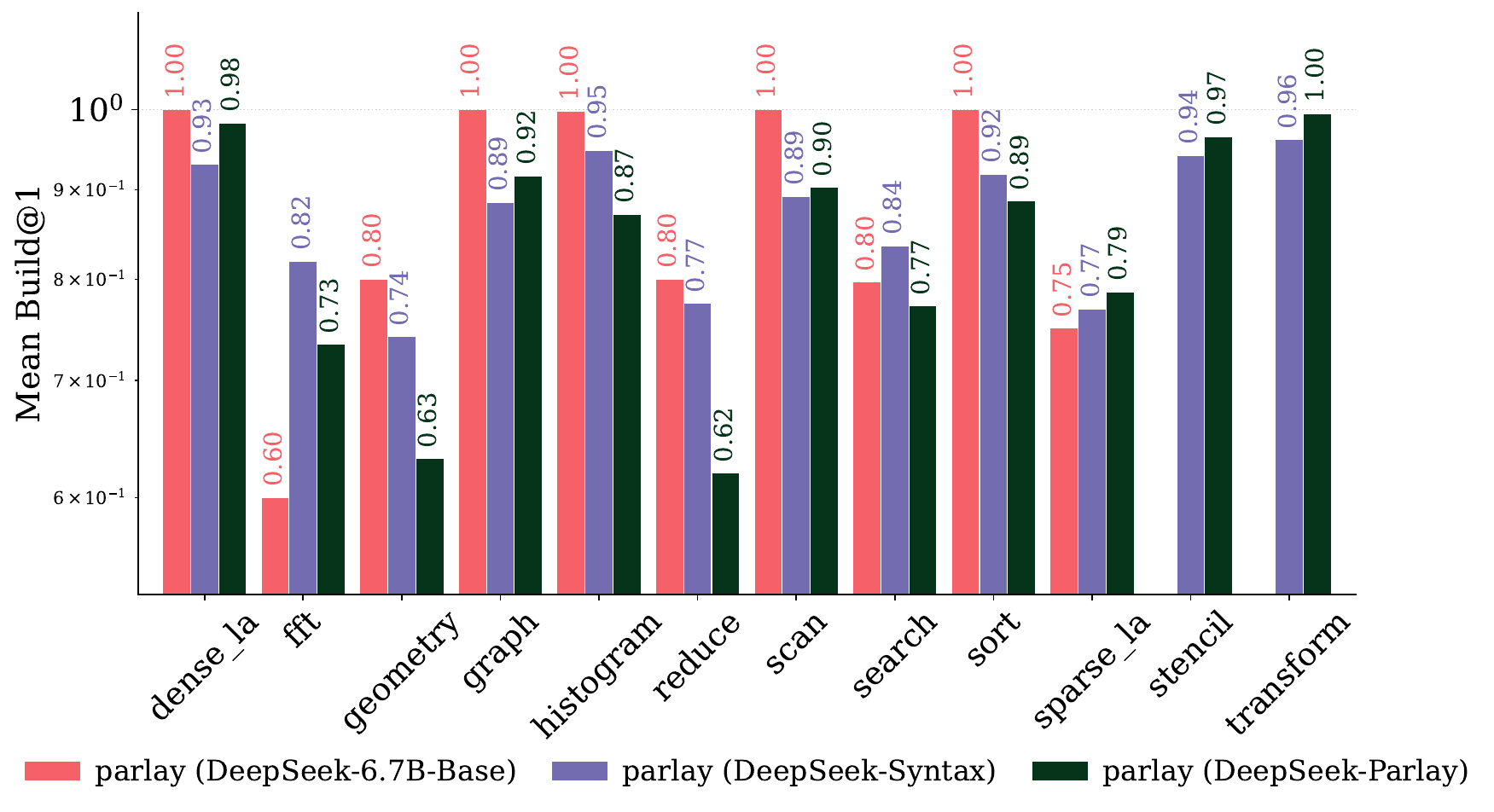}
        \caption{Mean Build@1}
    \end{subfigure}
    \begin{subfigure}[b]{0.32\textwidth}
        \includegraphics[width=\linewidth]{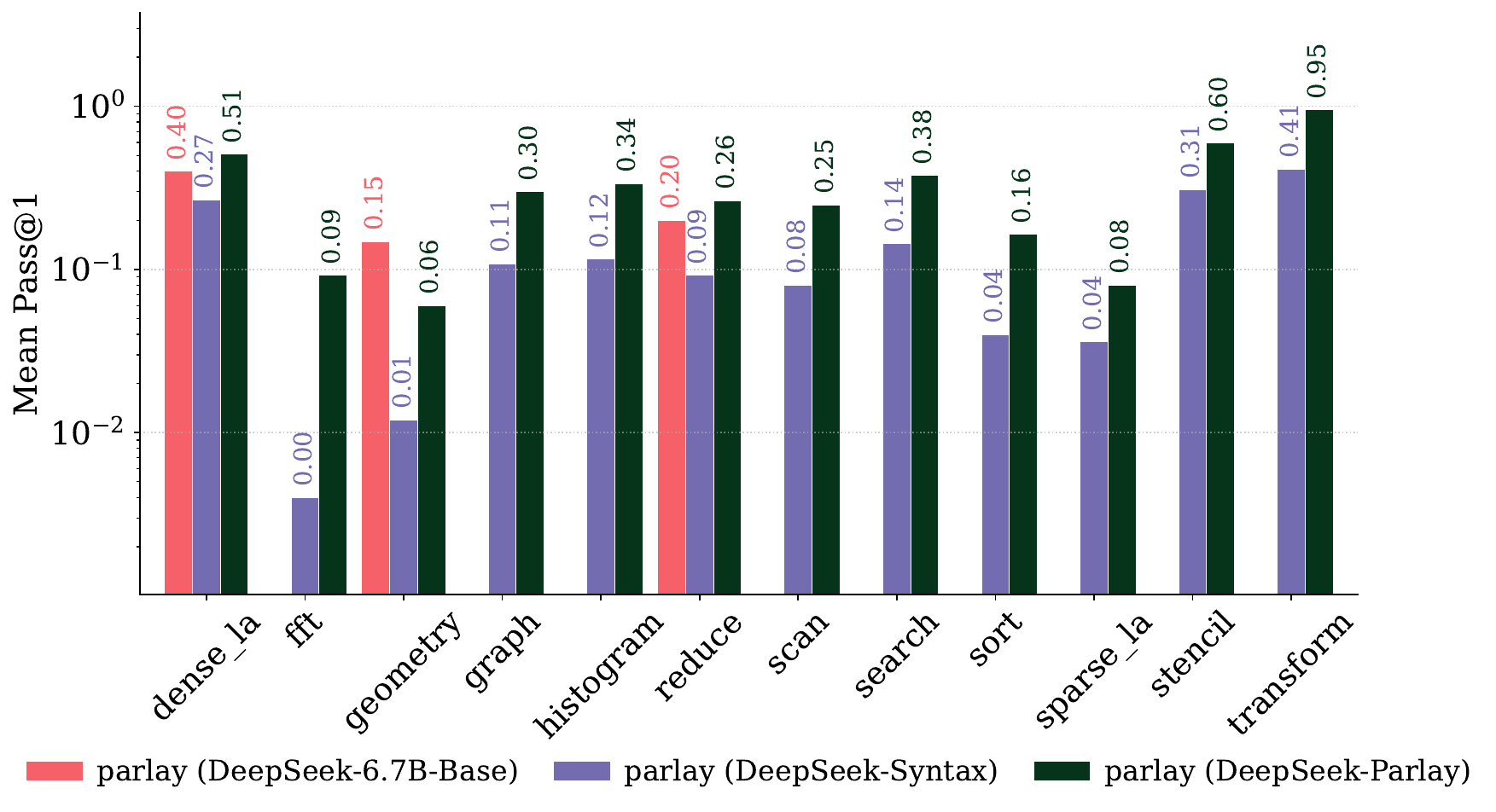}
        \caption{Mean Pass@1}
    \end{subfigure}
    \begin{subfigure}[b]{0.32\textwidth}
        \includegraphics[width=\linewidth]{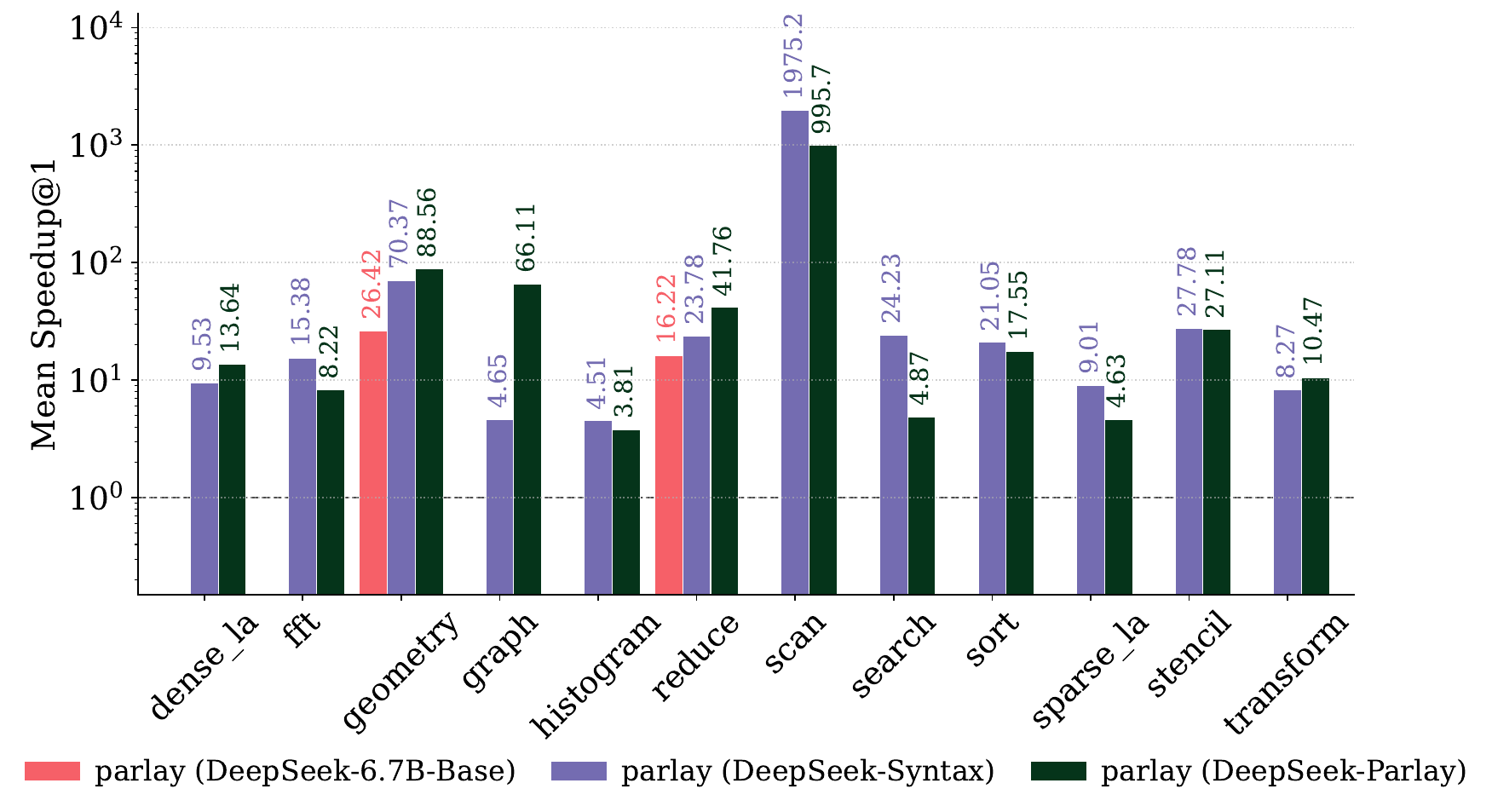}
        \caption{Mean Speedup@1}
    \end{subfigure}
    \caption{Impact of Fine-tuning on DeepSeek-6.7B. The fine-tuned model (DeepSeek-Parlay) shows massive gains in pass rate and speedup compared to the base model. The DeepSeek-Syntax model is the finetuned model of DeepSeek-6.7B-Base purely on ParlayLib syntax.}
    \label{fig:pareval-ft-deepseek-code-contest}
\end{figure*}

\begin{figure*}[t]
    \centering
    \begin{subfigure}[b]{0.32\textwidth}
        \includegraphics[width=\linewidth]{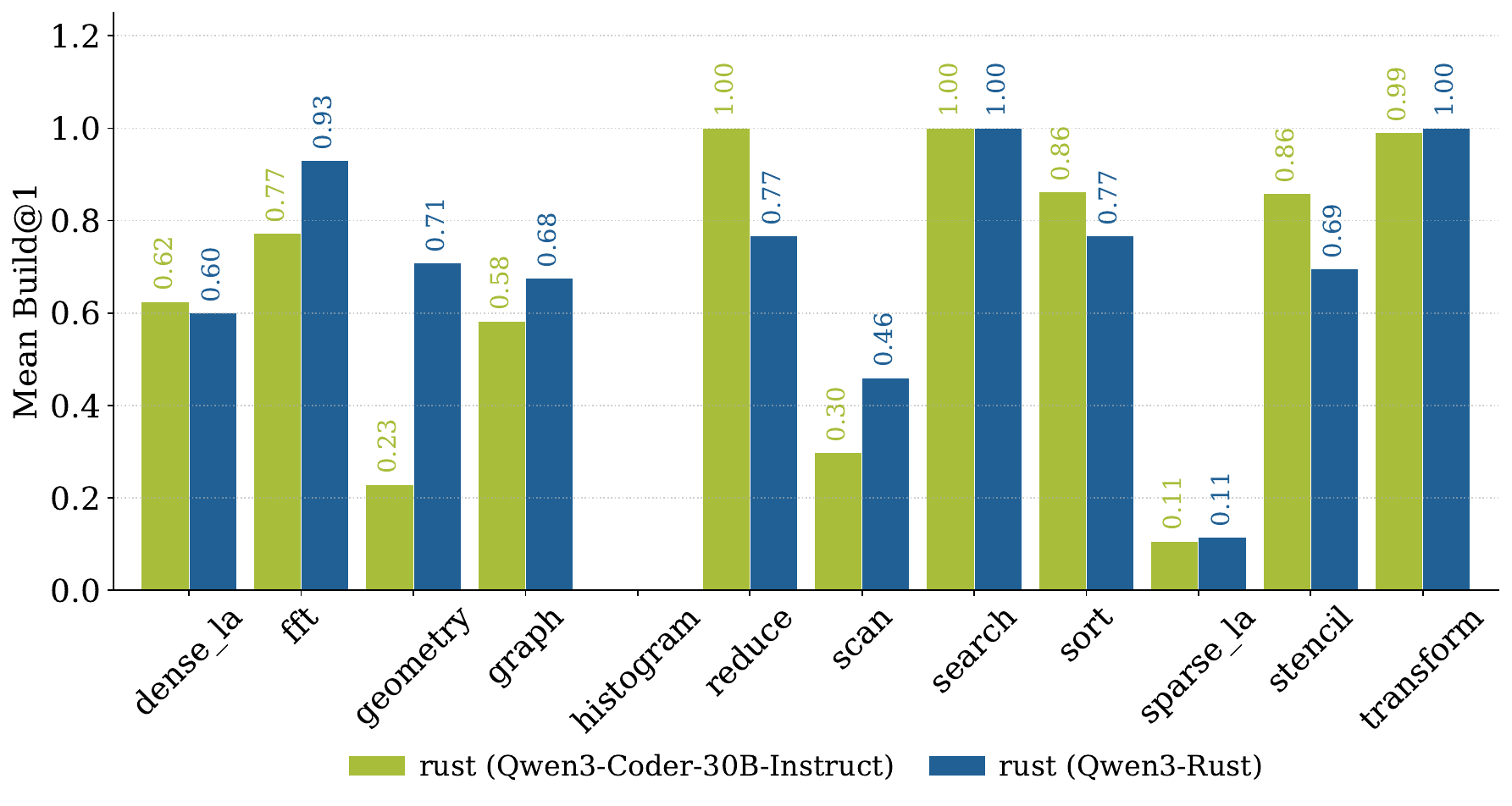}
        \caption{Mean Build@1}
    \end{subfigure}
    \begin{subfigure}[b]{0.32\textwidth}
        \includegraphics[width=\linewidth]{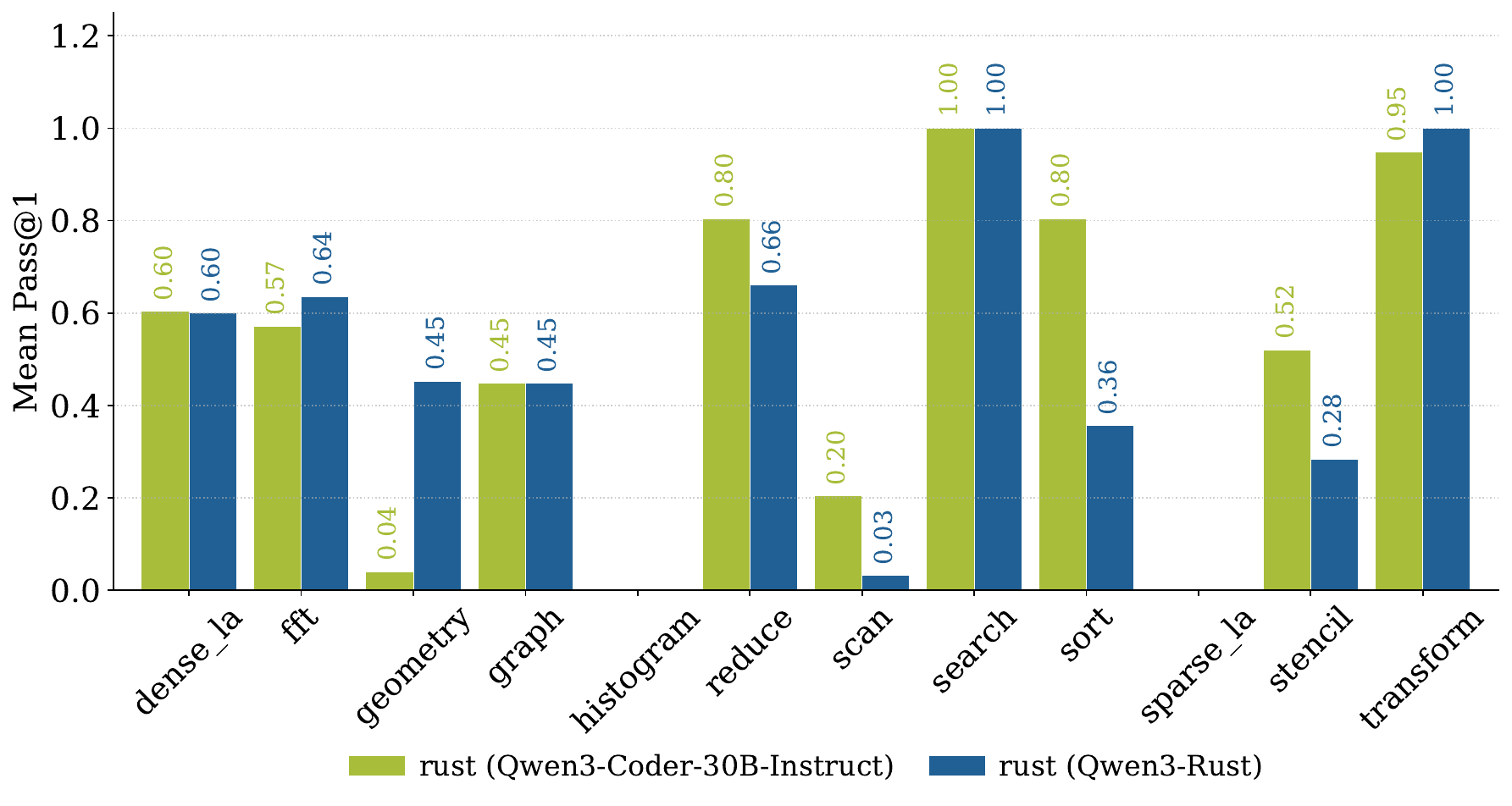}
        \caption{Mean Pass@1}
    \end{subfigure}
    \begin{subfigure}[b]{0.32\textwidth}
        \includegraphics[width=\linewidth]{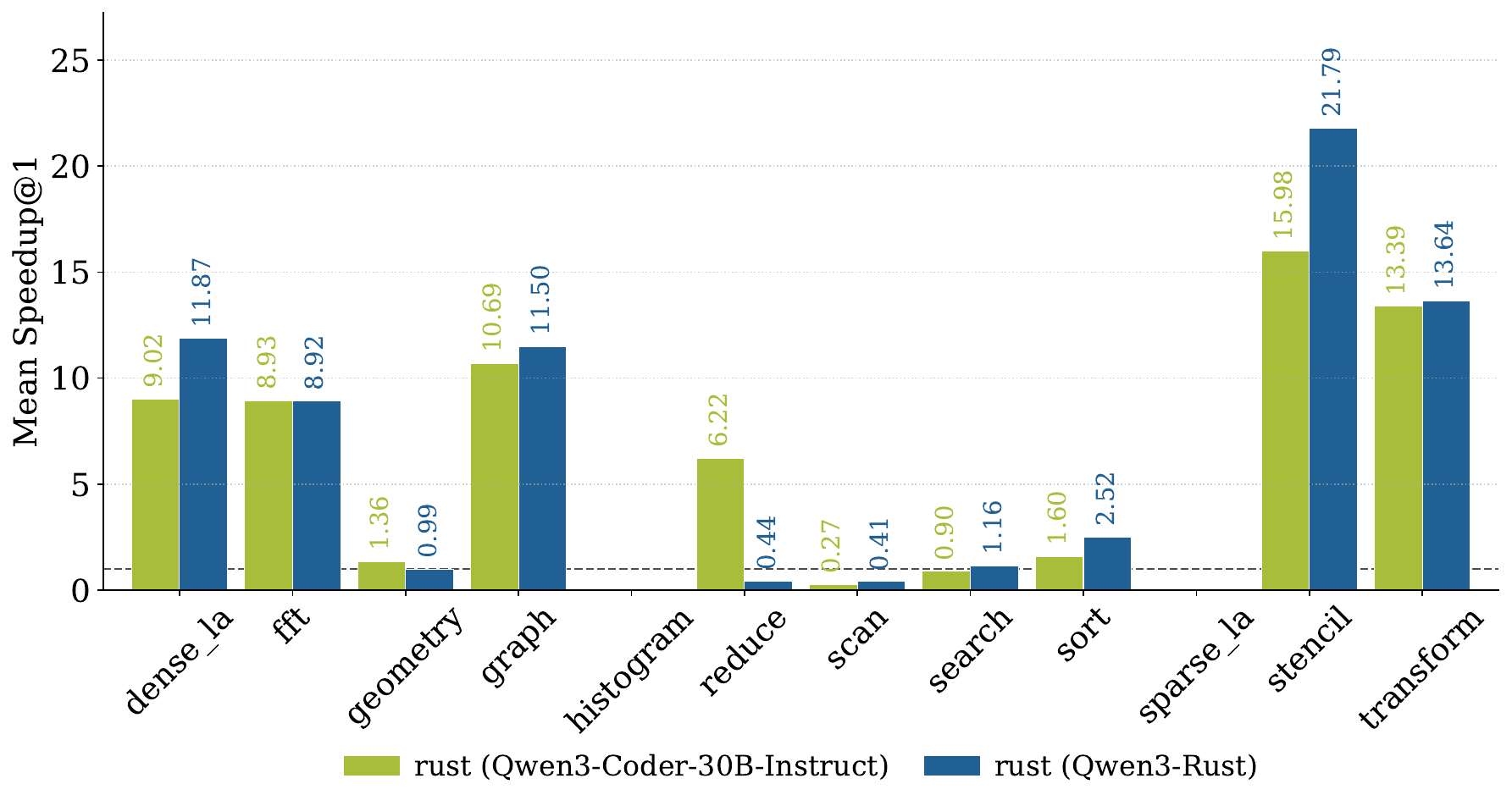}
        \caption{Mean Speedup@1}
    \end{subfigure}
    \caption{Impact of Fine-tuning on Qwen3-Coder-30B-A3B-Instruct. The fine-tuned model (Qwen3-Rust) shows gains in speedup compared to the base model. }
    \label{fig:pareval-ft-qwen3-rust}
\end{figure*}

To understand the specific impact of fine-tuning, we visualize the shift in metrics across problem types. \cref{fig:pareval-ft-gemini-code-contest} and \cref{fig:pareval-ft-deepseek-code-contest} demonstrate that while fine-tuning universally improves \texttt{Build} and \texttt{Pass} rates, the \texttt{Speedup} gains are most pronounced in the irregular graph and complex arithmetic categories. This applies to \cref{fig:pareval-ft-qwen3-rust} except that we see occasional lower \texttt{Pass} rate after finetuning.

\subsection{Comparison vs. Expert Human Baselines (PBBS \& RPB)}
\begin{figure}[h!]
    \centering
    \begin{subfigure}[b]{0.3\columnwidth}
        \includegraphics[width=\linewidth]{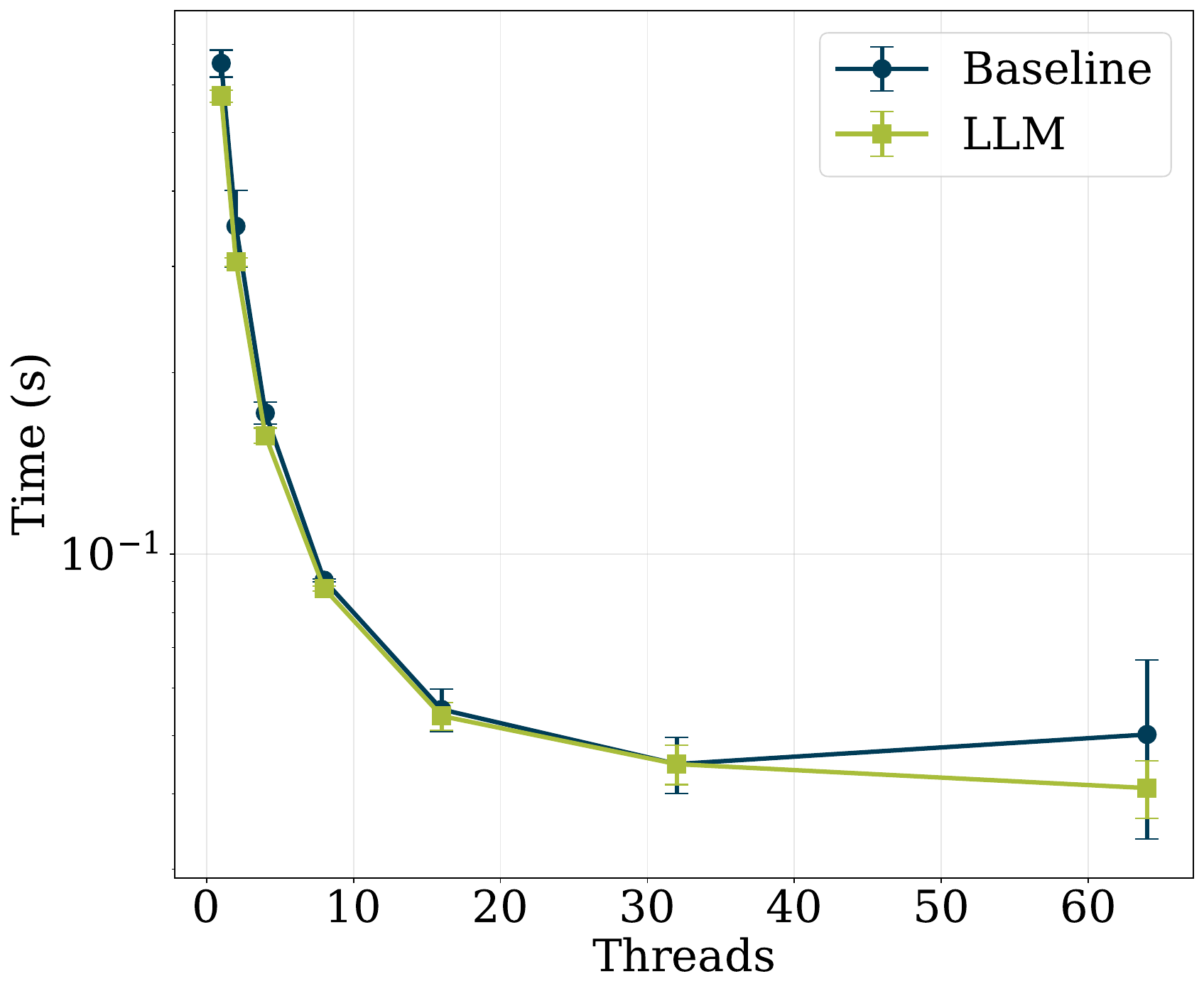}
        \caption{MM Runtime (Rust)}
    \end{subfigure}
    \begin{subfigure}[b]{0.3\columnwidth}
        \includegraphics[width=\linewidth]{RPB/mm_3Dgrid_E_8000000_Scalability.pdf}
        \caption{MM Scalability (Rust)}
    \end{subfigure}
    \begin{subfigure}[b]{0.3\columnwidth}
        \includegraphics[width=\linewidth]{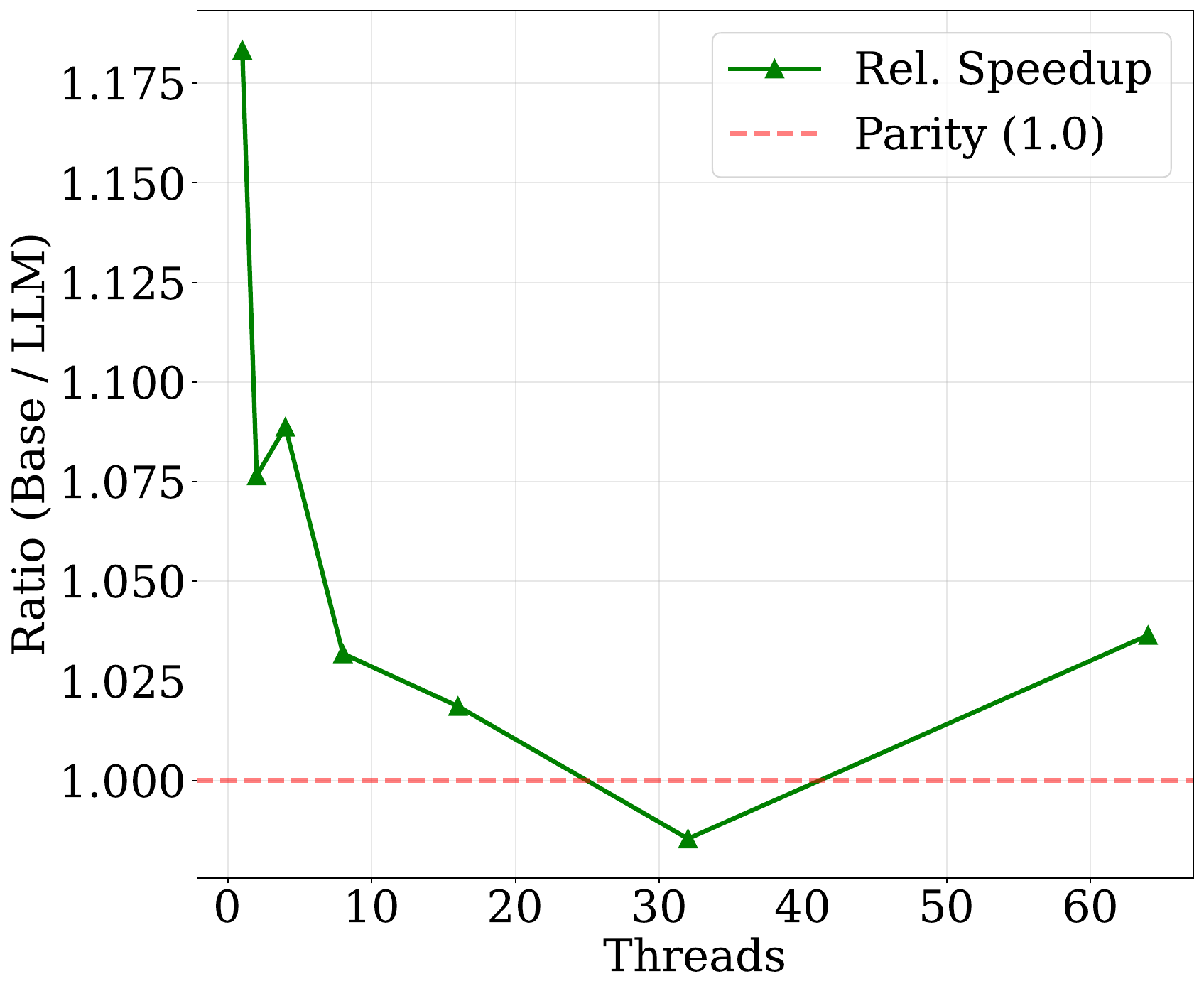}
        \caption{MM Relative Speedup (Rust)}
    \end{subfigure}
    
    \begin{subfigure}[b]{0.3\columnwidth}
        \includegraphics[width=\linewidth]{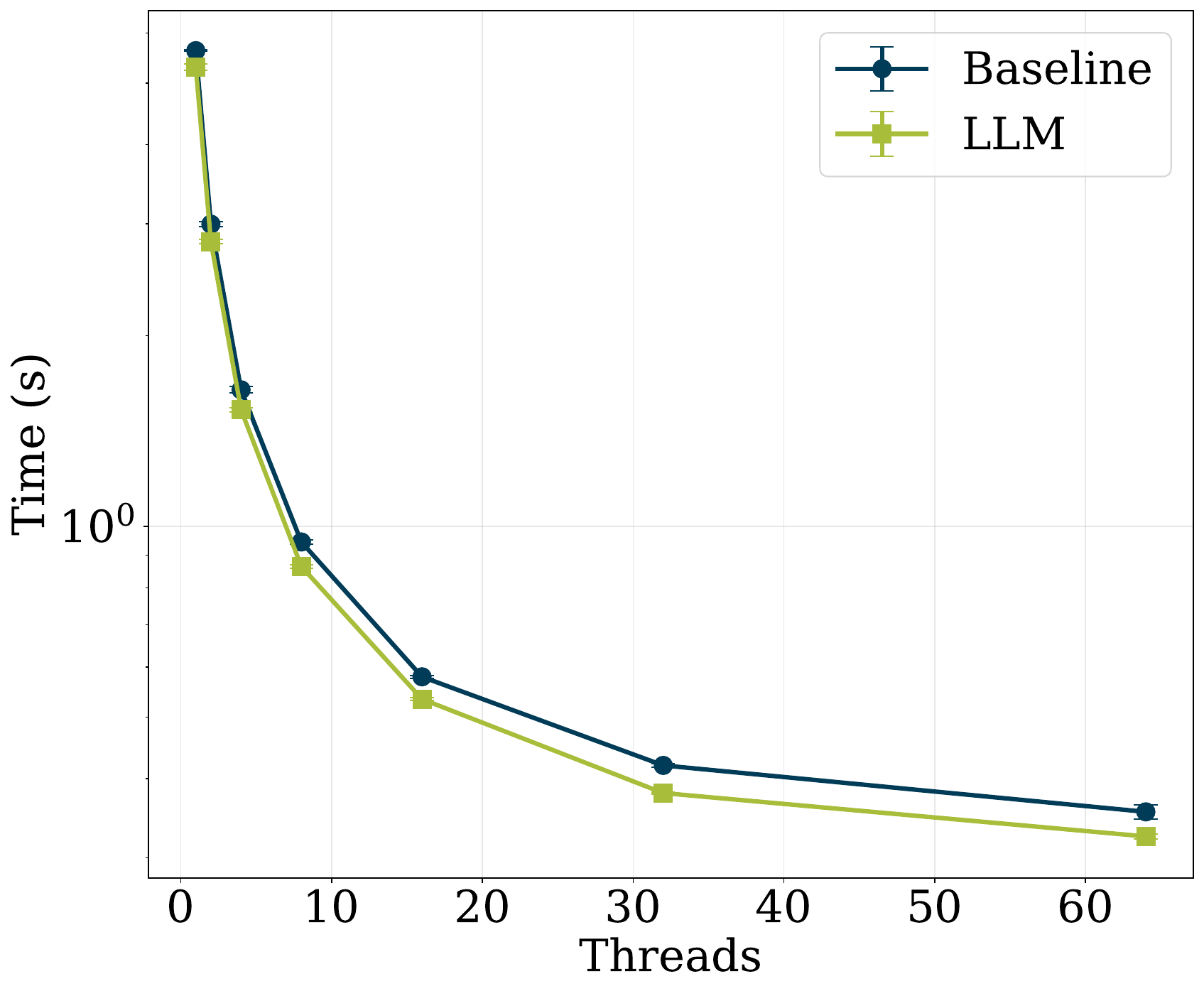}
        \caption{MSF Runtime (Rust)}
    \end{subfigure}
    \begin{subfigure}[b]{0.3\columnwidth}
        \includegraphics[width=\linewidth]{RPB/msf_rMatGraph_WE_12_2250000_Scalability.pdf}
        \caption{MSF Scalability (Rust)}
    \end{subfigure}
    \begin{subfigure}[b]{0.3\columnwidth}
        \includegraphics[width=\linewidth]{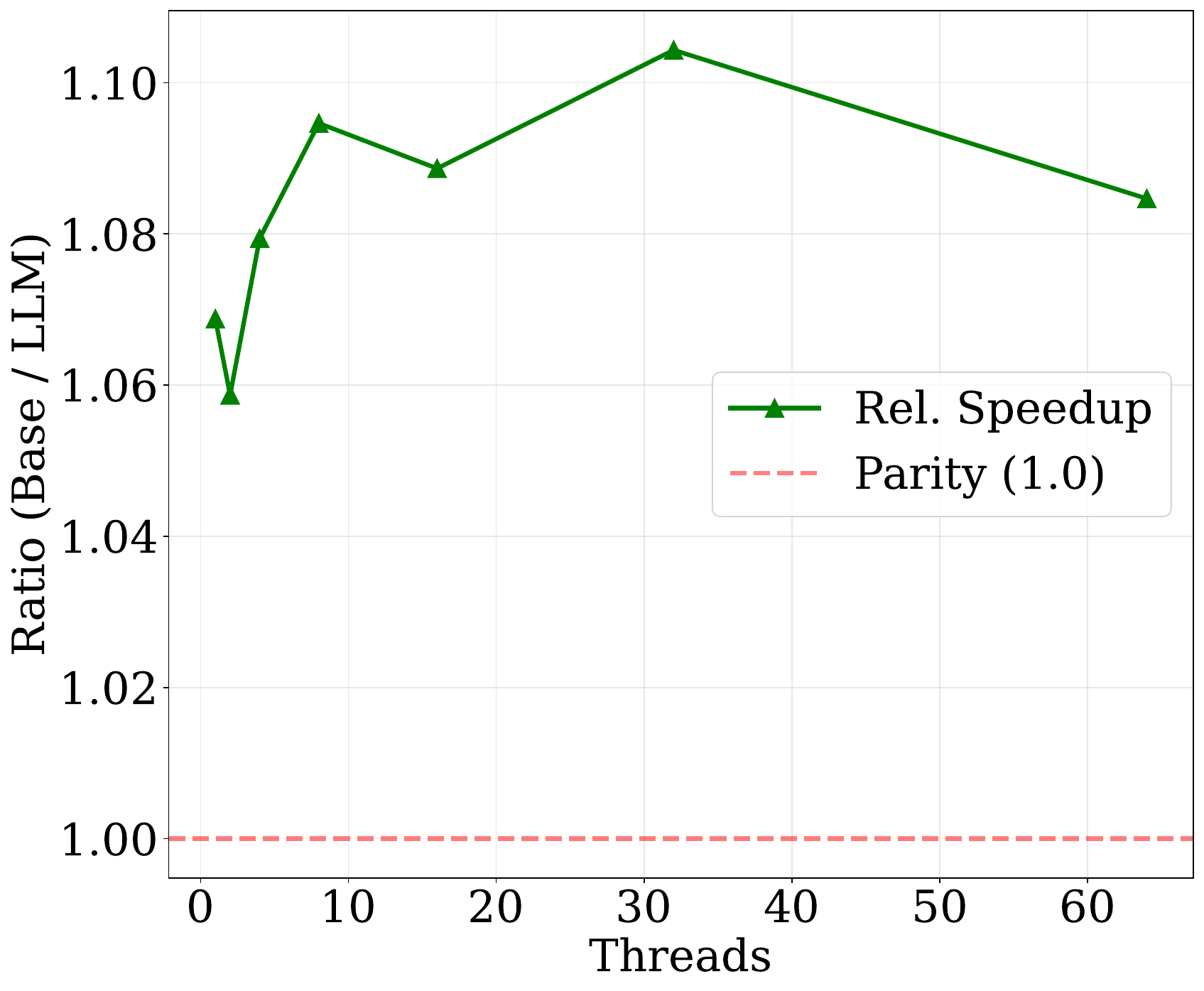}
        \caption{MSF Relative Speedup (Rust)}
    \end{subfigure}

    \begin{subfigure}[b]{0.3\columnwidth}
        \includegraphics[width=\linewidth]{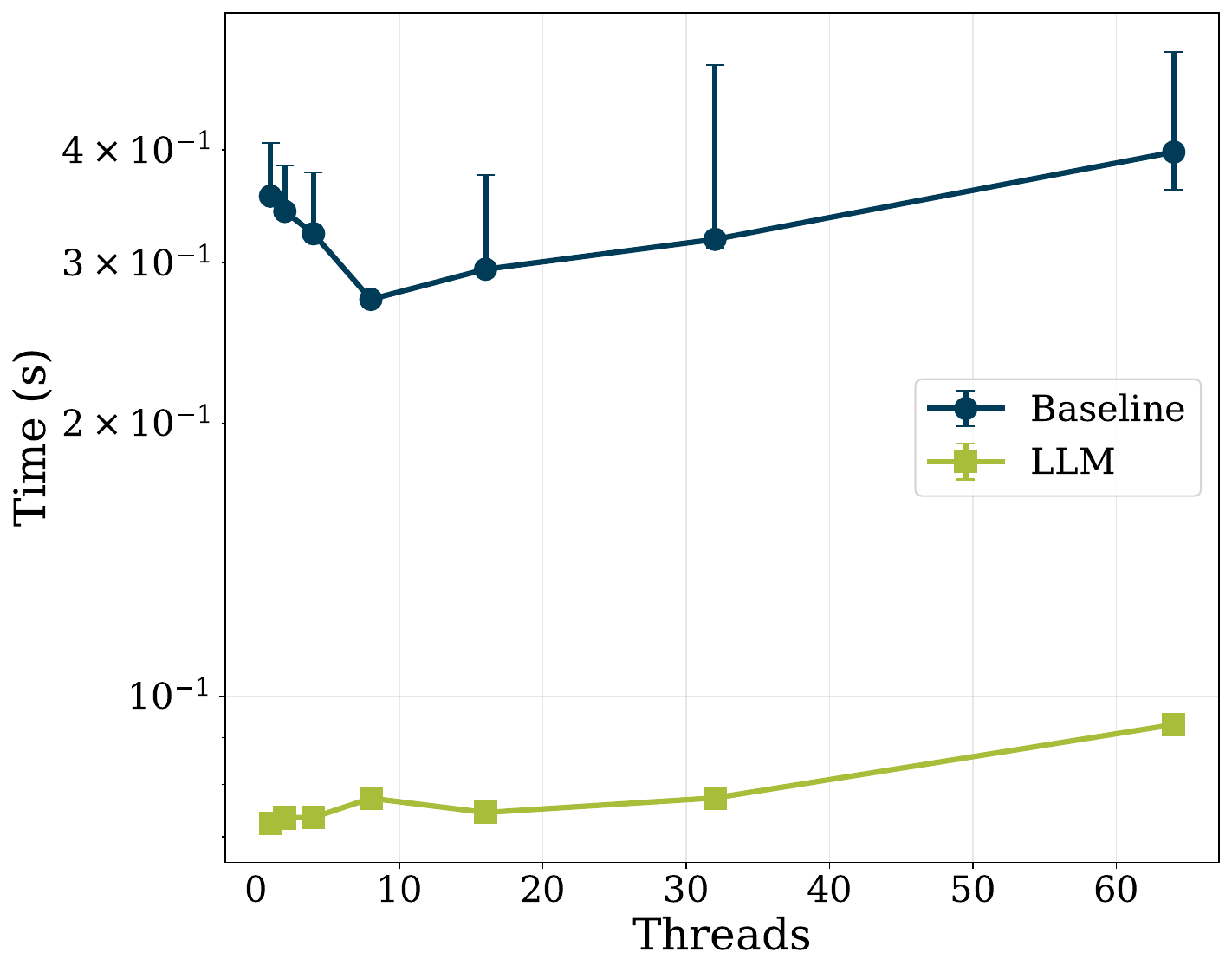}
        \caption{MIS (Rust)}
    \end{subfigure}
    \begin{subfigure}[b]{0.3\columnwidth}
        \includegraphics[width=\linewidth]{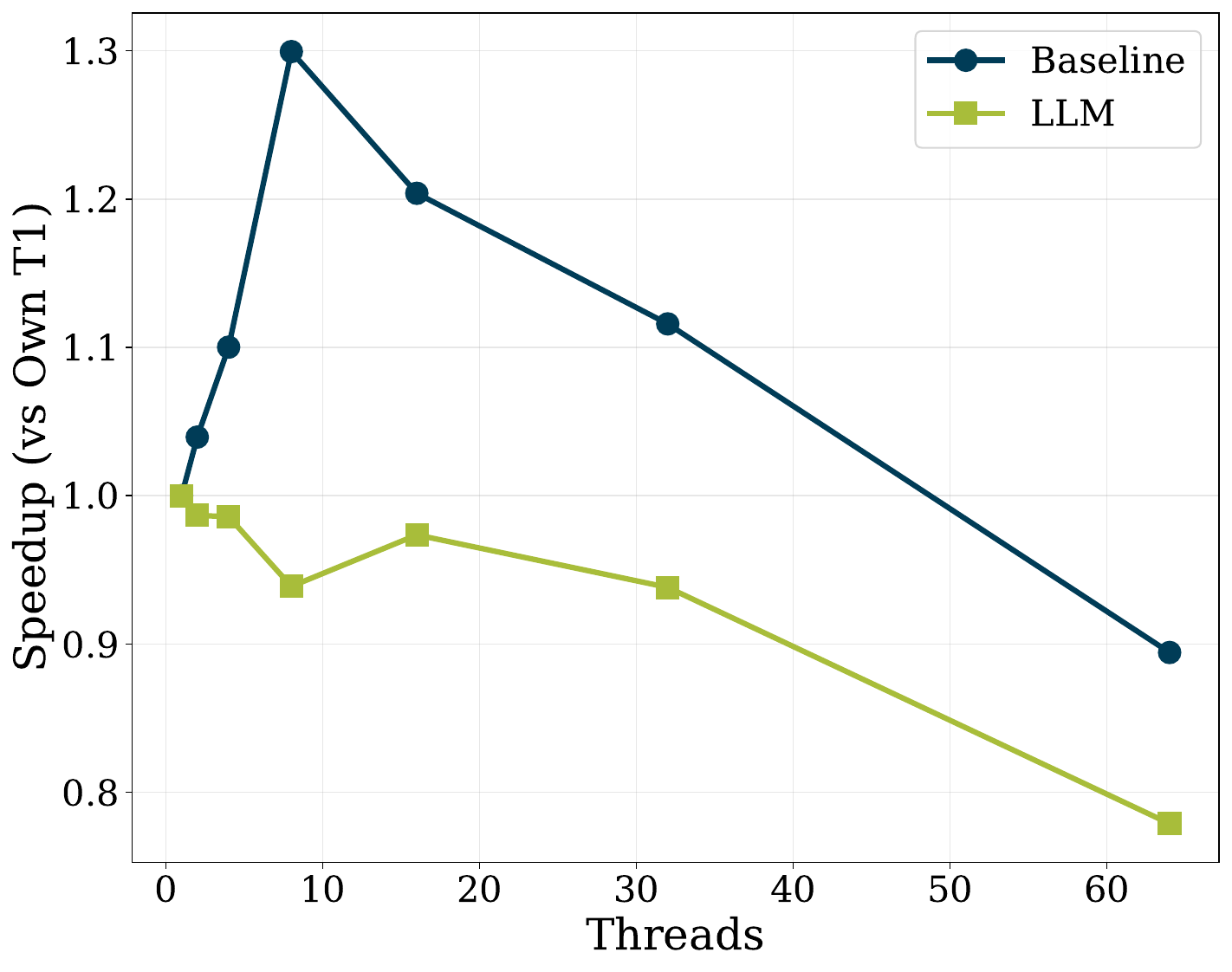}
        \caption{MIS (Rust)}
    \end{subfigure}
    \begin{subfigure}[b]{0.3\columnwidth}
        \includegraphics[width=\linewidth]{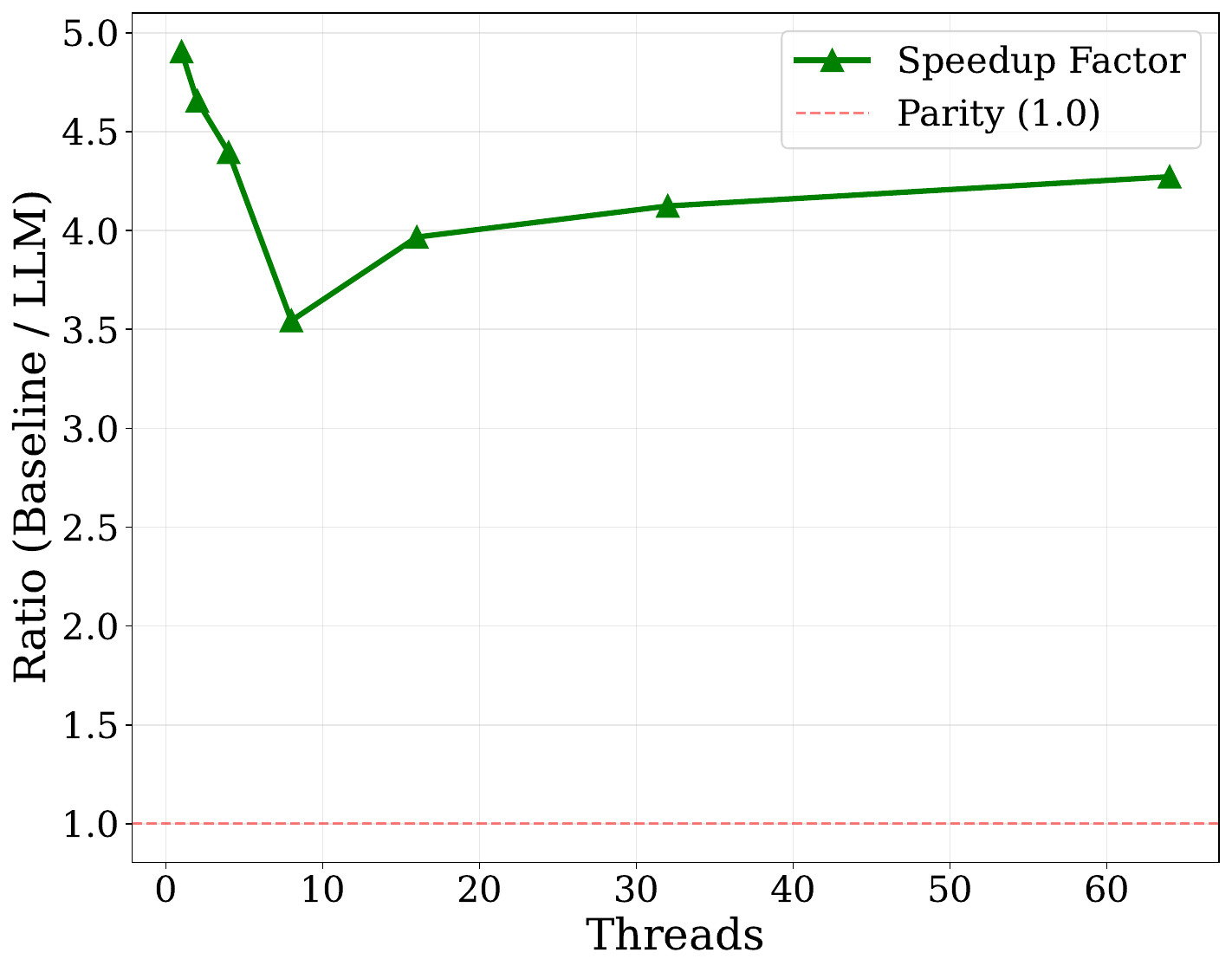}
        \caption{MIS (Rust)}
    \end{subfigure}
    
    \begin{subfigure}[b]{0.3\columnwidth}
        \includegraphics[width=\linewidth]{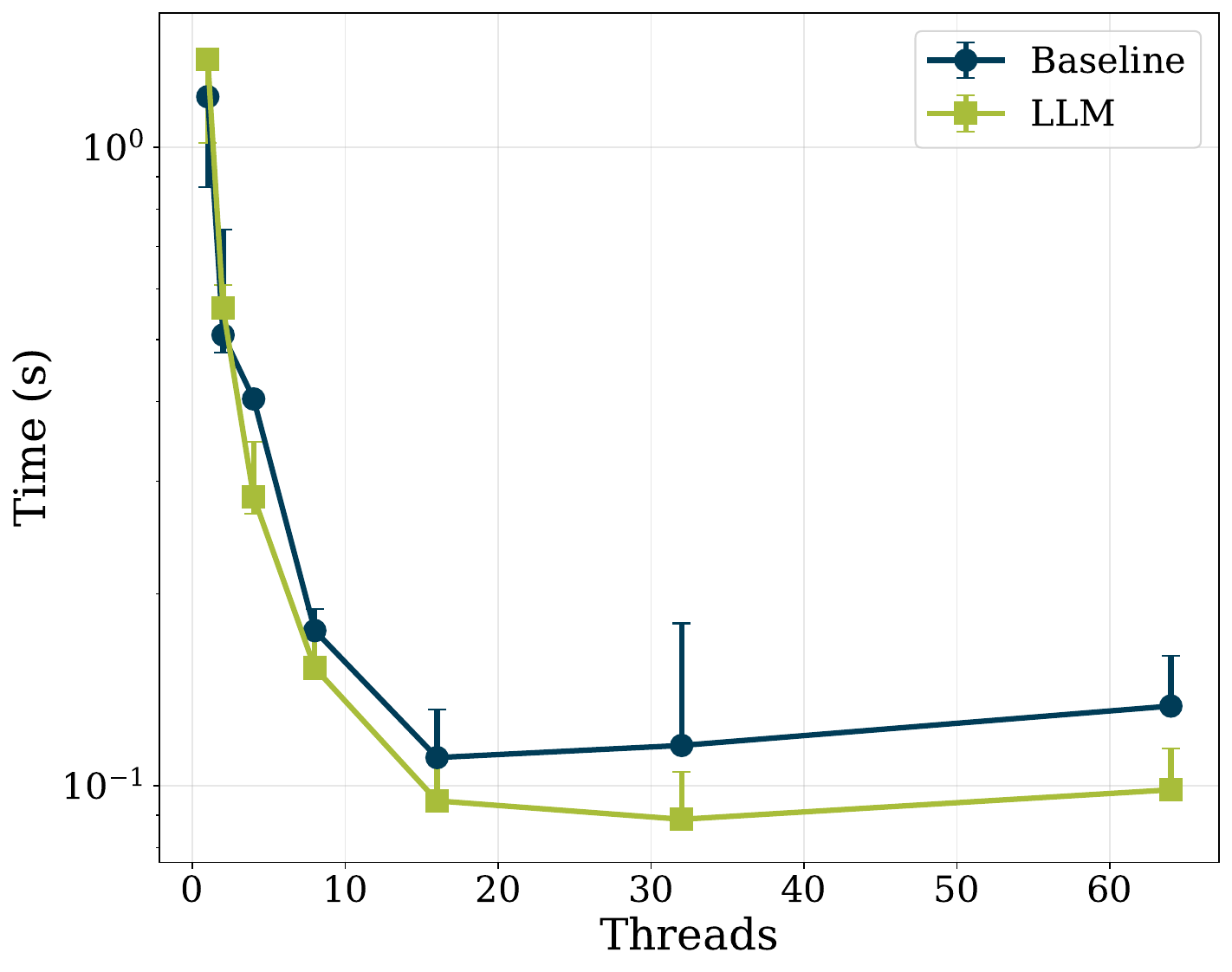}
        \caption{SP (Rust)}
    \end{subfigure}
    \begin{subfigure}[b]{0.3\columnwidth}
        \includegraphics[width=\linewidth]{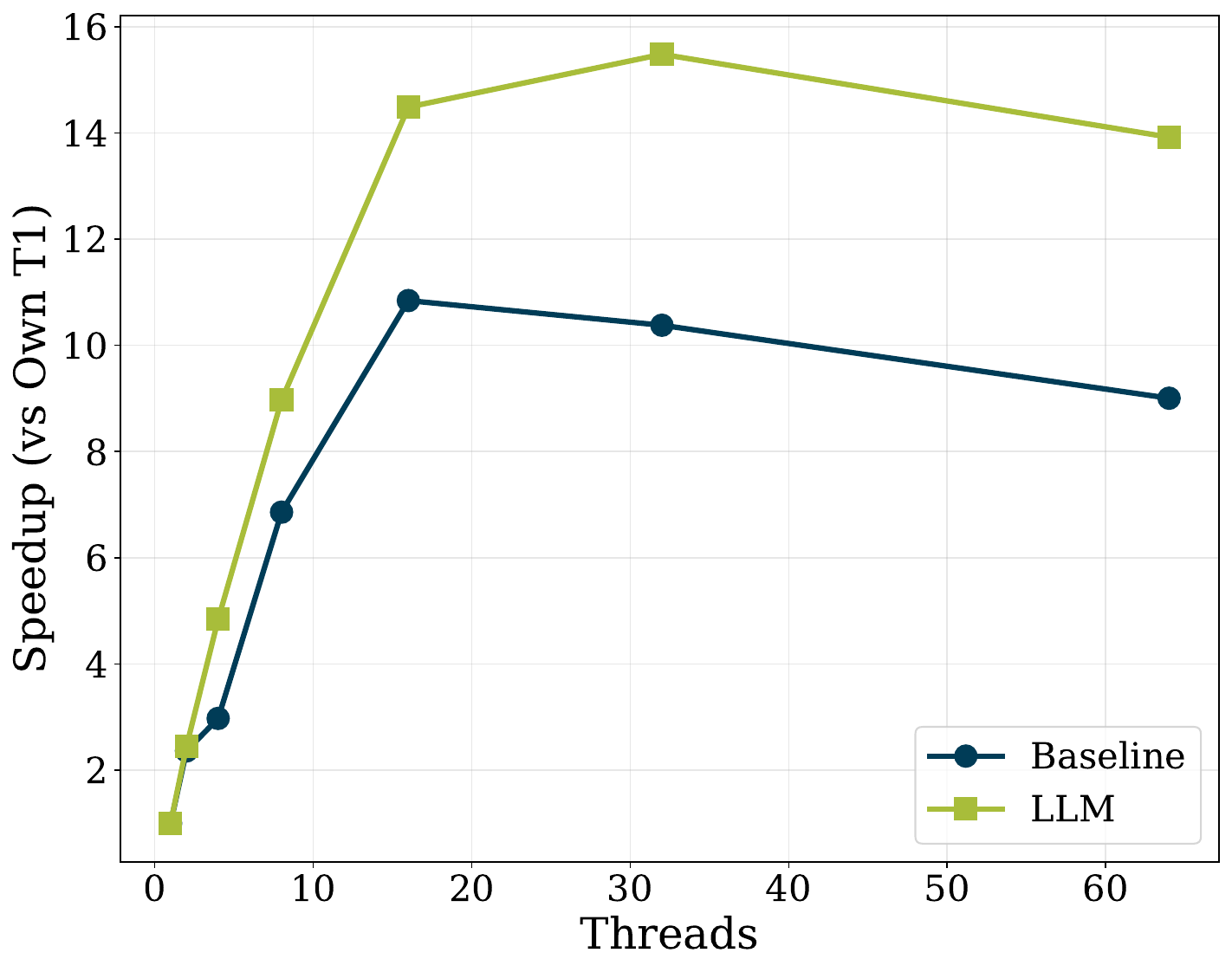}
        \caption{SP (Rust)}
    \end{subfigure}
    \begin{subfigure}[b]{0.3\columnwidth}
        \includegraphics[width=\linewidth]{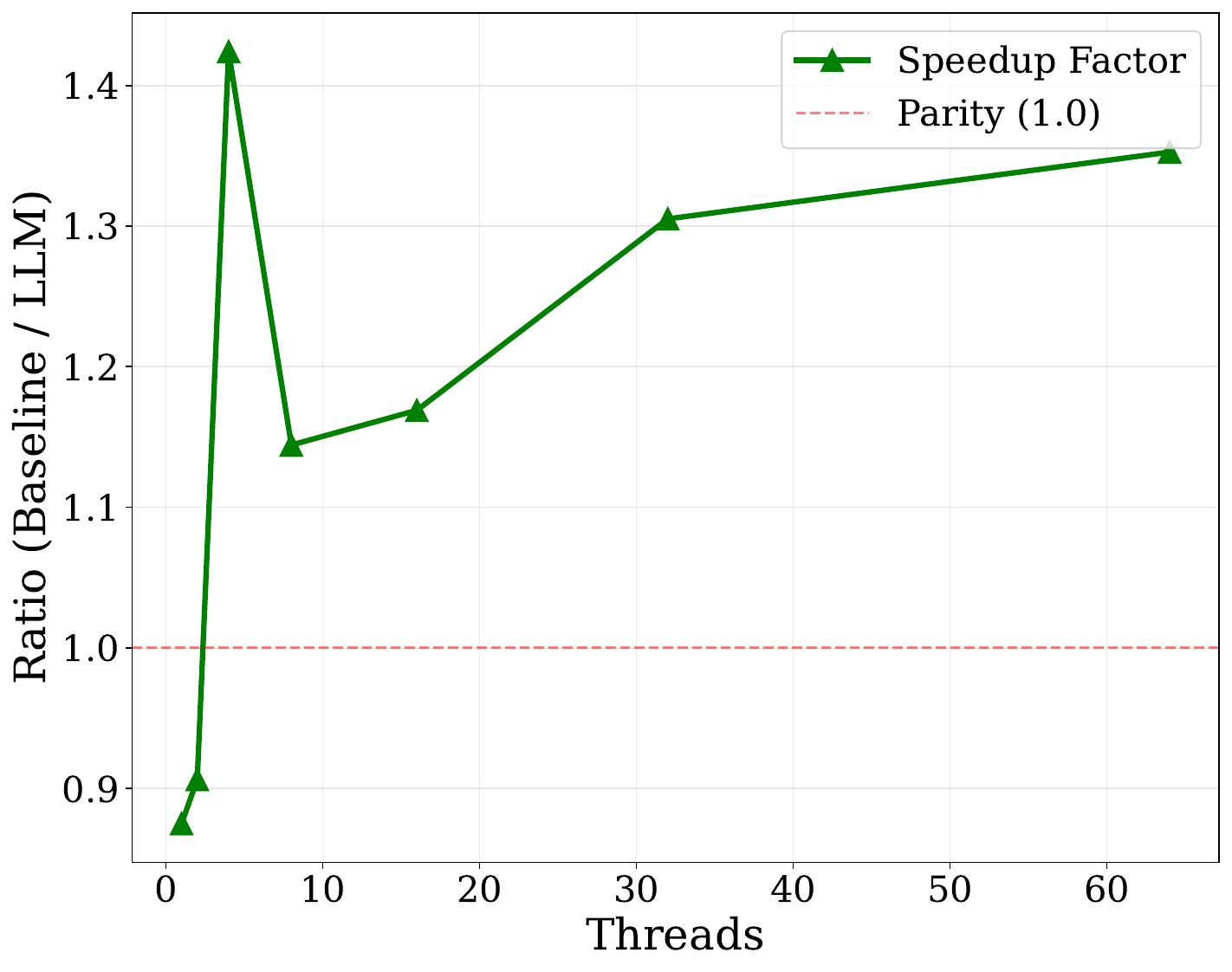}
        \caption{SP (Rust)}
    \end{subfigure}

    \begin{subfigure}[b]{0.32\columnwidth}
        \includegraphics[width=\linewidth]{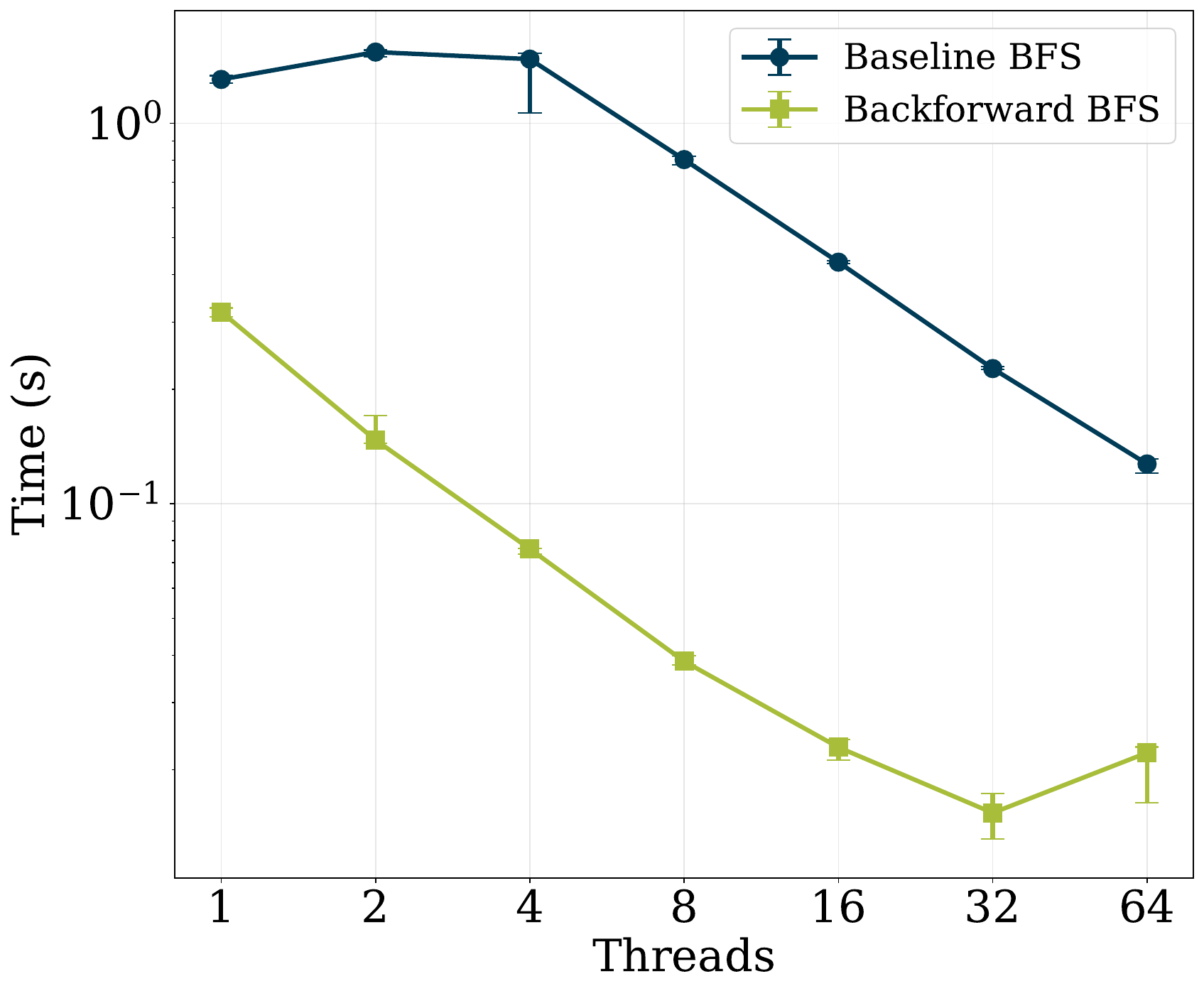}
        \caption{BFS (Rust)}
    \end{subfigure}
    \begin{subfigure}[b]{0.32\columnwidth}
        \includegraphics[width=\linewidth]{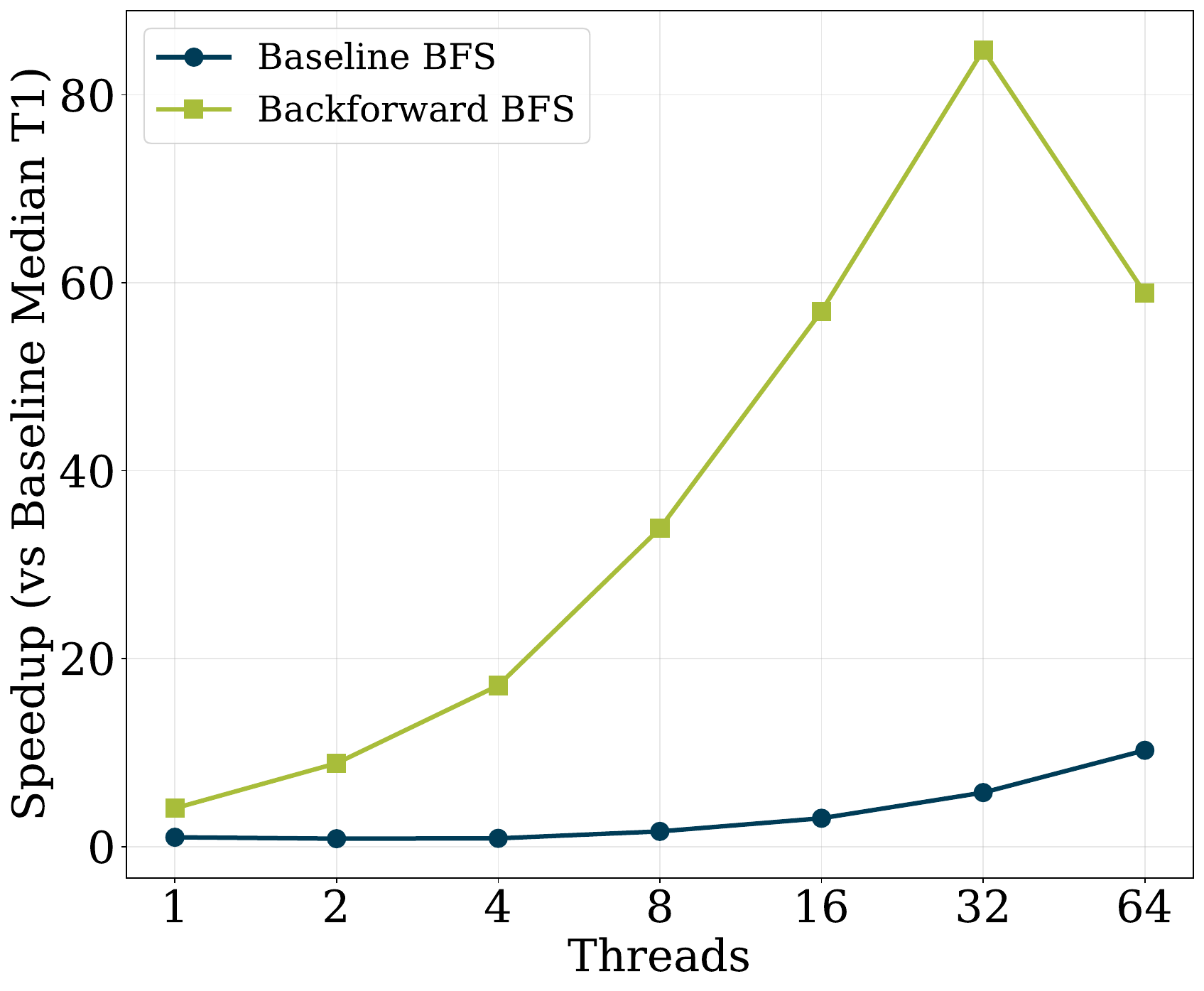}
        \caption{BFS (Rust)}
    \end{subfigure}
    \begin{subfigure}[b]{0.32\columnwidth}
        \includegraphics[width=\linewidth]{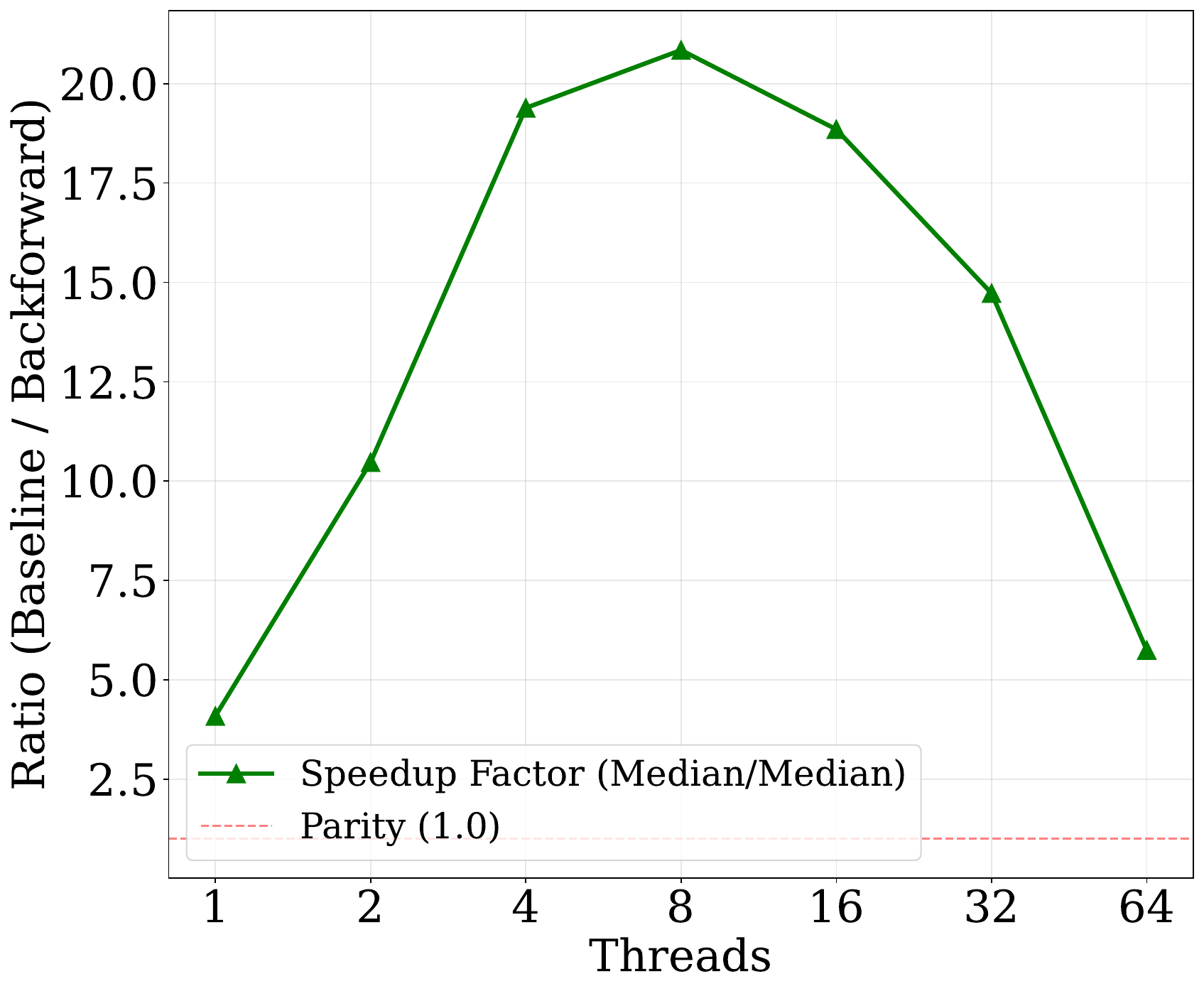}
        \caption{BFS (Rust)}
    \end{subfigure}
    \caption{Runtime and Scalability comparisons against expert Rust and C++ baselines. \sys solutions track or beat the scalability of hand-optimized code. In (m)-(o), \textit{BackForward BFS} specifically refers to the new BFS algorithm \sys generated, which uses a different method than the baseline implementation that uses multiqueue BFS.}
    \label{fig:rpb-pbbs-plots}
\end{figure}

A key contribution of this work is benchmarking against expert human code. \cref{fig:rpb-pbbs-plots} compares our best generated solutions against the PBBSBench (C++) and RPB (Rust) baselines. \sys solutions frequently match or exceed the human baselines. Figure \ref{fig:rpb-pbbs-plots} visualizes the runtime and scalability profiles.


\begin{figure}[h!]
    \centering
    \begin{minipage}[t]{0.48\textwidth}
        \begin{tcolorbox}[
            colback=gray!5, 
            colframe=gray!40, 
            title=\textbf{RPB (Unsafe Baseline)},
            fonttitle=\bfseries\small,
            boxrule=0.8pt,
            arc=2pt,
            left=2pt, right=2pt, top=2pt, bottom=2pt
        ]
        \begin{lstlisting}[language=Rust, basicstyle=\ttfamily\tiny, breaklines=true, escapechar=@]
#[cfg(not(feature = "openevolve"))]
pub fn maximal_matching(ea: &EdgeArray) -> Vec<DefInt> {
    let n = std::cmp::max(ea.num_rows, ea.num_cols);
    // ... setup ...
    let matched: Vec<bool> = (0..n).into_par_iter().map(|_| false).collect();
    let matched_ptr = matched.as_ptr() as usize; 

    let reserve = |i: usize| -> bool {
        let (u, v) = (ea[i].u as usize, ea[i].v as usize);
        // MIXED ACCESS: Reading safe memory, writing via raw pointer later.
        // Compiler may optimize this read incorrectly due to aliasing.
        if @\colorbox{red!15}{matched[u] || matched[v]}@ || u == v { false }
        else {
            rs[u].reserve(i as u32); rs[v].reserve(i as u32);
            true
        }
    };

    let commit = |i: usize| -> bool {
        // ... (check logic) ...
        if rs[v].check(i as u32) {
            rs[v].reset();
            if rs[u].check(i as u32) {
                unsafe { // UNSAFE WRITE via pointer alias
                    (matched_ptr as *mut bool).add(u).write(true);
                    (matched_ptr as *mut bool).add(v).write(true);
                }
                return true;
            }
        }
        // ...
    };

    (0..m).spec_for(reserve, commit, @\colorbox{red!15}{10}@, ...).unwrap();

    // DOUBLE PASS (Inefficient collection)
    let mut matching_idx = vec![];
    parlay::primitives::pack(
        &rs.par_iter().map(|r| r.get() as DefInt)@\colorbox{red!15}{.collect::<Vec<DefInt>>() }@,
        &rs.par_iter().map(|r| r.reserved())@\colorbox{red!15}{.collect::<Vec<bool>>() }@,
        &mut matching_idx
    );
    matching_idx
}
        \end{lstlisting}
        \end{tcolorbox}
    \end{minipage}
    \hfill
    \begin{minipage}[t]{0.48\textwidth}
        \begin{tcolorbox}[
            colback=gray!5, 
            colframe=gray!40, 
            title=\textbf{ParEVO (Optimized)},
            fonttitle=\bfseries\small,
            boxrule=0.8pt,
            arc=2pt,
            left=2pt, right=2pt, top=2pt, bottom=2pt
        ]
        \begin{lstlisting}[language=Rust, basicstyle=\ttfamily\tiny, breaklines=true, escapechar=@]
#[cfg(feature = "openevolve")]
pub fn maximal_matching(ea: &EdgeArray) -> Vec<DefInt> {
    let n = std::cmp::max(ea.num_rows, ea.num_cols);
    // ... setup ...
    let matched: Vec<bool> = (0..n).into_par_iter().map(|_| false).collect();
    let matched_ptr = matched.as_ptr() as usize;

    let reserve = |i: usize| -> bool {
        let (u, v) = (ea[i].u as usize, ea[i].v as usize);
        if u == v { return false; }
        
        // CONSISTENT RAW READ: Forces fresh memory access.
        unsafe {
            let mp = matched_ptr as *const bool;
            @\colorbox{green!15}{if *mp.add(u) || *mp.add(v) \{ return false; \}}@
        }
        rs[u].reserve(i as u32); rs[v].reserve(i as u32);
        true
    };

    let commit = |i: usize| -> bool {
        // ... (check logic) ...
        if rs[v].check(i as u32) {
            rs[v].reset();
            if rs[u].check(i as u32) {
                unsafe { // UNSAFE WRITE
                    let mp = matched_ptr as *mut bool;
                    *mp.add(u) = true; *mp.add(v) = true;
                }
                return true;
            }
        }
        // ...
    };

    // TUNED BLOCK SIZE
    (0..m).spec_for(reserve, commit, @\colorbox{green!15}{16}@, ...).unwrap();

    // SINGLE PASS (Optimized collection)
    let mut matching_idx = Vec::with_capacity(n / 2);
    let (values, flags): (Vec<DefInt>, Vec<bool>) = rs
        .par_iter()
        .map(|r| (r.get() as DefInt, r.reserved()))
        @\colorbox{green!15}{.unzip()}@;

    parlay::primitives::pack(&values, &flags, &mut matching_idx);
    matching_idx
}
        \end{lstlisting}
        \end{tcolorbox}
    \end{minipage}
    \caption{Code comparison for Maximal Matching. \textbf{Left (Baseline):} Uses mixed safe/unsafe access (potential aliasing bugs) and collects results using two separate passes (red highlights). \textbf{Right (ParEVO):} Uses consistent raw pointer access to ensure memory visibility, increases block granularity to 16, and uses a single-pass \texttt{unzip} for result collection (green highlights).}
    \label{fig:matching_comparison}
\end{figure}
\begin{figure}[h!]
    \centering
    \begin{minipage}[t]{0.48\textwidth}
        \begin{tcolorbox}[
            colback=gray!5, 
            colframe=gray!40, 
            title=\textbf{RPB (Unsafe Baseline)},
            fonttitle=\bfseries\small,
            boxrule=0.8pt,
            arc=2pt,
            left=2pt, right=2pt, top=2pt, bottom=2pt
        ]
        \begin{lstlisting}[language=Rust, basicstyle=\ttfamily\tiny, breaklines=true, escapechar=@]
#[cfg(not(feature="openevolve"))]
pub fn minimum_spanning_forest(wea: &WghEdgeArray, dest: &mut Vec<DefInt>) {
    // ... Initialization ...
    let rs: Vec<Reservation> = (0..n).map(|_| Reservation::new()).collect();
    
    // Raw pointers for some structures only
    let _uf_ptr = &uf as *const _ as usize; 
    let _iwea_ptr = iwea.as_ptr() as usize;

    let reserve = |i: usize| {
        // Unsafe access to Edge List
        let e = unsafe {
            (_iwea_ptr as *mut IndexedEdge).add(i).as_mut()@\colorbox{red!15}{.unwrap()}@ // Runtime check
        };
        // Unsafe access to UnionFind
        let luf = unsafe { (_uf_ptr as *mut UnionFind).as_mut()@\colorbox{red!15}{.unwrap()}@ };
        
        e.u = luf.find(e.u as DefIntS) as DefInt;
        e.v = luf.find(e.v as DefIntS) as DefInt;
        
        if e.u != e.v {
            // STANDARD INDEXING: Incurs bounds checking overhead
            @\colorbox{red!15}{rs[e.v as usize]}@.reserve(i as DefInt);
            @\colorbox{red!15}{rs[e.u as usize]}@.reserve(i as DefInt);
            true
        } else { false }
    };
    
    // ... Commit logic similar to above ...
    (0..iwea.len()).spec_for(reserve, commit, ...);
}
        \end{lstlisting}
        \end{tcolorbox}
    \end{minipage}
    \hfill
    \begin{minipage}[t]{0.48\textwidth}
        \begin{tcolorbox}[
            colback=gray!5, 
            colframe=gray!40, 
            title=\textbf{ParEVO (Optimized)},
            fonttitle=\bfseries\small,
            boxrule=0.8pt,
            arc=2pt,
            left=2pt, right=2pt, top=2pt, bottom=2pt
        ]
        \begin{lstlisting}[language=Rust, basicstyle=\ttfamily\tiny, breaklines=true, escapechar=@]
#[cfg(feature="openevolve")]
pub fn minimum_spanning_forest(wea: &WghEdgeArray, dest: &mut Vec<DefInt>) {
    // ... Initialization ...
    let rs: Vec<Reservation> = (0..n).into_par_iter().map(|_| Reservation::new()).collect();
    
    // Raw pointers for EVERYTHING (including Reservation array)
    let _rs_ptr = rs.as_ptr() as usize;
    let _uf_ptr = &uf as *const _ as usize;
    let _iwea_ptr = iwea.as_ptr() as usize;

    let reserve = |i: usize| {
        unsafe {
            // UNCHECKED UNWRAP: Eliminates null checks
            let e = (_iwea_ptr as *mut IndexedEdge).add(i)
                .as_mut()@\colorbox{green!15}{.unwrap\_unchecked()}@;
            let luf = (_uf_ptr as *mut UnionFind).as_mut()
                @\colorbox{green!15}{.unwrap\_unchecked()}@;
            
            e.u = luf.find(e.u as DefIntS) as DefInt;
            e.v = luf.find(e.v as DefIntS) as DefInt;
            
            if e.u != e.v {
                // POINTER ARITHMETIC: Eliminates bounds checks
                let rv = (_rs_ptr as *const Reservation)@\colorbox{green!15}{.add(e.v as usize)}@
                    .as_ref().unwrap_unchecked();
                let ru = (_rs_ptr as *const Reservation)@\colorbox{green!15}{.add(e.u as usize)}@
                    .as_ref().unwrap_unchecked();
                
                rv.reserve(i as DefInt);
                ru.reserve(i as DefInt);
            }
            true
        }
    };
    (0..iwea.len()).spec_for(reserve, commit, ...);
}
        \end{lstlisting}
        \end{tcolorbox}
    \end{minipage}
    \caption{Code comparison for Minimum Spanning Forest (MSF). \textbf{Left (Baseline):} Uses standard indexing for the reservation array (incurring bounds checks) and standard \texttt{unwrap()} (incurring branch checks), highlighted in red. \textbf{Right (ParEVO):} Adopts a ``Maximal Unsafe" strategy, converting all data structures to raw pointers. It uses \texttt{unwrap\_unchecked()} and pointer arithmetic (\texttt{.add()}) to eliminate all runtime safety checks, highlighted in green. This relies on the assumption that edge indices are always valid, allowing ParEVO to trade runtime safety checks for improved performance.}
    \label{fig:msf_comparison}
\end{figure}

\begin{figure}[h!]
    \centering
    \begin{minipage}[t]{0.48\textwidth}
        \begin{tcolorbox}[
            colback=gray!5, 
            colframe=gray!40, 
            title=\textbf{RPB (Unsafe Baseline)},
            fonttitle=\bfseries\small,
            boxrule=0.8pt,
            arc=2pt,
            left=2pt, right=2pt, top=2pt, bottom=2pt
        ]
        \begin{lstlisting}[language=Rust, basicstyle=\ttfamily\tiny, breaklines=true, escapechar=@]
#[cfg(not(feature = "openevolve"))]
pub fn maximal_independent_set(g: &Graph) -> Vec<u8> {
    let n = g.n;
    // UNSAFE: Standard Vec used for concurrent access
    let flags: Vec<u8> = (0..n).into_par_iter()
        .map(|_| 0)@\colorbox{red!15}{.collect()}@; // Standard allocation
    let flags_ptr = flags.as_ptr() as usize;

    let reserve = |i: usize, s: &mut MISState| -> bool {
        s.flag = 1;
        let v = g.index(i);
        for j in 0..v.degree {
            let ngh = v.neighbors[j] as usize;
            if ngh < i {
                // DATA RACE: Reading mutable memory without atomics
                // Compiler may optimize incorrectly; Undefined Behavior
                let f = @\colorbox{red!15}{flags[ngh]}@; 
                if f == 1 { s.flag = 2; return true; }
                else if f == 0 { s.flag = 0; }
            }
        }
        true
    };

    let commit = |i: usize, s: &mut MISState| -> bool {
        // UNSAFE POINTER WRITE: Bypassing borrow checker
        unsafe { (flags_ptr as *mut u8).add(i)@\colorbox{red!15}{.write(s.flag)}@; }
        s.flag > 0
    };

    (0..n).stateful_spec_for(
        reserve, commit, MISState { flag: 0 },
        @\colorbox{red!15}{20}@, Some(64), Some(256) // Small granularity
    ).expect("failed speculative for");

    return flags;
}
        \end{lstlisting}
        \end{tcolorbox}
    \end{minipage}
    \hfill
    \begin{minipage}[t]{0.48\textwidth}
        \begin{tcolorbox}[
            colback=gray!5, 
            colframe=gray!40, 
            title=\textbf{ParEVO (Optimized)},
            fonttitle=\bfseries\small,
            boxrule=0.8pt,
            arc=2pt,
            left=2pt, right=2pt, top=2pt, bottom=2pt
        ]
        \begin{lstlisting}[language=Rust, basicstyle=\ttfamily\tiny, breaklines=true, escapechar=@]
#[cfg(feature = "openevolve")]
pub fn maximal_independent_set(g: &Graph) -> Vec<u8> {
    let n = g.n;
    // PARALLEL ATOMICS: Safe concurrent access
    let flags: Vec<AtomicU8> = (0..n).into_par_iter()
        .map(|_| AtomicU8::new(0))@\colorbox{green!15}{.collect()}@;
    let flags_slice = &flags[..];

    let reserve = |i: usize, s: &mut MISState| -> bool {
        let v = g.index(i);
        let mut waiting = false;
        for &ngh in v.neighbors {
            let ngh = ngh as usize;
            if ngh < i {
                // SAFE ATOMIC LOAD: Correct synchronization
                let f = unsafe { 
                    flags_slice.get_unchecked(ngh)@\colorbox{green!15}{.load(Relaxed)}@ 
                };
                if f == 1 { s.flag = 2; return true; }
                if f == 0 { waiting = true; }
            }
        }
        s.flag = if waiting { 0 } else { 1 };
        true
    };

    let commit = |i: usize, s: &mut MISState| -> bool {
        if s.flag > 0 {
            // SAFE ATOMIC STORE
            unsafe { flags_slice.get_unchecked(i)@\colorbox{green!15}{.store(s.flag, Relaxed)}@; }
            true
        } else { false }
    };

    (0..n).stateful_spec_for(
        reserve, commit, MISState { flag: 0 },
        @\colorbox{green!15}{256}@, None, None // Larger granularity
    ).expect("failed speculative for");

    // ZERO-COPY TRANSFORMATION: AtomicU8 -> u8
    unsafe {
        let mut v = std::mem::ManuallyDrop::new(flags);
        @\colorbox{green!15}{Vec::from\_raw\_parts}@(v.as_mut_ptr() as *mut u8, v.len(), v.capacity())
    }
}
        \end{lstlisting}
        \end{tcolorbox}
    \end{minipage}
    \caption{Code comparison for Maximal Independent Set (MIS). \textbf{Left (Baseline):} Uses unsafe standard \texttt{Vec<u8>} (red), causing undefined behavior (data races) during reads and writing via raw pointers. It uses a small block size (20). \textbf{Right (ParEVO):} Uses \texttt{Vec<AtomicU8>} (green) for correct synchronization using \texttt{Relaxed} ordering. It optimizes throughput with a larger block size (256) and employs a zero-copy cast to convert the atomic vector back to a standard vector at the end.}
    \label{fig:mis_comparison}
\end{figure}

\begin{figure}[h!]
    \centering
    \begin{minipage}[t]{0.48\textwidth}
        \begin{tcolorbox}[
            colback=gray!5, 
            colframe=gray!40, 
            title=\textbf{RPB (Unsafe Baseline)},
            fonttitle=\bfseries\small,
            boxrule=0.8pt,
            arc=2pt,
            left=2pt, right=2pt, top=2pt, bottom=2pt
        ]
        \begin{lstlisting}[language=Rust, basicstyle=\ttfamily\tiny, breaklines=true, escapechar=@]
#[cfg(not(feature = "openevolve"))]
pub fn spanning_forest(ea: &EdgeArray) -> Vec<u32> {
    let n = ea.num_rows;
    // NON-ATOMIC & FRESH ALLOCATION
    let uf = UnionFind::new(n); 
    let uf_ptr = &uf as *const UnionFind as usize;
    
    // HEAVY ALLOCATION: Creates new Vec every call
    let rs: Vec<Reservation> = (0..n).into_par_iter()
        .map(|_| Reservation::new())@\colorbox{red!15}{.collect()}@;

    let reserve = |i: usize, s: &mut SFState| -> bool {
        let e = &ea[i]; // Bounds checked
        unsafe {
            // UNSAFE DEREF + RUNTIME CHECK (unwrap)
            s.u = (uf_ptr as *mut UnionFind).as_mut()@\colorbox{red!15}{.unwrap()}@.find(e.u as i32);
            s.v = (uf_ptr as *mut UnionFind).as_mut()@\colorbox{red!15}{.unwrap()}@.find(e.v as i32);
        }
        if s.u > s.v { swap(&mut s.u, &mut s.v); }
        
        if s.u != s.v {
            // BOUNDS CHECKED indexing
            @\colorbox{red!15}{rs[s.v as usize]}@.reserve(i as DefInt);
            true
        } else { false }
    };

    let commit = |i: usize, s: &mut SFState| -> bool {
        if rs[s.v as usize].check(i as DefInt) {
            unsafe {
                (uf_ptr as *mut UnionFind).as_mut().unwrap().link(s.v, s.u);
            }
            true
        } else { false }
    };

    (0..ea.non_zeros).stateful_spec_for(
        reserve, commit, SFState { u: -1, v: -1 }, 
        100, Some(1024), Some(4096)
    ).expect("failed speculative for");

    rs.into_par_iter().filter_map(|r| /*...*/).collect()
}
        \end{lstlisting}
        \end{tcolorbox}
    \end{minipage}
    \hfill
    \begin{minipage}[t]{0.48\textwidth}
        \begin{tcolorbox}[
            colback=gray!5, 
            colframe=gray!40, 
            title=\textbf{ParEVO (Optimized)},
            fonttitle=\bfseries\small,
            boxrule=0.8pt,
            arc=2pt,
            left=2pt, right=2pt, top=2pt, bottom=2pt
        ]
        \begin{lstlisting}[language=Rust, basicstyle=\ttfamily\tiny, breaklines=true, escapechar=@]
#[cfg(feature = "openevolve")]
pub fn spanning_forest(ea: &EdgeArray, rs_cache: &mut Option<Vec<Reservation>>) -> Vec<u32> {
    let n = ea.num_rows;
    let uf = AtomicUnionFind::new(n); // Safe Atomics

    // MEMORY RECYCLING: Reuses vector to skip allocation
    let mut rs = if let Some(mut vec) = @\colorbox{green!15}{rs\_cache.take()}@ {
        if vec.len() == n {
            vec.par_iter_mut().for_each(|r| *r = Reservation::new());
            vec
        } else { (0..n).into_par_iter().map(|_| Reservation::new()).collect() }
    } else { (0..n).into_par_iter().map(|_| Reservation::new()).collect() };

    let es = &ea.es;
    let reserve = |i: usize, s: &mut SFState| -> bool {
        // ZERO OVERHEAD ACCESS
        let e = unsafe { @\colorbox{green!15}{es.get\_unchecked(i)}@ }; 
        if e.u == e.v { return false; }

        let u_val = uf.find(e.u as i32);
        let v_val = uf.find(e.v as i32);
        if u_val == v_val { return false; }

        let (u, v) = if u_val > v_val { (v_val, u_val) } else { (u_val, v_val) };
        s.u = u; s.v = v;
        
        // UNCHECKED INDEXING
        unsafe { @\colorbox{green!15}{rs.get\_unchecked(s.v as usize)}@.reserve(i as DefInt); }
        true
    };

    let commit = |i: usize, s: &mut SFState| -> bool {
        unsafe {
            if rs.get_unchecked(s.v as usize).check(i as DefInt) {
                uf.link(s.v, s.u);
                true
            } else { false }
        }
    };
    // ... spec_for execution ...
    let res = rs.par_iter().filter_map(|r| /*...*/).collect();
    
    // RECYCLE: Return vector to cache
    @\colorbox{green!15}{*rs\_cache = Some(rs)}@;
    res
}
        \end{lstlisting}
        \end{tcolorbox}
    \end{minipage}
    \caption{Code comparison for Spanning Forest. \textbf{Left (Baseline):} Performs a fresh allocation for the reservation array on every call (red) and uses checked indexing/unwrapping inside the hot loop. \textbf{Right (ParEVO):} Implements a memory recycling mechanism via \texttt{rs\_cache} (green) to reuse the large reservation vector across calls. It also employs \texttt{get\_unchecked} and \texttt{AtomicUnionFind} to eliminate bounds checking and pointer dereference overheads.}
    \label{fig:sf_comparison}
\end{figure}

\begin{figure}[h!]
    \centering
    \begin{minipage}[t]{0.48\textwidth}
        \begin{tcolorbox}[
            colback=gray!5, 
            colframe=gray!40, 
            title=\textbf{RPB (Unsafe Baseline)},
            fonttitle=\bfseries\small,
            boxrule=0.8pt,
            arc=2pt,
            left=2pt, right=2pt, top=2pt, bottom=2pt
        ]
        \begin{lstlisting}[language=Rust, basicstyle=\ttfamily\tiny, breaklines=true, escapechar=@]
fn process_node(val: ValType, graph: &Graph, data: &SharedData,
    pq: &MultiQueue<ValType> // Concurrent Queue Overhead
) {
    let (dist, src) = (val.0, val.1);
    if data.shortest_distance[src].load(Ordering::Relaxed) < dist { return; }

    let new_distance = dist + 1;
    for i in graph.nodes[src]..graph.nodes[src + 1] {
        let target = graph.edges[i].target;
        let mut old_distance = data.shortest_distance[target].load(Ordering::Relaxed);

        // HOT LOOP: High Contention Point
        while new_distance < old_distance {
            // HEAVY SYNC: Compare-And-Swap loop
            match data.shortest_distance[target]@\colorbox{red!15}{.compare\_exchange\_weak}@(
                old_distance, new_distance,
                @\colorbox{red!15}{Ordering::SeqCst}@, // Strong Ordering
                Ordering::Relaxed,
            ) {
                Ok(_) => {
                    // QUEUE PUSH: Locking overhead
                    @\colorbox{red!15}{pq.push(ValType(new\_distance, target))}@;
                    break;
                },
                Err(x) => old_distance = x, // Retry on failure
            }
        }
    }
}
        \end{lstlisting}
        \end{tcolorbox}
    \end{minipage}
    \hfill
    \begin{minipage}[t]{0.48\textwidth}
        \begin{tcolorbox}[
            colback=gray!5, 
            colframe=gray!40, 
            title=\textbf{ParEVO (Optimized)},
            fonttitle=\bfseries\small,
            boxrule=0.8pt,
            arc=2pt,
            left=2pt, right=2pt, top=2pt, bottom=2pt
        ]
        \begin{lstlisting}[language=Rust, basicstyle=\ttfamily\tiny, breaklines=true, escapechar=@]
impl<'a, Fa, Cond> EdgeMap<'a, Fa, Cond> {
    pub fn apply(&self, frontier: VertexSubset) -> VertexSubset {
        let n = self.g_out.num_nodes();
        let m = self.g_out.num_edges();

        // HEURISTIC: Check frontier density
        if frontier.is_sparse {
            let l = frontier.sparse.len();
            // Calculate exact workload
            let out_degree = delayed::reduce_map(&fview, |v| degree(self.g_out, v));

            // THRESHOLD: Switch based on Edge Count vs Vertices
            if l + out_degree > m / 20 {
                // PULL PHASE (Dense Optimization)
                // Scans unvisited nodes to find ANY parent (Early Exit)
                // Uses Transpose Graph (g_in) significantly reducing checks
                let next_dense = @\colorbox{green!15}{edge\_map\_dense(self.g\_in, ...)}@;
                VertexSubset::from_dense(next_dense)
            } else {
                // PUSH PHASE (Sparse Standard)
                // Traditional BFS only for small frontiers
                let next_sparse = @\colorbox{green!15}{edge\_map\_sparse(self.g\_out, ...)}@;
                VertexSubset::from_sparse(next_sparse)
            }
        } else {
            // ... Dense -> Dense or Dense -> Sparse logic ...
        }
    }
}
        \end{lstlisting}
        \end{tcolorbox}
    \end{minipage}
    \caption{Code comparison for BFS. \textbf{Left (Baseline):} Uses a standard asynchronous approach where every edge relaxation requires a CAS loop and a queue push (red), leading to high contention on scale-free graphs. \textbf{Right (ParEVO):} Implements Direction-Optimizing BFS (Ligra-style). It dynamically switches between "Push" (Sparse) and "Pull" (Dense) modes based on the frontier density (green), drastically reducing edge checks during the heavy middle levels of the traversal.}
    \label{fig:bfs_comparison}
\end{figure}


\subsection{Case Study: The Safety vs. Performance Trade-off}
\begin{figure}[h!]
    \centering
    \begin{subfigure}[b]{0.48\columnwidth}
        \includegraphics[width=\linewidth]{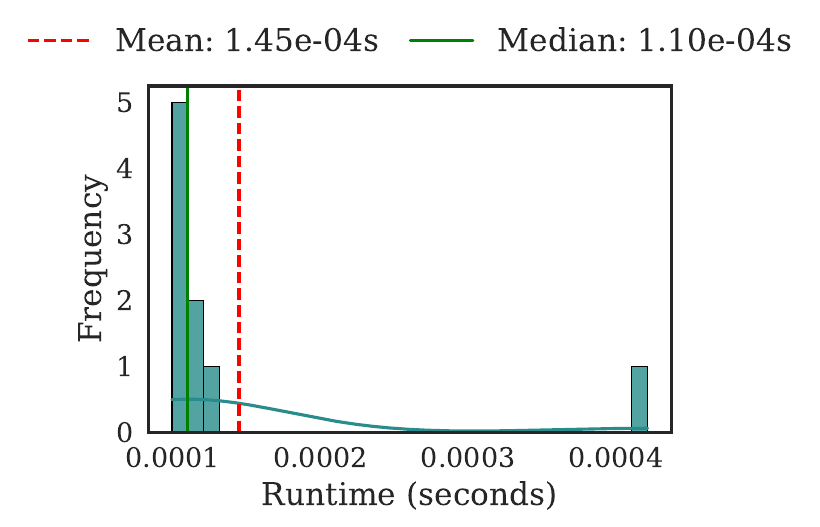}
        \caption{Base Model Runtime Distribution}
    \end{subfigure}
    \begin{subfigure}[b]{0.48\columnwidth}
        \includegraphics[width=\linewidth]{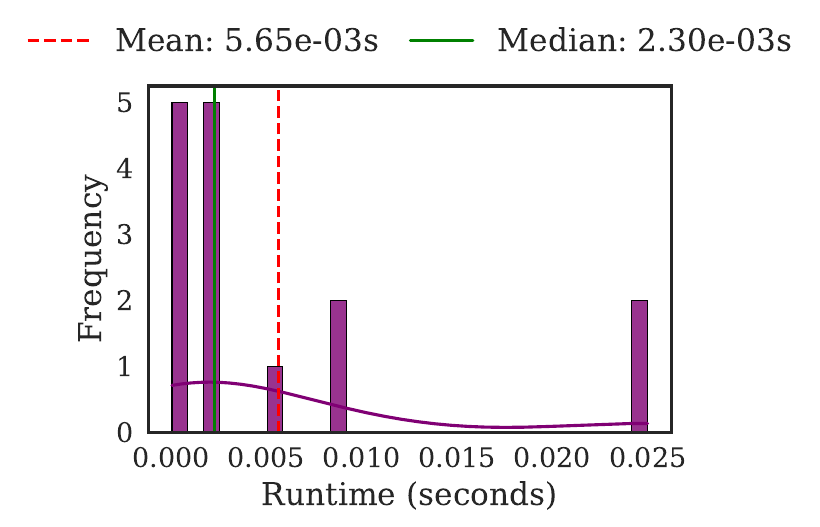}
        \caption{Fine-tuned Runtime Distribution}
    \end{subfigure}
    \caption{Runtime Histograms for Graph Shortest Path. The fine-tuned model (b) exhibits tighter variance (reliability) but a higher median runtime due to overhead from safety-focused primitives.}
    \label{fig:shortest-path-hist}
\end{figure}

A deeper analysis of the Graph Shortest Path problem reveals a subtle trade-off introduced by fine-tuning. As shown in \cref{fig:shortest-path-hist}, the base model produces a wide variance of runtimes, occasionally hitting a very fast (but risky) solution using atomic operations. The fine-tuned \sys model produces highly consistent but slightly slower code, preferring safe high-level primitives (like `parlay::unique') over raw memory manipulation. The detailed code samples are shown in \cref{fig:pareval_19_gen_comparison}.

\begin{figure}[h!]
    \centering
    \begin{minipage}[t]{0.48\textwidth}
        \begin{tcolorbox}[
            colback=gray!5, 
            colframe=gray!40, 
            title=\textbf{Baseline: Gemini-2.5-Pro},
            fonttitle=\bfseries\small,
            boxrule=0.8pt,
            arc=2pt,
            left=2pt, right=2pt, top=2pt, bottom=2pt
        ]
        \textbf{Runtime:} 0.0000775394
        \begin{lstlisting}[language=C++, basicstyle=\ttfamily\tiny, breaklines=true, escapechar=@]
{
  if (source == dest) { return 0; }

  // ATOMIC DISTANCE ARRAY
  // Allows safe concurrent updates without locks
  auto distances = @\colorbox{green!15}{parlay::sequence<std::atomic<int>>(N)}@;
  parlay::parallel_for(0, N, [&](size_t i) {
    distances[i].store(-1);
  });
  distances[source].store(0);

  parlay::sequence<int> frontier(1, source);
  int current_dist = 0;

  while (!frontier.empty()) {
    if (distances[dest].load() != -1) return distances[dest].load();
    current_dist++;

    // Parallel neighbor generation
    auto all_neighbors = parlay::flatten(parlay::map(frontier, [&](int u) {
      auto indices = parlay::iota(N);
      return parlay::filter(indices, [&](int v) {
        return A[static_cast<size_t>(u) * N + v] == 1;
      });
    }));

    auto next_frontier = parlay::filter(all_neighbors, [&](int v) {
      int expected = -1;
      // ATOMIC CAS: Thread-safe claim of visitation
      return @\colorbox{green!15}{distances[v].compare\_exchange\_strong(expected, current\_dist)}@;
    });

    frontier = std::move(next_frontier);
  }
  return -1;
}
        \end{lstlisting}
        \end{tcolorbox}
    \end{minipage}
    \hfill
    \begin{minipage}[t]{0.48\textwidth}
        \begin{tcolorbox}[
            colback=gray!5, 
            colframe=gray!40, 
            title=\textbf{Finetuned: Gemini-2.5-Parlay},
            fonttitle=\bfseries\small,
            boxrule=0.8pt,
            arc=2pt,
            left=2pt, right=2pt, top=2pt, bottom=2pt
        ]
        \textbf{Runtime:} 0.0020898816
        \begin{lstlisting}[language=C++, basicstyle=\ttfamily\tiny, breaklines=true, escapechar=@]
{
  // HEAVY INITIALIZATION (O(N^2) Copy)
  // Reconstructs adjacency matrix explicitly before starting
  auto adj = @\colorbox{red!15}{parlay::tabulate(N, [\&](size\_t i) \{}@
    return parlay::tabulate(N, [&](size_t j) {
      return A[i * N + j];
    });
  });

  if (source == dest) return 0;
  parlay::sequence<int> dist(N, -1);
  parlay::sequence<int> q; // Standard non-concurrent structure
  q.push_back(source);
  dist[source] = 0;

  size_t head = 0;
  while (head < q.size()) {
    size_t current_size = q.size();
    // ... neighbor gathering ...
    auto new_frontiers = parlay::flatten(/*...*/);

    if (new_frontiers.empty()) { head = current_size; continue; }

    // SORT/UNIQUE OVERHEAD
    auto unique_frontiers = @\colorbox{red!15}{parlay::unique(new\_frontiers)}@;

    // SEQUENTIAL BOTTLENECK
    // Updates and pushes must happen serially here
    @\colorbox{red!15}{for (int v : unique\_frontiers) \{}@
      dist[v] = dist[q[head]] + 1;
      if (v == dest) return dist[v];
      q.push_back(v);
    }
    head = current_size;
  }
  return -1;
}
        \end{lstlisting}
        \end{tcolorbox}
    \end{minipage}
    \caption{Code comparison for Shortest Path (Problem 19 in ParEval). \textbf{Left (Baseline):} Effectively uses \texttt{std::atomic} and Compare-and-Swap (CAS) to manage visitation state in parallel, resulting in a significantly faster runtime. \textbf{Right (Finetuned):} Chooses a high-overhead initialization step (copying the adjacency matrix via \texttt{tabulate}) and falls back to sequential logic for the queue update loop (red), causing $O(N^2)$ startup cost and serialization bottlenecks. Nonetheless, it shows heavier abstraction usage.}
    \label{fig:pareval_19_gen_comparison}
\end{figure}

\subsection{Case Study: Performance Stability on ParEval Problem 34 (Scan)}
\begin{figure}[h!]
    \centering
    \begin{subfigure}[b]{0.48\columnwidth}
        \includegraphics[width=\linewidth]{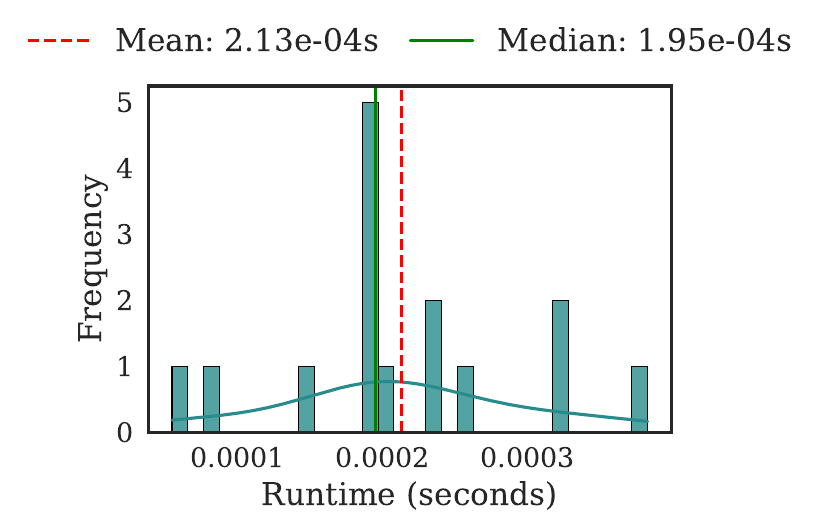}
        \caption{Base Model Runtime Distribution}
    \end{subfigure}
    \begin{subfigure}[b]{0.48\columnwidth}
        \includegraphics[width=\linewidth]{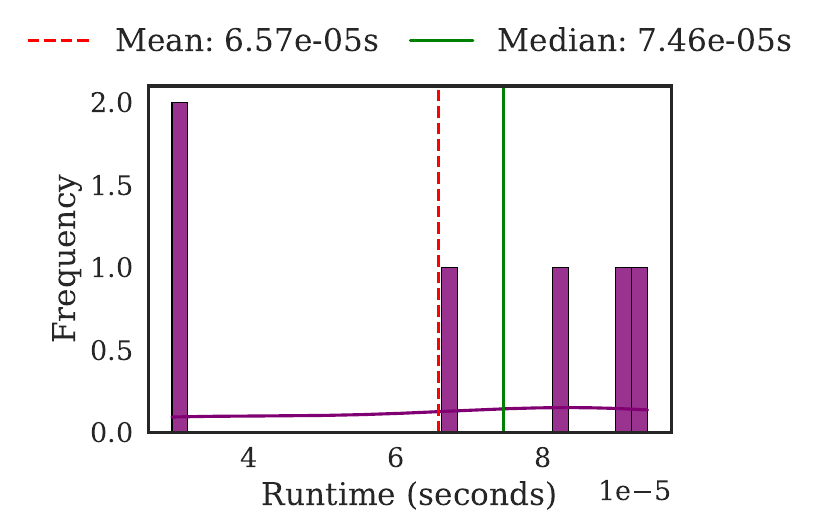}
        \caption{Fine-tuned Runtime Distribution}
    \end{subfigure}
    \caption{Runtime Histograms for ParEval Problem 34 (Scan). The fine-tuned model (b) exhibits tighter variance and highly predictable performance compared to the wider distribution of the base model (a).}
    \label{fig:scan-hist}
\end{figure}

Similarly to the Shortest Path problem, we observe a distinct stabilization of performance in the fine-tuned model for ParEval Problem 34 (Scan), as shown in \cref{fig:scan-hist}. The base model's runtime distribution \cref{fig:scan-hist}(a) is somewhat disjointed, with some runs being very slow and others faster. On the other hand, the fine-tuned model \cref{fig:scan-hist}(b) demonstrates a much tighter, more predictable runtime distribution. This consistency confirms that the fine-tuned \sys model systematically converges on stable and reliable parallel patterns.

\subsection{Failure Modes: Geometric Hallucinations}
While fine-tuning improves general syntax, it can induce ``confident hallucinations'' in domains with specialized logic. In the Convex Hull task (\cref{tab:geometry-results}), the fine-tuned model failed by repeatedly calling a non-existent \texttt{parlay::convex\_hull} function, whereas the base model attempted (and occasionally succeeded at) a manual implementation. This highlights the necessity of the ECA's compiler-feedback loop to catch API hallucinations.

\begin{table}[h]
\centering
\scriptsize
\resizebox{0.5\columnwidth}{!}{
\begin{tabular}{llcc}
\toprule
Problem Type & Model & Pass@1 & Speedup@1 \\
\midrule
10\_convex\_hull & Gemini-2.5-Pro & 0.45 & 1.43 \\
10\_convex\_hull & \textbf{DS-Parlay (\sys)} & \textbf{0.00} & \textbf{0.00} \\
13\_closest\_pair\_2d & Gemini-2.5-Pro & 0.40 & 74.48 \\
13\_closest\_pair\_2d & \textbf{DS-Parlay (\sys)} & \textbf{0.45} & \textbf{188.03} \\
\bottomrule
\end{tabular}
}
\caption{Detailed Geometry Results. The fine-tuned model dominates in Closest Pair but hallucinates APIs in Convex Hull.}
\label{tab:geometry-results}
\end{table}


\section{Prompts}
\paragraph{ParEval Prompts} For the ParEval benchmarks, we adopt the prompting specifications outlined by Nichols et al.~\cite{nichols2024pareval}. We utilize a fixed system instruction alongside language-specific templates for C++ and Rust.

The system prompt provided to the model is as follows (see \cref{fig:pareval_prompts}).

\begin{figure}[h!]
\centering
\begin{tcolorbox}[
    colback=gray!5, 
    colframe=gray!40, 
    title=\textbf{ParEval Prompting Specification},
    fonttitle=\bfseries\small,
    boxrule=0.8pt,
    arc=2pt
]

\textbf{System Instruction:} \\
\small Fixed instruction prepended to all queries.
\begin{lstlisting}[basicstyle=\ttfamily\scriptsize, breaklines=true, numbers=none, frame=none]
You are a **helpful** coding assistant.
You are helping a programmer write a C++ function. Write the body of the function and put it in a markdown code block. 
**Requirements**:
- **DO NOT WRITE ANY COMMENTS OR EXPLANATIONS** in the code!!! Generate **PURE** code!!!
- Before you return the code, make sure to **remove any comments or explanations** that you may have added.
\end{lstlisting}
\vspace{0.5em}
\hrule
\vspace{0.5em}
\textbf{C++ User Template:}
\begin{lstlisting}[basicstyle=\ttfamily\scriptsize, breaklines=true, numbers=none]
Complete the C++ function {function_name}. Only write the body of the function {function_name}. 
```cpp
{prompt}
```
\end{lstlisting}
\vspace{0.5em} \hrule \vspace{0.5em}
\textbf{Rust User Template:} 
\begin{lstlisting}[basicstyle=\ttfamily\scriptsize, breaklines=true, numbers=none] 
Complete the Rust function {function_name}. Only write the body of the function {function_name}.
```Rust
{prompt}
```
\end{lstlisting}
\end{tcolorbox} 
\caption{The prompting strategy adopted from the ParEval paper~\cite{nichols2024pareval}. The templates include specific placeholders (\texttt{{function\_name}}, \texttt{{prompt}}) populated dynamically during evaluation.} 
\label{fig:pareval_prompts} 
\end{figure}

\paragraph{Extending ParEval for Parallel Libraries}
Since the original ParEval dataset lacks native support for ParlayLib and Rust, we manually curated task-specific prompts to bridge this gap. These prompts preserve the original problem semantics while explicitly requesting the use of specific parallel frameworks (ParlayLib for C++ and Rayon for Rust). Figure~\ref{fig:pareval_extension} demonstrates how a standard Discrete Fourier Transform (DFT) task is adapted for both languages.

\begin{figure}[h]
\centering
\begin{tcolorbox}[
    colback=gray!5, 
    colframe=gray!40, 
    title=\textbf{ParEval Extension Examples},
    fonttitle=\bfseries\small,
    boxrule=0.8pt,
    arc=2pt
]

\textbf{C++ Prompt (ParlayLib):}
\begin{lstlisting}[language=C++, basicstyle=\ttfamily\scriptsize, keywordstyle=\color{blue}, commentstyle=\color{green!60!black}, breaklines=true, numbers=none]
/* Compute the discrete fourier transform of x. Store the result in output.
   Use ParlayLib to compute in parallel.
   Example:
     input: [1, 4, 9, 16]
     output: [30+0i, -8-12i, -10-0i, -8+12i]
*/
void dft(parlay::sequence<double> const& x, 
         parlay::sequence<std::complex<double>> &output) {
\end{lstlisting}

\vspace{0.5em}
\hrule
\vspace{0.5em}

\textbf{Rust Prompt (Rayon):}
\begin{lstlisting}[language=Rust, basicstyle=\ttfamily\scriptsize, keywordstyle=\color{purple}, commentstyle=\color{gray}, breaklines=true, numbers=none]
/* Compute the discrete fourier transform of x. Store the result in output.
   Use Rust Rayon to compute in parallel.
   Example:
     input: [1, 4, 9, 16]
     output: [30+0i, -8-12i, -10-0i, -8+12i]
*/
pub fn dft(x: &[f64], output: &mut [num_complex::Complex<f64>]) {
\end{lstlisting}

\end{tcolorbox}
\vspace{-1em}
\caption{Representative examples of our manual extensions to the ParEval dataset. The prompts are tailored to enforce specific parallel backends while maintaining identical input/output specifications.}
\label{fig:pareval_extension}
\end{figure}

\paragraph{PBBSBench Prompting Strategy}
We employ two distinct prompting strategies for PBBSBench to evaluate the model's ability to utilize context:
\begin{itemize}
    \item \textbf{Concise Prompts:} These contain only the natural language problem description and the target function signature.
    \item \textbf{Augmented Prompts:} These extend the concise version by including definitions for necessary ParlayLib primitives, custom data structures (e.g., \texttt{Graph}), and helper utilities (e.g., \texttt{Graph\_io}) defined within the PBBSBench environment.
\end{itemize}

Figure~\ref{fig:pbbs_concise} illustrates an example of the concise prompting format.

\begin{figure}[h]
\centering
\begin{tcolorbox}[
    colback=gray!5, 
    colframe=gray!40, 
    title=\textbf{PBBSBench Strategy: Concise Prompt},
    fonttitle=\bfseries\small,
    boxrule=0.8pt,
    arc=2pt
]

\textbf{System Instruction:}
\begin{lstlisting}[basicstyle=\ttfamily\scriptsize, breaklines=true, numbers=none, frame=none]
You are an expert C++ programmer with extensive experience in parallel programming. 
Write a parallel {} procedure in C++ that is correct and is the fastest parallel {} program you can generate.
Return the code between `// --- Start of file:` and `// --- End of file:` markers.
\end{lstlisting}

\vspace{0.5em}
\hrule
\vspace{0.5em}

\textbf{User Input (Example: Maximal Independent Set):}
\begin{lstlisting}[language=C++, basicstyle=\ttfamily\scriptsize, keywordstyle=\color{blue}, breaklines=true, numbers=none]
Returns a maximal independent set for an undirected graph. Use ParlayLib to compute in parallel.

#include "common/graph.h"

using vertexId = uint;
using edgeId = uint;
using Graph = graph<vertexId,edgeId>;

parlay::sequence<char> maximalIndependentSet(Graph const &G);
\end{lstlisting}

\end{tcolorbox}
\vspace{-1em}
\caption{An example of the \textit{Concise} prompt formulation for the PBBSBench \texttt{maximalIndependentSet} task. The model is provided with the function signature and a request to use ParlayLib, but implementation details of the \texttt{Graph} structure are omitted.}
\label{fig:pbbs_concise}
\end{figure}

\paragraph{RPB Prompting Strategy}
The prompting strategy for the RPB benchmarks relies on a composite structure. Each prompt comprises two distinct segments: (1) a context block defining Rust primitives that replicate ParlayLib functionality (e.g., \texttt{flatten}), and (2) the specific problem statement, including allowed libraries and the target function signature. 

\begin{figure}[h!]
\centering
\begin{tcolorbox}[
    colback=gray!5, 
    colframe=gray!40, 
    title=\textbf{RPB Prompt Structure},
    fonttitle=\bfseries\small,
    boxrule=0.8pt,
    arc=2pt
]


\textbf{Part 1: Context (Excerpt of ParlayLib-Rust Primitives)} \\
\small The prompt begins by providing the full suite of helper functions (truncated here for brevity).
\begin{lstlisting}[language=Rust, basicstyle=\ttfamily\scriptsize, keywordstyle=\color{purple}, commentstyle=\color{gray}, breaklines=true, numbers=none]
// Here are the primitives you may use
// ... [Full list of primitives omitted] ...

/* -------------------- Flatten -------------------- */
pub fn flatten<T>(arr: &[&Vec<T>], dest: &mut Vec<T>)
where T: Copy + Send + Sync + Default {
    // ... implementation details ...
}

// ... [Additional primitives like scan, reduce, etc.] ...
\end{lstlisting}

\vspace{0.5em}
\hrule
\vspace{0.5em}

\textbf{Part 2: Task Definition \& Signature} \\
\small The specific algorithm request follows the context.
\begin{lstlisting}[language=Rust, basicstyle=\ttfamily\scriptsize, keywordstyle=\color{purple}, commentstyle=\color{gray}, breaklines=true, numbers=none]
// Given an undirected graph, return a maximal independent set (MIS).
// The input graph can be in any format.
// The code cannot reorder the graph for locality.
// The output must be a sequence of vertices in the MIS (order irrelevant).

#[cfg(feature = "AW_safe")]
use std::sync::atomic::{AtomicU8, Ordering::Relaxed};
use rayon::prelude::*;
use pbbs::common::graph::Graph;

#[path="../../common/spec_for.rs"] mod spec_for;
use spec_for::StatefulSpecFor;

#[derive(Clone)]
struct MISState {
    flag: u8,
}

pub fn maximal_independent_set(g: &Graph) -> Vec<u8> {
    // LLM_OUTPUT_HERE
}
\end{lstlisting}

\end{tcolorbox}
\vspace{-1em}
\caption{An example of the RPB prompting template. We inject the full set of parallel primitive definitions (represented by the \texttt{flatten} excerpt in Part 1) prior to the specific task instructions (Part 2) to ground the model in the available Rust-ParlayLib equivalence layer.}
\label{fig:rpb_prompts}
\end{figure}



\section{Fine-tuning on Low-level APIs vs. ParlayLib Token-efficiency}
\label{app:parlay-vs-omp}

To assess whether the choice of parallel abstraction affects fine-tuning efficiency, we compare \texttt{deepseek-coder-6.7b-base} fine-tuned on our Parlay-Instruct dataset against an equivalent model fine-tuned on a token-matched OpenMP dataset.

The OpenMP dataset was sourced from \texttt{hpcgroup/hpc-instruct} and cleaned to match the 2.14M token budget of Parlay-Instruct, isolating core algorithmic loops to provide a dense learning signal. The filtering process consisted of:

\begin{enumerate}
    \item Dropping translation tasks and non-native OpenMP languages (Python, Chapel, OpenCL, CUDA).
    \item Retaining only samples with actual \texttt{\#pragma omp} or \texttt{!\$omp} directives in the solution.
    \item Dropping outputs from weaker generator models (e.g., Mixtral, DBRX).
    \item Deduplicating by seed, keeping the best model output per seed.
    \item Trimming the longest samples to hit the 2.14M token target, retaining shorter, more focused examples.
\end{enumerate}

Both models were trained with identical SFT+LoRA hyperparameters (LoRA $r=8$, $\alpha=16$, dropout $=0.05$, \texttt{max\_seq\_length}\,$=2048$, LR\,$=2\times10^{-4}$). Results on three representative PBBS problems are shown in Table~\ref{tab:parlay-vs-omp}.

\begin{table}[h]
\centering
\small
\begin{tabular}{llrrrr}
\toprule
\textbf{Model} & \textbf{Exec. Model} & \textbf{Problem Name} & \textbf{build@1} & \textbf{pass@1} & \textbf{speedup@1} \\
\midrule
DeepSeek-Syntax & parlay & \texttt{06\_fft\_dft}                 & 0.937 & 0.020 & 30.754 \\
DeepSeek-Syntax & parlay & \texttt{16\_graph\_largest\_component} & 0.856 & 0.060 & 3.517  \\
DeepSeek-Syntax & parlay & \texttt{19\_graph\_shortest\_path}    & 0.919 & 0.080 & 2.187  \\
DeepSeek-OMP    & omp    & \texttt{06\_fft\_dft}                 & 1.000 & 0.550 & 23.180 \\
DeepSeek-OMP    & omp    & \texttt{16\_graph\_largest\_component} & 1.000 & 0.000 & 0.000  \\
DeepSeek-OMP    & omp    & \texttt{19\_graph\_shortest\_path}    & 0.960 & 0.100 & 0.000  \\
\bottomrule
\end{tabular}
\caption{Token-matched comparison of Parlay-Instruct vs. OpenMP fine-tuning on \texttt{deepseek-coder-6.7b-base}.}
\label{tab:parlay-vs-omp}
\end{table}

Under a fixed token budget, the Parlay-Instruct model achieves higher speedups on successful runs, while the OpenMP model produces more frequently compiling and passing solutions but with limited parallel gains. This suggests that high-level abstractions like ParlayLib provide a more token-efficient signal for learning parallel reasoning, whereas explicit low-level thread and memory management consumes capacity that might otherwise go toward algorithmic structure. Full PBBSBench evaluations will appear in a subsequent revision.

\section{Ablation Study on Agent Architecture and Search Strategies}
\label{app:agent-ablation}

To rigorously evaluate the structural contributions of our evolutionary pipeline, we conducted an ablation study utilizing an open-weight local model (\textbf{Qwen3-Coder-30B-Instruct}). A mid-sized open model is employed to eliminate confounding variables associated with opaque, massive-scale models, and to ensure that any observed performance gains are directly attributable to our architectural advancements—specifically, the integration of MAP-Elites and dynamic execution feedback—rather than sheer parameter count. 

The baseline model for all non-finetuned configurations is \textbf{Qwen3-Coder-30B-Instruct}, while our complete pipeline utilizes the fine-tuned \textbf{Qwen-Parlay}. We evaluate the following configurations:

\begin{itemize}
    \item \textbf{Best-of-N}: Standard independent sampling without execution-driven feedback.
    \item \textbf{Best-of-N + CoT}: Independent sampling augmented with Chain-of-Thought prompting.
    \item \textbf{Self-Refine}: A single-lineage, text-only iterative refinement strategy.
    \item \textbf{ECA-base}: Our Evolutionary Coding Agent (ECA) pipeline applied to the baseline model, lacking domain-specific fine-tuning.
    \item \textbf{ECA\_no\_div}: The ECA pipeline utilizing runtime execution feedback, but with the MAP-Elites diversity archive disabled.
    \item \textbf{ECA-ft (Full ParEVO)}: Our proposed methodology, combining the fine-tuned \textbf{Qwen-Parlay} model with the complete, diversity-driven ECA pipeline.
\end{itemize}

\begin{table}[h]
\centering
\small
\begin{tabular}{lrrrrrr}
\toprule
\textbf{Problem ID} & \textbf{Best-of-N} & \textbf{Best-of-N + CoT} & \textbf{Self-Refine} & \textbf{ECA-base} & \textbf{ECA\_no\_div} & \textbf{ECA-ft (Full)} \\
\midrule
\texttt{coci11c1p3} & 0.1019 & 0.1018 & 0.3200 & 0.1054 & 0.1017 & \textbf{0.0297} \\
\texttt{ccc16j3}    & 0.0026 & 0.0025 & 0.0031 & 0.0030 & 0.0026 & \textbf{0.0019} \\
\texttt{coci06c1p5} & -      & -      & 0.0040 & 0.0068 & -      & 0.0069 \\
\texttt{coci12c1p4} & -      & 0.0265 & 0.0264 & 0.0774 & 0.0267 & \textbf{0.0147} \\
\texttt{coci17c4p1} & 0.0030 & 0.0030 & 0.0030 & 0.0032 & 0.0018 & \textbf{0.0017} \\
\texttt{coci22c3p1} & 0.0027 & 0.0031 & 0.0032 & 0.0030 & -      & 0.0030 \\
\texttt{crci08p1}   & 0.0030 & 0.0031 & 0.0018 & 0.0033 & -      & \textbf{0.0017} \\
\bottomrule
\end{tabular}
\caption{Ablation results evaluating runtime performance (in seconds) across diverse competitive programming challenges. Problems ids refer to problems in the DMOJ dataset. A ``-'' indicates that the generated solution failed to pass the required unit tests, resulting in no valid runtime.}
\label{tab:agent-ablation}
\end{table}

The results demonstrate that ParEVO's primary advantage over standard iterative repair lies in its ability to escape local optima. Single-lineage refinement methods (e.g., Self-Refine and \texttt{ECA\_no\_div}) frequently converge on structurally flawed algorithms, exhaustively attempting syntactic patches without exploring alternative logic. By leveraging MAP-Elites to enforce structural diversity (e.g., varying synchronization primitives and underlying data structures), \texttt{ECA-ft} successfully navigates the broader search space to discover highly optimized parallel implementations where simpler pipelines fail to yield functionally correct solutions.

\section{Ablation Study on Iteration Budgets and Fine-Tuning Synergy}
\label{app:eca-iteration-ablation}
This section investigates the combined efficacy of fine-tuning and the Evolutionary Coding Agent (ECA). We evaluate the synergy of fine-tuning alongside the ECA pipeline across various iteration budgets and diversity settings on selected DMOJ problems.

For these experiments, the base model is \textbf{Gemini-2.5-Pro}, and the fine-tuned variant is \textbf{Gemini-2.5-Parlay}. We define the following ablation configurations:

\begin{itemize}
    \item \textbf{ECA-Iter5, ECA-Iter15, ECA-Iter30}: The baseline Gemini-2.5-Pro model utilizing the standard ECA pipeline (with a diversity archive of 5 programs) capped at 5, 15, and 30 evolutionary iterations, respectively.
    \item \textbf{ECA-No-Diversity}: The baseline Gemini-2.5-Pro model running for 30 iterations, but with the MAP-Elites diversity mechanism disabled.
    \item \textbf{ECA-Finetuned (Full Synergy)}: Our fully proposed pipeline, pairing the fine-tuned Gemini-2.5-Parlay model with the complete ECA pipeline for 30 iterations.
\end{itemize}

\begin{table}[h]
\centering
\small
\begin{tabular}{lrrrrrr}
\toprule
\textbf{Problem ID} & \textbf{ECA-Iter5} & \textbf{ECA-Iter15} & \textbf{ECA-Iter30} & \textbf{ECA-No-Diversity} & \textbf{ECA-Finetuned} \\
\midrule
\texttt{cco08p4}    & -       & 0.00753 & 0.00698 & 0.00927 & \textbf{0.00706} \\
\texttt{coci19c1p3} & -       & -       & -       & -       & \textbf{0.02424} \\
\texttt{coci11c1p3} & 0.02606 & 0.02565 & 0.02534 & \textbf{0.00822} & 0.02681 \\
\texttt{coci23c2p2} & 0.03474 & 0.03166 & \textbf{0.03130} & 0.03500 & 0.03608 \\
\bottomrule
\end{tabular}
\caption{Ablation results detailing runtime performance (in seconds) on DMOJ problems. A ``-'' indicates that the generated solution failed to pass the requisite tests, yielding no valid runtime. Iteration configurations dictate the maximum number of evolutionary cycles permitted.}
\label{tab:eca-iteration-ablation}
\end{table}

Crucially, \textbf{ECA-Finetuned} emerges as the most robust model configuration. On exceptionally complex algorithms—such as the graph problem \texttt{coci19c1p3}—every base-model configuration failed to generate a valid parallel solution, regardless of the iteration allowance. Only the combination of domain-specific fine-tuning and the evolutionary search successfully navigated the solution space to yield a correct, optimized implementation. Furthermore, this variant maintains highly competitive runtimes across all other evaluated problems.

Additionally, isolating the search parameters reveals two critical methodological trends:
\begin{enumerate}
    \item \textbf{Iteration Budgets:} Expanding the evolutionary horizon consistently yields tighter, more optimized execution schedules. For instance, on \texttt{cco08p4}, a constrained budget of 5 iterations fails entirely, 15 iterations secures a valid solution, and 30 iterations discovers the most time-efficient implementation.
    \item \textbf{Advantage of MAP-Elites Diversity:} Disabling the structural diversity constraints (\texttt{ECA-No-Diversity}) reliably traps the agent in local optima. Without the pressure to explore alternative concurrency primitives or data structures, the search can sometimes lead to invalid or degraded solutions.
\end{enumerate}

\section{Integrity of Unit Tests in Training Data Verification}
\paragraph{Test integrity.} A natural concern with LLM-generated training data is whether the model effectively grades itself with trivial tests. Our pipeline avoids this by reusing the human-authored test logic from the 593 seed programs. The mutation operators alter intermediate algorithmic logic and datatypes for diversity, but they preserve the deterministic input--output mapping of each seed. The Teacher therefore synthesizes new test code only when $\mathcal{M}_{type}$ changes the input or output type; in all other cases the original human-written unit test is reused. The vast majority of the corpus is thus evaluated against tests the LLM did not write. Additionally, we uniformly sampled 100 programs from the 13{,}820 tasks for manual review, confirming that reused tests matched their seed and that synthesized datatype-adapted tests preserved the original semantics. All sampled programs achieved 100\% line coverage, ensuring no parallel logic escaped verification.

We additionally compiled and executed the full corpus under ThreadSanitizer (TSan). TSan flagged approximately 150 candidate races or runtime errors; manual inspection of 50 flagged cases identified no genuine data races. This is consistent with the structural properties of the code we generate: ParlayLib's functional, lock-free data-parallel primitives, used in place of explicit thread management, make data races and deadlocks structurally unlikely.

\section{Leakage and Near-Duplicate Overlap Analysis}
\label{app:leakage}

To verify that our reported results reflect genuine generalization rather than memorization of evaluation problems, we analyze the relationship between our training corpora and the benchmarks used for evaluation.

\paragraph{Held-out tasks.} The 700 held-out tasks in Parlay-Instruct corpus function exclusively as a validation set during fine-tuning and are not used in any of the final evaluations (PBBSBench, ParEval, RPB, DMOJ). Overlap between the training and validation splits therefore does not affect reported results.

\paragraph{Evaluation benchmarks.} The evaluation benchmarks are disjoint from our training data. The Parlay-Instruct corpus, synthesized from ParlayLib, is used to fine-tune the C++ models, while a Rust version of DMOJ is used to fine-tune the Rust models. All base and fine-tuned models share identical pretraining data to ensure fair comparison.

\paragraph{N-gram overlap analysis.} We performed an exact $N$-gram overlap analysis ($N = 10, 15$) using Jaccard similarity, comparing our fine-tuning corpora against each evaluation suite (Parlay-Instruct against ParEval, DMOJ, and PBBSBench; Rust-DMOJ against RPB). Results are summarized below.

\begin{itemize}
    \item \textbf{ParEval and DMOJ (C++):} Negligible overlap ($< 0.00002$ for 10-grams). All matches consisted of standard library dependencies (e.g., \texttt{\#include <algorithm>}) and generic variable assignments.
    \item \textbf{PBBSBench (C++):} Minimal overlap ($< 0.6\%$ for 10-grams). Manual review confirmed these matches are predominantly C++ header usage and structural \texttt{parlaylib} boilerplate required for compilation.
    \item \textbf{Rust Performance Benchmarks (RPB):} Near-zero leakage ($0.0$ for 15-grams; $0.000004$ for 10-grams). The single overlapping 10-gram is standard Rust comparison boilerplate (\texttt{fn partial\_cmp...}).
    \item \textbf{DMOJ fine-tune problems vs. DMOJ test problems:} $0.002559$ for 10-grams.
\end{itemize}

\paragraph{Summary.} The maximum structural overlap across all evaluation suites is below $0.6\%$, and manual inspection confirms these matches consist of syntactic boilerplate rather than solution content. We find no evidence of evaluation problem solutions appearing in the training data.

\section{Semantic Alignment Example}
\cref{fig:code_comparison} is an example of the semantic alignment achieved by finetuning.
\begin{figure}[H]
    \centering
    \lstset{
        language=C++,
        basicstyle=\ttfamily\footnotesize,
        keywordstyle=\color{blue}\bfseries,
        commentstyle=\color{green!50!black}\itshape,
        stringstyle=\color{red!60!black},
        numbers=left,
        numberstyle=\tiny\color{gray},
        stepnumber=1,
        frame=single,
        rulecolor=\color{black!30},
        breaklines=true,
        captionpos=b,
        tabsize=2,
        showstringspaces=false
    }

    \textbf{Input Prompt:}
    \begin{lstlisting}[numbers=none, frame=tb]
#include <parlay/primitives.h>
#include <complex>
/* Sort vector x of complex numbers by magnitude (ascending) using ParlayLib. */
void sortComplexByMagnitude(parlay::sequence<std::complex<double>>& x) {
    \end{lstlisting}

    \vspace{0.2cm}

    \begin{minipage}[t]{0.48\textwidth}
        \textbf{Base Model (Gemini-2.5-Pro)}
        \begin{lstlisting}[firstnumber=1]
{
    // Build Error: parlay::sort expects 
    // different comparator signature
    parlay::sort(x, [](const std::complex<double>& a, const std::complex<double>& b) {
        return std::abs(a) < std::abs(b);
    });
}
        \end{lstlisting}
        \textit{Status: Compilation Failed (Build@1=0)}
    \end{minipage}
    \hfill
    \begin{minipage}[t]{0.48\textwidth}
        \textbf{Fine-tuned Model (\sys)}
        \begin{lstlisting}[firstnumber=1]
{
    // Correct: uses sort_inplace and 
    // generic auto params for complex type
    parlay::sort_inplace(x, [](const auto& a, const auto& b) {
        return std::abs(a) < std::abs(b);
    });
}
        \end{lstlisting}
        \textit{Status: Passed ($17.5\times$ Speedup)}
    \end{minipage}

    \vspace{-0.2cm}
    \caption{\textbf{Semantic Alignment Example.} The base model (top) fails to compile due to incorrect API usage and strict type definitions in the lambda. The fine-tuned model (bottom) correctly identifies \texttt{sort\_inplace} and uses \texttt{auto} to handle the complex number types safely.}
    \label{fig:code_comparison}
\end{figure}

\end{document}